\documentclass[sn-mathphys]{sn-jnl}% Math and Physical Sciences Reference Style
\jyear{2021}%

\theoremstyle{thmstyleone}%
\theoremstyle{thmstyletwo}%

\theoremstyle{thmstylethree}%

\usepackage{hyperref}
\usepackage{amsfonts}
\usepackage{amsmath}
\usepackage{amssymb}
\usepackage{mathtools}
\usepackage{bm}   % bm command
\usepackage{cleveref} % auto refs formulas
\usepackage{subfigure}
\usepackage[export]{adjustbox}
\usepackage{algorithm}
\usepackage{algpseudocode}
\usepackage{xcolor}
\usepackage{wrapfig}
\usepackage[nodisplayskipstretch]{setspace}
\usepackage{caption}

\usepackage{array}
\newcommand{\PreserveBackslash}[1]{\let\temp=\\#1\let\\=\temp}
\newcolumntype{C}[1]{>{\PreserveBackslash\centering}p{#1}}
\newcolumntype{R}[1]{>{\PreserveBackslash\raggedleft}p{#1}}
\newcolumntype{L}[1]{>{\PreserveBackslash\raggedright}p{#1}}

\newcommand{\centered}[1]{\begin{tabular}{c} #1 \end{tabular}}

\begin{document}

\title[Learning Multi-Agent Coordination through Connectivity-driven Communication]{Learning Multi-Agent Coordination through Connectivity-driven Communication}


\author[2]{\fnm{Emanuele} \sur{Pesce}}\email{e.pesce@warwick.ac.uk}

\author*[1,2,3]{\fnm{Giovanni} \sur{Montana}}\email{g.montana@warwick.ac.uk}

\affil[1]{\orgdiv{Department of Statistics}, \orgname{University of Warwick}, \orgaddress{\city{Coventry}, \postcode{ CV4 7AL}, \country{UK}}}

\affil[2]{\orgdiv{WMG}, \orgname{University of Warwick}, \orgaddress{\city{Coventry}, \postcode{ CV4 7AL}, \country{UK}}}

\affil[3]{\orgdiv{Alan Turing Institute}, \orgaddress{\city{London}, \postcode{ NW1 2DB}, \country{UK}}}


\abstract{
In artificial multi-agent systems, the ability to learn collaborative policies  is predicated upon the agents' communication skills: they must be able to encode the information received from the environment and learn how to share it with other agents as required by the task at hand. We present a deep reinforcement learning approach, Connectivity Driven Communication (CDC), that facilitates the emergence of multi-agent collaborative behaviour only through  experience. The agents are modelled as nodes of a weighted graph whose state-dependent edges encode pair-wise messages that can be exchanged. We introduce a graph-dependent attention mechanisms that controls how the agents' incoming messages are weighted. This mechanism takes into full account the current state of the system as represented by the graph, and builds upon a diffusion process that captures how the information flows on the graph. The graph topology is not assumed to be known a priori, but depends dynamically on the agents' observations, and is learnt concurrently with the attention mechanism and policy in an end-to-end fashion. Our empirical results show that CDC is able to learn effective collaborative policies and can over-perform competing learning algorithms on cooperative navigation tasks.
}


\maketitle

\section{Introduction}

In reinforcement learning (RL), an agent learns to take sequential decisions by mapping its observations of the world to actions using a reward as feedback signal \cite{sutton1998introduction}. In the last few years, deep artificial neural networks \cite{lecun2015deep,schmidhuber2015deep} have been leveraged to improve the learning ability of RL algorithms in a number of ways, e.g. as policy function approximators to map observations to actions and to learn informative data representations. The resulting deep reinforcement learning algorithms (DRL) have recently achieved unprecedented performance in  single-agent tasks, e.g. in playing Go  \cite{silver2016mastering} and Atari games \cite{mnih2015human,vinyals2019grandmaster}.

Multi-agent reinforcement learning (MARL) extends RL to problems characterized by the interplay of multiple agents operating in a shared environment. This is a scenario that is typical of many real-world applications including robot navigation \cite{tanner2005towards}, autonomous vehicles coordination \cite{brunet1995multi}, traffic management \cite{dresner2004multiagent}, and supply chain management \cite{lee2008multi}. Compared to single-agent systems, MARL  presents additional layers of complexity. When multiple learners interact with each other, the environment becomes highly non-stationary from the point of view of each individual actor \cite{hernandez2017survey}. Moreover, credit assignment \cite{rahaie2009toward}, which is the ability to determine how the actions of each individual agent impact on the overall system performance, becomes particularly difficult  \cite{harati2007knowledge,yliniemi2014multi,agogino2004unifying}.

We are interested in systems involving agents that autonomously learn how to collaborate  in order to achieve a shared outcome. When multiple agents are expected to develop a cooperative behaviour, an important need emerges: an adequate communication protocol must be established to support the level of coordination that is necessary to solve the task. The fact that communication plays a critical role in achieving synchronization in multi-agent systems has been extensively documented \cite{vorobeychik2017does,demichelis2008language,miller2004communication,kearns2012experiments,foerster2016learning,sukhbaatar2016learning,singh2018learning,pesce2019improving}. Building upon this evidence, a number of multi-agent DRL algorithms (MADRL) have been developed lately which try to facilitate the spontaneous emergence of communication strategies during training. In particular, significant efforts have gone into the development of attention mechanisms for filtering out irrelevant information \cite{jiang2018learning,mao2018modelling,liu2020multi,hoshen2017vain,das2018tarmac,iqbal2018actor,wang2019learning} (see also Section 4).

In this paper we introduce a MADRL algorithm for cooperative multi-agent tasks. Our approach relies on learning a state-dependent communication graph whose topology controls what information should be exchanged within the system and how this information should be distributed across agents. As such, the communication graph plays a dual role. First, it represents how every pair of agents jointly encodes their observations to form local messages to be shared with others. Secondly, it controls a mechanism by which local messages are propagated through the network to form agent-specific information content that is ultimately used to make decisions. As we will demonstrate, this approach supports the emergence of a collaborative decision making policy. The core idea we intend to exploit is that, given any particular state of the environment, the graph topology should be self-adapting to support the most efficient information flow. This raises the question: how should efficiency be measured?  

Our proposed approach, {\it connectivity-driven communication} (CDC), is inspired by the process of heat transference in a graph, and specifically the heat kernel (HK). The HK describes the effect of applying a heat source to a network and observing the diffusion process over time. As such, it can be used to characterise the way in which the information flows across nodes. The HK has been used in a number of different application domains  where there is a need to characterise the topology of graph, e.g. in 3D object recognition \cite{zhang2008graph} and neuroimaging \cite{chung2016classifying,chung2016characterising}. Various metrics obtained from the HK have been used to organise the  intrinsic  geometry  of a  network  over  multiple-scales by capturing  local  and  global shapes’ in relation to a node via a time parameter. The HK also incorporates a concept of node influence as  measured  by  heat  propagation in a network, which can be exploited to characterise how efficiently the information propagates between any pair of nodes. To the best of our knowledge, this is the first time that the HK has been used to develop an end-to-end learnable attention mechanism enabling multi-agent cooperation.

Our approach relies on an actor-critic paradigm \cite{degris2012off,silver2014deterministic,lillicrapHPHETS15} and is intended to extend the centralized-learning with decentralized-execution (CLDE) framework \cite{foerster2016learning,lowe2017multi}. In CDC, all the observations from each agent are assumed known only during the training phase whilst during execution each agent makes autonomous decisions using only their own information. The entire model is learned end-to-end supported by the fact that the heat-kernel is a differentiable operator allowing the gradients to flow throughout the architecture.  The performance of CDC has been evaluated against alternative methods on four cooperative navigation tasks. Our experimental evidence demonstrates that CDC is capable of outperforming other relevant state-of-the-art algorithms. In addition, we analyse the communication patterns discovered by the agents to illustrate how interpretable topological structures can emerge in different scenarios.

The structure of this work is as follows. In Section \ref{sec:related_works} we discuss related state-of-the-art MADRL methods focusing on cooperating systems with communication mechanisms. In Section \ref{sec:method} we provide the details of the proposed CDC algorithm. Experimental results are then provided in Section \ref{sec:experiments}. Finally, in Section \ref{sec:conclusions}, we discuss the benefits and potential limitations of the proposed methodology with a view on further improvements in future work.  

\section{Related Work} \label{sec:related_works}

Multi-agent systems have been widely studied in a number of different domains, such as machine learning \cite{stone2000multiagent}, game theory \cite{parsons2002game} and distributed systems \cite{shoham2008multiagent}. Recent advances in deep reinforcement learning have allowed multi-agent systems capable of autonomous decision-making \cite{nguyen2020deep,hernandez2019survey,albrecht2018autonomous} improving tabular-based solutions \cite{busoniu2008comprehensive}. In this section, we briefly review recent developments in MADRL  with a focus on communication strategies that have been proposed to improve cooperation. 

\subsection{Centralised learning with decentralised execution}

When multiple learners interact with each other, the environment becomes  non-stationary from the perspective of individual agents which results in increased training instability \cite{tuyls2012multiagent,laurent2011world}. An approach that has proved particularly effective consists of training the agents assuming centralised access to the entire system's information whilst executing the policies in a decentralised manner (CLDE) \cite{kraemer2016multi,foerster2016learning,foerster2017counterfactual,lowe2017multi,pesce2019improving,iqbal2018actor}. During training, a critic module  has access to information related to other agents, i.e.  their actions and observations. MADDPG \cite{lowe2017multi}, for example, extends DDPG \cite{silver2014deterministic} in this fashion: each agent has a centralised critic providing feedback to the actors, which decide what actions to take. A variant of this approach has recently been proposed to deal with partially observable environments through the use of recurrent neural networks \cite{wang2020r,hochreiter1997long}. In \cite{foerster2017counterfactual}, a centralised critic is used to estimate the Q-function whilst decentralised actors optimise the agents' policies. In \cite{lin2018efficient}, an action-value critic network coordinates decentralised policy networks for a fleet management problem. 

\subsection{Communication methods}

Communication has always played a crucial role in facilitating synchronization and coordination  \cite{scardovi2008synchronization,wen2012consensus,wunder2009communication,ito2011innovations,fox2000probabilistic}. 
Some of the recent MADRL approaches facilitate the emergence of novel communication protocols through communication mechanisms. 
For example, in CommNet \cite{sukhbaatar2016learning}, the hidden states of an agent' neural network are first averaged and then used jointly with the agent's own observations to decide what action to take. Similarly, in \cite{peng2017multiagent}, communication is enabled by connecting agents' policies through a bidirectional recurrent neural network that can produce higher-level information to be shared. In IC3Net \cite{singh2018learning}, a gating mechanism decides whether to allow or block access to other agents' hidden states. 

Other approaches have introduced explicit communication mechanisms that can be learnt from experience. For instance, in RIAL \cite{foerster2016learning}, each agent learns a simple encoding that is transferred over a differentiable channel and allows the gradient of the Q-function to flow; this enables an agent's feedback to take into account the exchanged information. In our previous work, \cite{pesce2019improving}, the agents are equipped with a memory device allowing them to write and read signals to be shared within the system. The communication mechanism we propose in this paper is also explicit; messages are signals that must be shared within the system in order to maximize the shared rewards and serve no other purpose.  

\subsection{Attention mechanisms to support communication}

In a collaborative decision making context, attention mechanisms are used to selectively identify relevant information coming from the environment and other agents that should be prioritised to infer better policies. For example, in \cite{jiang2018learning}, the agents first encode their observations to produce messages; then an attention unit, implemented as a recurrent neural network (RNN), probabilistically controls which incoming messages are used as inputs for the action selection network. The CommNet algorithm \cite{sukhbaatar2016learning} has been extended using a multi-agent predictive modeling approach \cite{hoshen2017vain} which captures the locality of interactions and improves performance by determining which agents will share information. 
{\color{black}
In the IS algorithm \cite{kim2020communication} the agents predict their future trajectories, and these predictions are utilised by an attention mechanism module to compose a message determining the next actions to take. 
} % end color_red
The TarMac algorithm \cite{das2018tarmac} instead leverages the signature-based attention model originally proposed in \cite{vaswani2017attention}. Here, each agent receives the messages broadcasted by others and produces a query that helps select what information to keep and what to discard. The latter approach is closely related to the work proposed in this paper; ours agents also aggregate information coming from different sources in order to maximise their final reward.

\subsection{Diffusion processes on graphs}

Spectral graph theory allows to relate the properties of a graph to its spectrum by analysing its associated eigenvectors and eigenvalues \cite{chung1997spectral,brouwer2011spectra,cvetkovic1980spectra}. The heat kernel falls in this category; it is a powerful and well-studied operator allowing to study certain properties of a graph by solving the heat diffusion equation. The HK is determined by exponentiating the graph's Laplacian eigensystem \cite{differentialgeometry} over time. The resulting features can be used to study the graph's topology and have been utilised across different applications whereby graphs are naturally occurring data structures; e.g. the HK has been used for community detection \cite{kloster2014heat}, data manifold extraction \cite{lafferty2005diffusion}, network classification \cite{chung2016characterising} and image smoothing \cite{zhang2008graph} amongst others. In recent work, the HK has been adopted to extend graph convolutional networks \cite{xu2020graph} and define edge structures supporting convolutional operators \cite{klicpera2019diffusion}. In this work, we use the HK to characterise the state-dependent topology of a multi-agent communication network and learn how the information should flow within the network. 

\subsection{Graph-based communication mechanisms}

Graph structures provides a natural framework for modelling interactions in RL domains \cite{kschischang2001factor,kuyer2008multiagent,guestrin2002multiagent}. Lately, Graph Neural Networks (GNNs) have also been adopted to learn useful graph representations in cooperative multi-agent systems \cite{liao2021review,zhou2021ast,huang2019stgat,mohamed2020social,xu2021learning}. For example, graphs have been used to model spatio-temporal dependencies within episodes for traffic light control \cite{wang2019stmarl}, and to infer a multi-agent connectivity structure which, once processed by a GNN, generates the features required to decide what action to take \cite{li2020deep,jiang2018graph,chen2020gama}. Heterogeneous graph attention networks \cite{seraj2021heterogeneous} have been introduced to learn efficient and diverse communication models for coordinating heterogeneous  agents. Graph convolutional networks capturing multi-agent interactions have also been combined with a counterfactual policy gradient algorithm to deal with the credit assignment problem \cite{su2020counterfactual}. 

GNNs have also supported the development of multi-stage attention mechanisms. For instance, \cite{liu2020multi} describe a two-stage approach whereby multi-agent interactions are first determined, and their importance is then estimated to generate actions. In GraphComm \cite{yuan2021graphcomm}, the agents share their encoded observations over a multi-step communication process; at each step a GNN processes a graph and generates signals for the subsequent  communication round. This multi-round process is designed to increase the length of the communication mechanism and favour a longer range exchange of information.  The MAGIC algorithm \cite{niu2021multi} consists of a scheduled learning when to communicate and
whom to address messages to, and a message processor to process communication signals; both components have been implemented using GNNs and the entire architecture is learned end-to-end.

In our proposed model, the attention mechanism depends on how the encoded information exchanged amongst the agents flows within the graph; the graph topology itself depends on the encoded observations and the heat kernel is used as a topology-dependent feature to control the agent's communication. The process of encoding the observations, inferring the graph topology, and learning the attention mechanism are all coupled with the aim to learn an optimal policy.

\section{Connectivity-driven Communication} \label{sec:method}

\subsection{Problem setting} 

We consider Markov Games, partially observable extension of Markov decision processes \cite{littman1994markov} involving $N$ interacting agents.  We use $\mathcal{S}$ to denote the set of  environmental states; $ \mathcal{O}_i$ and $\mathcal{A}_i$ indicate the sets of all possible observations and actions for the $i^{th}$ agent, with $i \in {1, \dots N} $, respectively. The agent-specific (private) observations at time $t$ are denoted by $\bm{o}^t_i \in \mathcal{O}_i$, and each action $a^t_i \in \mathcal{A}_i$ is deterministically determined by a mapping, $ \bm{\mu}_{\theta_i}: \mathcal{O}_i \mapsto \mathcal{A}_i $, which is parametrised by $\theta_i$. 
A transition function $ \mathcal{T}: S \times \mathcal{A}_1 \times \mathcal{A}_2 \times \dots \times \mathcal{A}_N  $ describes the stochastic behaviour of the environment.
Each agent receives a reward, defined as a function of states and actions $ r_i : \mathcal{S} \times \mathcal{A}_1 \times \mathcal{A}_2 \times \dots \times \mathcal{A}_N  \mapsto \mathbb{R}$ and learns a policy that maximises the expected discounted future rewards over a period of $T$ time steps, $J(\theta_i)  = \mathbb{E} [R_i]$, where $R_i  =\sum_{t=0}^{T} \gamma^t r^t_i(s^t,a^t_1, \dots, a^t_N) $ is the discounted sum of future rewards, where  $ \gamma \in [0,1] $ is the discount factor.

\subsection{Learning the dynamic communication graph} \label{sec:topology}

We model each agent as the node of a time-depending, undirected (and unknown) weighted graph, $ G^t=(V, \bm{S}^t )$, where {\color{black}$V$} is a set of $N$ nodes and $ \bm{S}^t$ is an $N \times N$ matrix of edge weights. Each $\bm{S}^t(u,v) = \bm{S}^t(v,u) = s^t_{u,v} $ quantifies the degree of communication or connectivity strength between a given pair of agents, $u$ and $v$. Specifically, we assume that each $ s^t_{u,v} \in [0,1]$ with values close to 1 indicating strong connectivities, and to 0 a lack of connectivity.

%A zero weight, i.e. $ s_{u,v} = 0 $ indicates there two agents are not connected.

%Each element of $\bm{S}^t$, denoted by $ s_{u,v} $, quantifies the degree of connectivity between a pair of agents, $ u $ and $ v $. 

%All the edges an then arranged in a matrix $ \bm{S}_{u,v} = s_{u,v} $  with $\bm{S} \subseteq V \times V $. 

%A graph is a discrete structure, $ G = (\bm{E},V) $, where $ V $ represents a set of $ V $ nodes (or vertices) and $ \bm{E} \subseteq V \times V $ the set of edges. 
%In our approach, each node represents an agent. 

\begin{figure}
\centering
\includegraphics[width=1\linewidth]{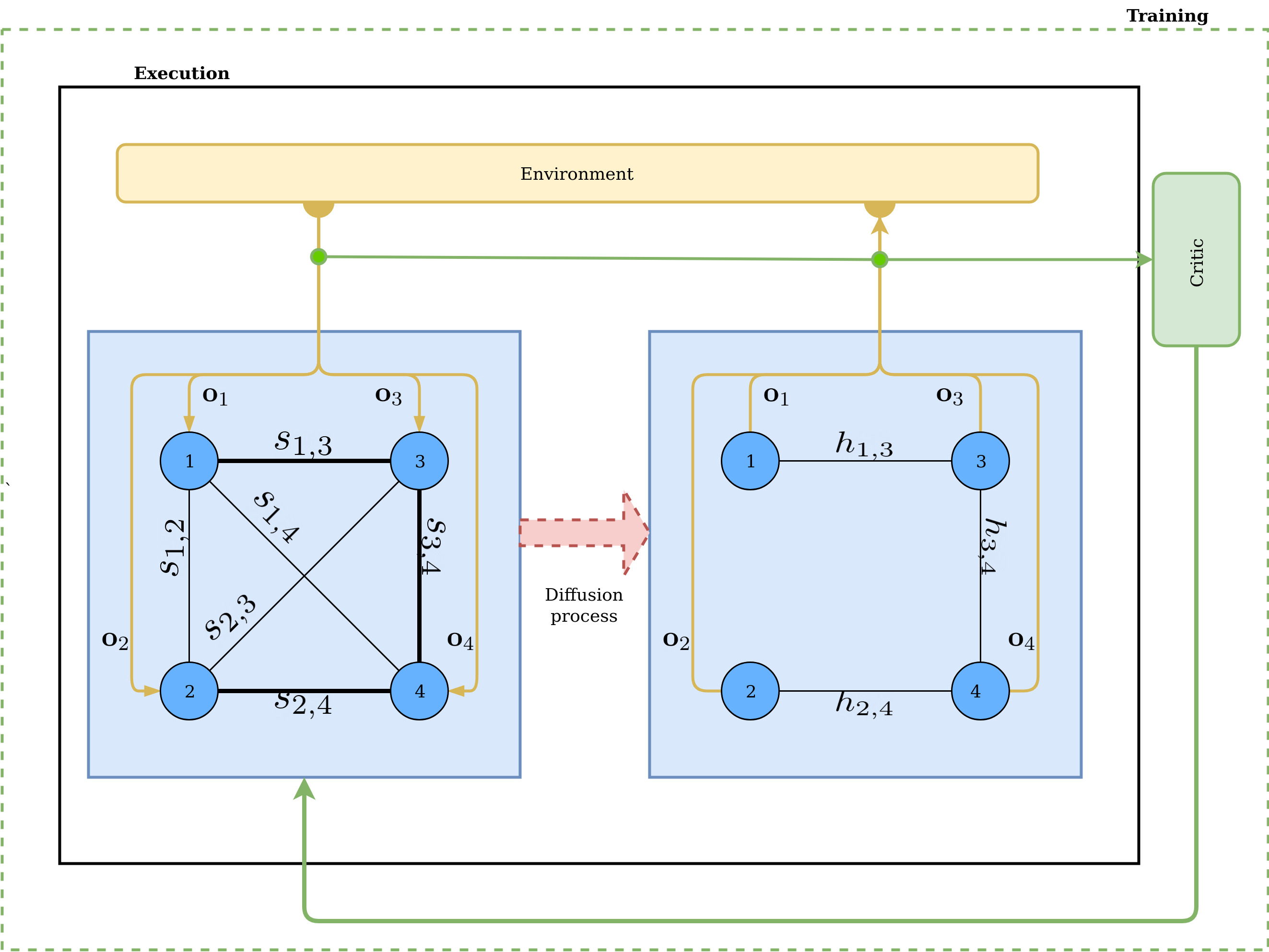}
\caption{ \textmd{Diagrammatic representation of CDC at a fixed time-step. Agents' observations are encoded to generate a graph topology (blue box on the left). The diffusion process is used to quantify global information flow throughout the graph and to control the communication process (blue box on the right). In this example, the line thickness is proportional to communication strength. At training time, observations and actions are utilised by the critic to receive feedback on the graph components.}}
\label{img:architecture}
\end{figure}

%These weights are not known \emph{a priori} and must be learnt. 

In our formulation, each $s^t_{u,v}$ is not known \emph{a priori}. Instead, each one of these connectivities is assumed to be a time-dependent parameter that varies as a function of the current state of the environment. This is done through the following two-step process. First, given a pair of agents, $u$ and $v$, their private observations at time-step $ t $ are  encoded to form a local message, 
\begin{equation}\label{eq:communicationtensor}
\bm{c}^t_{u,v}= \bm{c}^t_{v,u} = \varphi_{\theta^c}(\bm{o}^t_u,\bm{o}^t_v) 
\end{equation}
where $ \varphi_{\theta^{c}} $ is a non-linear mapping modelled as a neural network with parameter $\theta^{c}$. Each local message is then encoded non-linearly to produce the corresponding  connectivity weight,
\begin{equation}\label{eq:strengthmatrix}
s^t_{u,v} = s^t_{v,u}  = \sigma({\varphi_{\theta^s}(\bm{c}^t_{u,v})})
\end{equation}
where  $ \varphi_{\theta^{s}} $ is a neural network parameterised by $\theta^{s}$  and $ \sigma $ is the sigmoid function.

\subsection{{\color{black}Learning a time-dependent attention mechanism}} \label{sec:heat_kernel}

Once the time-dependent connectivities in Eq. \ref{eq:strengthmatrix} are estimated, the communication graph $G^t$ is fully specified. Given this graph, our aim is to characterise the relative contribution of each node to the overall flow of information over the entire network, and let these contributions define a attention mechanism controlling what messages are being exchanged. The resulting attention mechanism should be differentiable with respect to the network parameters to ensure that, during backpropagation, all the gradients correctly flow throughout the architecture to enable end-to-end training. 

%A message-passing strategy is introduced that takes into account the current network topology. 
Our observation is that a diffusion process over graphs can be deployed to  quantify how the information flows across all agents for any given communication graph, $G^t$. The information flowing process is conceptualised as the amount of energy that propagates throughout the network \cite{kondor2002diffusion}. Specifically, we deploy the heat diffusion process: we mimic the process of applying a source of heat over a network and observe how it varies as a function of time. In our context, the heat transfer patterns reflect how efficiently the information propagates at time $ t $.

First, we introduce a diagonal matrix $ \mathbf{D}(u)$ of dimension $N \times N$ with diagonal elements given by
$$
D(u,u) = \sum\limits_{v \in V}s_{u,v},~ \forall u \in V.
$$
Each such element provides a measure of strength of node $u$. The Laplacian of the communication graph $G$ is given by 
$$
\mathcal{L} = \bm{D} - \bm{S}
$$ 
and its normalised version is defined as  
%	\begin{equation} % \label{eq:normlaplac}
$$
\mathcal{\hat{L}} = \frac{1}{\sqrt{\bm{D}}}\mathcal{L} \frac{1}{\sqrt{\bm{D}}}.
$$
%	\end{equation} 
%where $ \mathcal{\hat{L}} \in \mathbb{R}^{V \times V}$. 
%The eigenspectrum formulation  of the normalized Laplacian is defined as 
%$$
%\mathcal{\hat{L}} = \bm{\phi} \Lambda \bm{\phi} ^{\intercal}
%$$ 
%in which $ \Lambda = diag(\lambda_1,\dots,\lambda_V) $ is a diagonal matrix formed by the eigenvalues of $ \mathbf{S} $ ordered by increasing magnitude, and $ \bm{\phi} = (\phi_1,\dots,\phi_V) $ is a matrix with the corresponding eigenvector as columns. 

The differential equation describing the heat diffusion process over time $p$ \cite{fiedler1989laplacian, chung1997spectral} is defined as 
\begin{equation}\label{eq:heatequation}
\frac{\partial H(p)}{\partial p} = - \mathcal{\hat{L}}H(p).
\end{equation}
where $ H(p)$ is the fundamental solution representing the energy flowing through the network at time $ p $. To avoid confusion, the environment time-step is denoted by $t$ whilst $p$ indicates the time variable related to the diffusion process. For each pair of nodes $ u $ and $ v $, the corresponding heat kernel entry is given by 
{\color{black}
\begin{equation}\label{eq:hknumerical}
H(p)_{u,v} =   \bm{\phi} ~ \text{exp}[\Lambda p]\bm{\phi}^{\intercal} = \sum\limits_{i = 1}^{N}\text{exp}[-\lambda_ip]\phi_i(u)\phi_i(v)
\end{equation}
}
%$H^t(p)_{u,v} =   \bm{\phi}^t \text{exp}[\Lambda^t p]\bm{\phi}^{t^\intercal} = \sum\limits_{i = 1}^{|V|}\text{exp}[-\lambda^t_ip]\phi^t_i(u)\phi^t_i(v)$
where $ H(p)_{u,v} $ quantifies the amount of heat that started in $ u $ {\color{black} and reached $ v $ at time $ p $, $ \phi_i $ represents the $ i^{th}$ eigenvector, $ \bm{\phi} = (\phi_1,\dots,\phi_N) $ is a matrix with the corresponding eigenvectors as columns and $ \Lambda = diag(\lambda_1,\dots,\lambda_V) $ is a diagonal matrix formed by the eigenvalues of $ \mathbf{S} $ ordered by increasing magnitude. }

In practice, Eq. \eqref{eq:hknumerical} is approximated using Pad\'e approximant \cite{al2009new},
%	\begin{equation}\label{eq:pade}
$$
H(p) = \text{exp}[-p \mathcal{\hat{L}}].
$$
%	\end{equation}
%\begin{equation}\label{eq:hk}
%H(p) = \text{exp}[-p \mathcal{\hat{L}}^t],
%\end{equation}

A useful property of $ H(p) $ is that it is differentiable with respect to neural network parameters that define the Laplacian. This allows us to train an architecture where all the relevant quantities are estimated end-to-end via backpropagation. Additional details are provided in Section \ref{sec:supp_heatkerneldetails}. 

%A hard requirement of the backpropagation algorithm is indeed that all the components involved in the training process are differentiable in order to allow the gradients to be propagated through the entire system. 

%We exploit this formulation to determine the most important messages within a framework, where the intra-agent message passing is optimized through reinforcement learning \textbf{not sure I understand this}. We compute the heat kernel by using Equations \ref{eq:hknumerical} and \ref{eq:pade} on the connectivity strength matrix $ \bm{S}^t $ to extract $ H(p)_{u,v} $ which is amount of heat that flowed from $ u $ and reached $ v $ \textbf{why do we need to mention $S$ here which does not appear in the equations?}.  The heat diffusion over the graph can also be interpreted as the potential energy transmitted over time by each edge, which provides an analogy for the propagation of information between nodes of the graph \cite{thanou2017learning}.  

%The fact that the heat kernel equations are differentiable is a key aspect which allows us to train our policy networks utilising a gradient based approach avoiding the flow of computation to be interrupted. This is a fundamental property exploited by the backpropagation algorithm, during the learning process, to calculate all the derivates necessary to correctly update the set of the learnable weights in order to maximise the expected reward function.

We leverage this information to develop an attention mechanism that identifies the most important messages within the system, given the current graph topology. First, for every pair of nodes, we identify the critical time point $\hat p$ at which the heat transfer drops by a pre-determined percentage $\delta$ and becomes stable, i.e. for each pair of $u$ and $v$, we identify that critical value $\hat p(u,v)$ such that  
\begin{equation}\label{eq:threshold}
\Big\lvert\frac{ H^t(p+1)_{u,v} - H^t(p)_{u,v}  }{H^t(p)_{u,v}} \Big\lvert < \delta.
%p_{max} := \argmax_p \Big\lvert\frac{ H^t(p+1)_{u,v} - H^t(p)_{u,v}  }{H^t(p)_{u,v}} \Big\lvert < \delta.
\end{equation}
In practice, the search of these critical values is carried out over a uniform grid of points. Once these critical time points are identified, we use them to evaluate the HK values, and arrange them into an $N \times N$ matrix,  
$$
H^t_{u,v} = H^t (\hat p(u,v))
$$ 
which is used to define a multi-agent message-passing mechanism. Specifically, the final information content (or message) for an agent $u$ is determined by a linear combination of the local messages received from all other agents, 
\begin{equation}\label{eq:message}
\bm{m}^t_u = \sum\limits_{v \in V} H^t_{u,v}\bm{c}^t_{u,v}
\end{equation}
where the HK values are used to weight the importance of the incoming messages. Finally, the agent's action depends deterministically by its message,
\begin{equation}\label{eq:actionselector}
a^t_u  = \varphi_{\theta^p_u} (\bm{m}^t_u) 
\end{equation}
where $\varphi_{\theta^p_u} $ is a neural network with parameters $ \theta^p_u $. A lack of communication between a pair of agents results when no stable HK values can be found. In such cases, for a pair of agents $(u,v)$, the corresponding entry in $ H^t_{u,v}$ will be zero hence no value of $ \hat p(u,v) $ satisfies Eq. \ref{eq:threshold}.

%Once each agent has taken an action, a reward is provided by the environment, 
%$$
%r^t_i : \mathcal{S} \times \mathcal{A}_1 \times \mathcal{A}_2 \times \dots \times \mathcal{A}_N  \mapsto \mathbb{R}.
%$$
%Each agent objective's is to maximise the discounted sum of future rewards over time, 
%\small
%\begin{equation}\label{eq:objective}
%J(\theta_i)  = \mathbb{E}_{a^t_1 \sim \bm{\mu}_1,\dots,a^t_N \sim \bm{\mu}_N, s^t \sim \mathcal{T}} \Bigg[ \sum_{t=0}^{T} \gamma^t r^t_i(s^t,a^t_1, \dots, a^t_N) \Bigg]
%\end{equation}
%\normalsize
%where $\gamma \in [0,1]$ is the discount factor.  

{\color{black}

\subsection{Heat kernel: additional details and an illustration} \label{sec:supp_heatkerneldetails}

The heat kernel is a technique from spectral geometry  \cite{differentialgeometry}, and is a fundamental solution of the \emph{heat equation}:
\begin{equation}\label{eq:supp_heat}
\frac{\partial H^t(p)}{\partial p} = - \mathcal{\hat{L}}^tH^t(p).
\end{equation}
Given a graph $ G $ defined on $ n $ vertices, the normalized Laplacian $ \mathcal{\hat{L}} $, acting on functions with Neumann boundary conditions \cite{cheng2005heritage}, is associated with the rate of heat dissipation. $ \mathcal{\hat{L}} $ can be written as
%		\begin{equation}
$$
\mathcal{\hat{L}} = \sum\limits_{i = 0}^{n-1} \lambda_i I_i
$$
%		\end{equation}
where $ I_i $ is the projection onto the $ i^{th} $ eigenfunction $ \phi_i $. For a given time $ p \geq 0 $, the heat kernel $ H(p) $ is defined as a $ n \times n $ matrix:
\begin{equation}\label{eq:supp_hksol}
H(p) = \sum\limits_{i}\text{exp}[-\lambda_ip]I_i = \text{exp}[-p\mathcal{\hat{L}}].
\end{equation}

Eq. \ref{eq:supp_hksol} represents an analytical solution to Eq. \ref{eq:supp_heat}. Furthermore, the heat kernel H(t) for a graph $ G $ with eigenfunctions $ \theta_i $ satisfies
$$
H(p)_{u,v} =   \sum\limits_{i = 1} \text{exp} [-\lambda_ip]\phi_i(u)\phi_i(v).
$$
The proof follows from the fact that
$$
H(p) =   \sum\limits_{i}\text{exp}[-\lambda_ip]I_i
$$
and 
$$
I(u,v) = \phi_i(u)\phi_i(v).
$$

}

%\vspace{1cm}
%\begin{lemma}\label{lemma1} \cite{chung1997spectral}  
%\end{lemma}

{\color{black}
%\subsection{Edge cutting example}\label{sec:supp_varna}
In this work the heat kernel is used to introduce a mechanism for the selection of important edges in a network to support communication between nodes. In this context, the importance of an edge is determined by both its weight and the role it plays to allow agents to exchange information correctly in the network structure.  Figure \ref{fig:example_graph_cutting} illustrates the advantages of selecting edges through the heat kernel features over a naive thresholding approach. The heat diffusion considers the edge weights as well as their relevance within the graph structure, e.g. edge connecting two communities.

\begin{figure}[h]
	\begin{minipage}[b]{0.33\linewidth}
		\centering
		\includegraphics[width=1\linewidth]{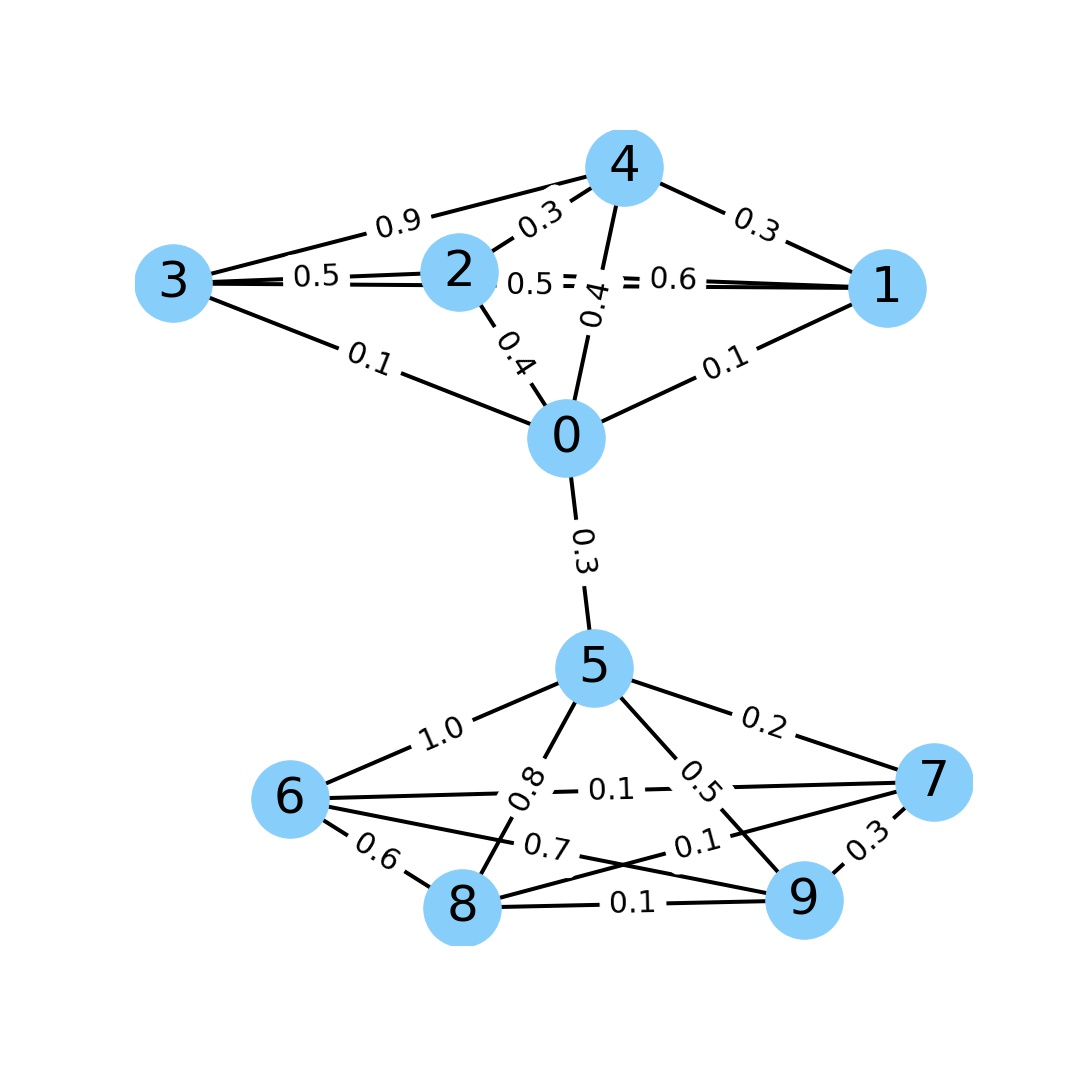} 
		\caption*{\textmd{(a)}}
	\end{minipage} 
	\begin{minipage}[b]{0.32\linewidth}
		\centering
		\includegraphics[width=1\linewidth]{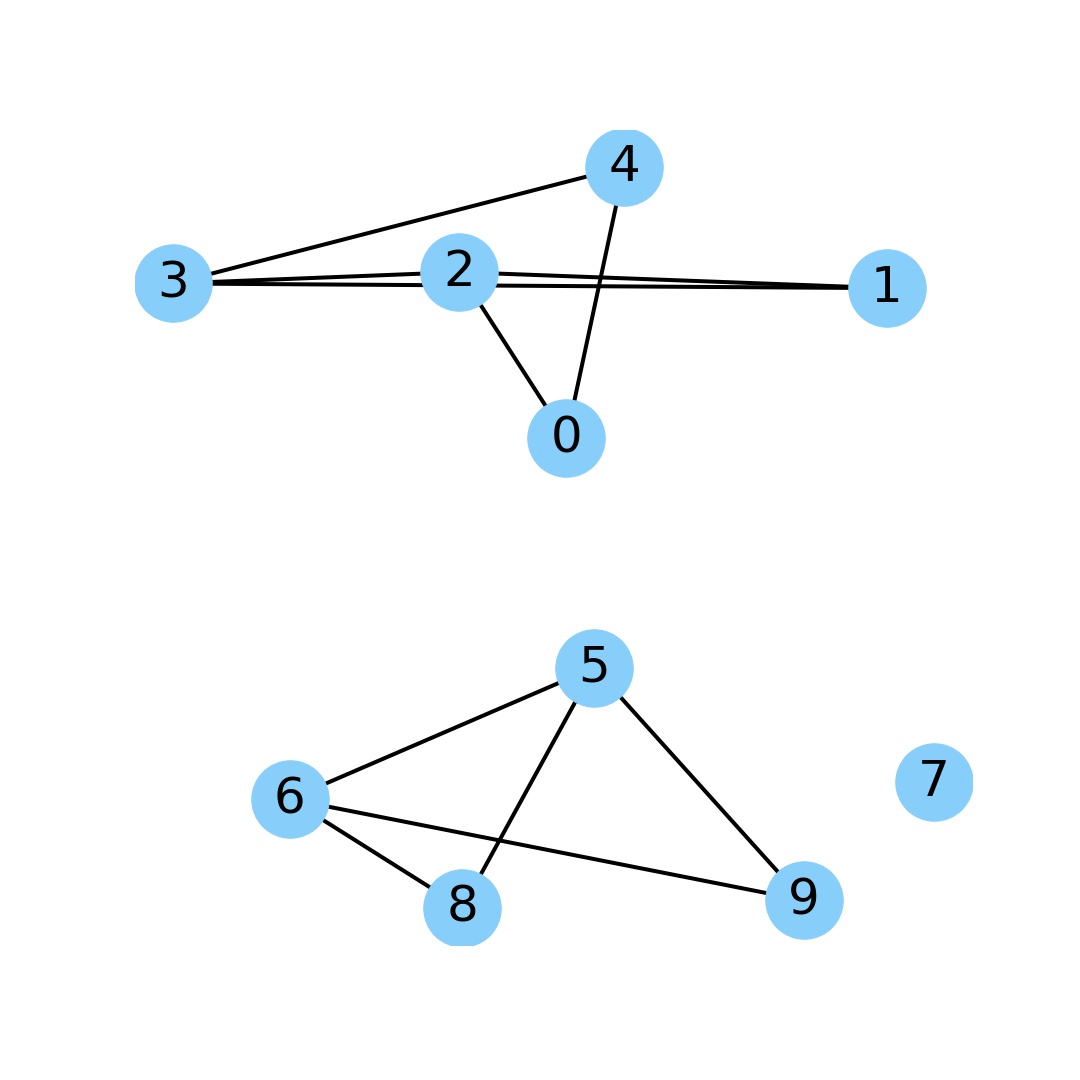} 
		\caption*{\textmd{(b)}}
	\end{minipage} 
	\begin{minipage}[b]{0.32\linewidth}
		\centering
		\includegraphics[width=1\linewidth]{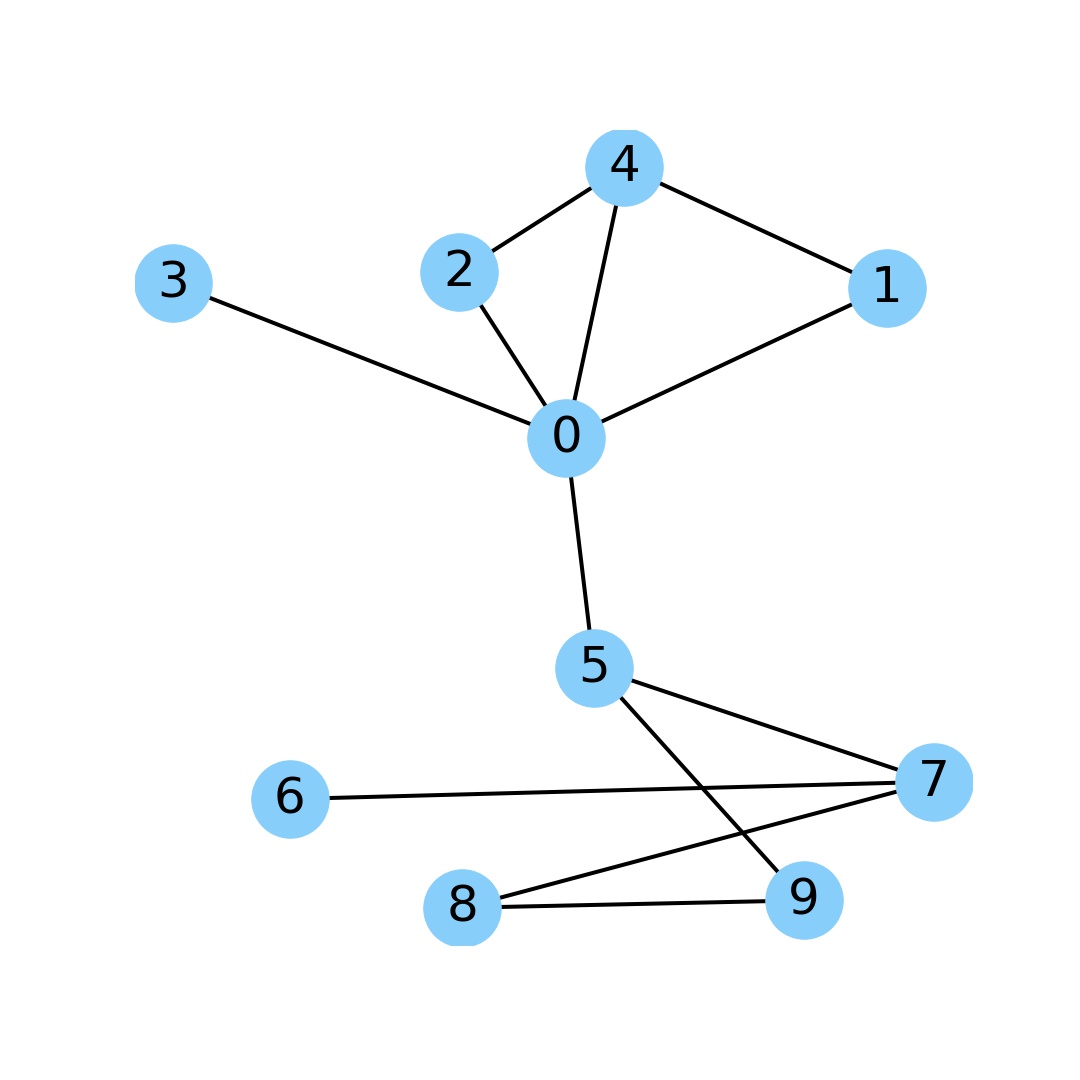} 
		\caption*{\textmd{(c)}}
	\end{minipage} 
	\caption{\textmd{{\color{black}An illustration of two edge selection methods. Starting from graph (a), we want to remove the less relevant edges. The relevance of an edge is measured considering both its weight and structural role in allowing information to pass through the network. The edge connecting nodes 0 and 5, despite its relatively low weight (0.3), has an important structural role as it serves as bridge connecting two communities hence allowing the information to propagate throughout the entire network. In (b), removing edges with smaller weights (e.g. all those falling below the 40th percentile of the edge weight distribution) results in the loss of the bridge. In (c), edges are selected based on the heat kernel weights, which recognise the importance of the bridge.}}}
	\label{fig:example_graph_cutting} 
\end{figure}

} % end color_red

\subsection{Reinforcement learning algorithm}\label{sec:learning}

In this section, we describe how the reinforcement learning algorithm is trained in an end-to-end fashion. We extend the actor-critic framework \cite{degris2012off} in which an actor produces actions and a critic provides feedback on the actors' moves. In our architecture, multiple actors, one per each agent, receive feedback from a single, centralised critic. 

In the standard DDPG algorithm \cite{silver2014deterministic,lillicrapHPHETS15}, the actor $\bm{\mu}_{\theta} : \mathcal{O}  \mapsto \mathcal{A}$ and the critic $ Q^{\bm{\mu}_{\theta}}: \mathcal{O} \times \mathcal{A} \mapsto \mathbb{R} $  are parametrised by neural networks with the aim to maximize the expected return,  
$$
J(\theta)  = \mathbb{E} \Big[\sum_{i = 1}^{T}r(\bm{o^t},a^t)\Big].
$$ 

where $\theta$ is the set of parameters that characterise the return. The gradient $ \nabla_\theta J(\theta) $ required to update the parameter vector $ \theta $ is calculated as follows,
$$
\nabla_{\theta} J(\theta) = \mathbb{E}_{\bm{o}^t \sim \mathcal{D}} \big[ \nabla_{\theta} \bm\mu_{\theta}(\bm{o}^t) \nabla_{a^t} Q^{\bm\mu_{\theta}}(\bm{o}^t,a^t) \lvert_{a^t=\bm\mu_\theta(\bm{o}^t)} \big] .
$$
whilst $ Q^{\bm{\mu}_{\theta}} $ is obtained by minimizing the following loss,
\small
%\begin{equation}
%\begin{split}
%L(\theta) &= \mathbb{E}_{{\bm{o}^t, a^t, r^t, \bm{o}^{t+1}} \sim \mathcal{D}} \Big[\big(Q^{\bm{\mu}_{\theta}}(\bm{o}^t, a^t) - y \big)^2\Big], \\
%%\text{where  }  y &= r^t + \gamma Q^{\bm{\mu'}_{\theta}}(\bm{o}^{t+1}, a^{t+1})
%\end{split}
%\end{equation}
%\begin{equation}
%L(\theta) = \mathbb{E}_{{\bm{o}^t, a^t, r^t, \bm{o}^{t+1}} \sim \mathcal{D}} \Big[\big(Q^{\bm{\mu}_{\theta}}(\bm{o}^t, a^t) - r^t - \gamma Q^{\bm{\mu'}_{\theta}}(\bm{o}^{t+1}, a^{t+1})\big)^2 \Big]
%\end{equation}
$$
L(\theta) = \mathbb{E}_{{\bm{o}^t, a^t, r^t, \bm{o}^{t+1}} \sim \mathcal{D}} \Big[\big(Q^{\bm{\mu}_{\theta}}(\bm{o}^t, a^t) - y \big)^2\Big]
$$
\normalsize
where 
$$
y = r^t + \gamma Q^{\bm{\mu'}_{\theta}}(\bm{o}^{t+1}, a^{t+1}).
$$
Here, $Q^{\bm{\mu'}_{\theta}}$ is a target critic whose parameters are only periodically updated with the parameters of $Q^{\bm{\mu}_{\theta}}$, which is utilised to stabilize the training.

Our developments follow the CLDE paradigm \cite{kraemer2016multi,foerster2016learning,lowe2017multi}. The critics are employed during learning, but otherwise only the actor and communication modules are used at test time. At training time, a centralised critic uses the observations and actions of all the agents to produce the $ Q $ values. In order to make the critic unique for all the agents and keep the number of parameters constant, we approximate our $ Q $ function with a recurrent neural network (RNN). We treat the observation/action pairs as a sequence,
\begin{equation}\label{eq:rnncritic}
\bm{z}^t_i = \text{RNN}(\bm{o}^t_i,a^t_i \lvert \bm{z}^t_{i-1}) 
\end{equation}	
where $ \mathbf{z}^t_i $ and $ \bm{z}^t_{i-1} $ are the hidden state produced for the $ i^{th} $ and $  i-1^{th} $ agent, respectively. 
Upon all the observation and action pairs from all the $N$ agents are available, we use the last hidden state $ \bm{z}^t_N $ to produce the $ Q $-value:
$$
Q(\bm{o}^t_1,\dots,\bm{o}^t_N,a^t_1, \dots,a^t_N) = \varphi_{\theta_Q} (\bm{z}^t_N)
$$		
where $ \varphi $ is a neural network with parameters $ \theta_Q $. The parameters of the $ i^{th} $ agent are adjusted to maximize the objective function $J(\theta_i)  = \mathbb{E} [R_i]$ following the direction of the gradient $J(\theta_i)$,
\begin{equation} \label{eq:gradienttheta}
\nabla_{\theta_i} J(\theta_i) = \mathbb{E}_{\bm{o}^t_i, a^t_i, r^t, \bm{o}^{t+1}_i \sim \mathcal{D}} \big[ \nabla_{\theta_i} \bm{\mu}_{\theta_i}(\bm{m}^t_i) \nabla_{a^t_i} Q(\bm{x}) \lvert_{a^t_i=\bm{\mu}_{\theta_i}(\bm{m}^t_i)} \big] 
\end{equation}	
where $ \bm{x} = (\bm{o}^t_1,\dots,\bm{o}^t_N,a^t_1, \dots,a^t_N)$ and  $ Q $ minimizes the temporal difference error, i.e. 
$$
L(\theta_i) = \mathbb{E}_{\bm{o}^t_i, a^t_i, r^t, \bm{o}^{t+1}_i \sim \mathcal{D}} \Big[(Q(\bm{x}) - y)^2 \Big]
$$	
where
$$
y = {}   r^t_i + \gamma Q(\bm{o}^{t+1}_1,\dots,\bm{o}^{t+1}_N,a^{t+1}_1, \dots,a^{t+1}_N). 
$$	

The differentiability of the heat kernel operator allows the gradient in Eq. \eqref{eq:gradienttheta} to be evaluated. Since the actions are modelled by a neural network parametrised $\theta_u$ in Eq.\eqref{eq:actionselector}, we have that
$$
\nabla_{\theta_u} \bm{\mu}_{\theta_u}(\bm{m}^t_u) =   \nabla_{\theta_u} \varphi_{\theta_u} (\bm{m}^t_u).
$$
and from Eq.\eqref{eq:message} the gradient is
\begin{align*}
\frac{\partial \varphi(\bm{m}^t_u)}{\partial{\theta_u}}   &=  \frac{\partial  \varphi \Big( \sum\limits_{v \in V} H^t_{u,v}\bm{c}^t_{u,v} \Big)}{\partial  \varphi_{\theta_u}} \\
&=  \sum\limits_{v \in V}  \frac{\partial  \varphi ( H^t_{u,v} \bm{c}^t_{u,v} )}{\partial  \varphi_{\theta_u}} \\
&=  \sum\limits_{v \in V} \Big( \frac{\partial  \varphi  (H^t_{u,v}) }{\partial  \varphi_{\theta_u}} \bm{c}^t_{u,v} + H^t_{u,v} \frac{\partial  \varphi   (\bm{c}^t_{u,v}) }{\partial  \varphi_{\theta_u}} \Big).
\end{align*}

whilst the gradients of the HK values are
\begin{align*}
\frac{\partial  \varphi  (H^t_{u,v}) }{\partial  \varphi_{\theta_u}}  &= \frac{\partial  \varphi  ({H}^t_{u,v}(\hat p)) }{\partial  \varphi_{\theta_u}}  \\
&=   \frac{\partial (\text{exp}[-\hat p \mathcal{\hat{L}}]  )}{\partial  \varphi_{\theta_u}} \\
&=  \frac{\partial (\text{exp}[-\hat p \frac{1}{\sqrt{\bm{D}}}\mathcal{L} \frac{1}{\sqrt{\bm{D}}}]  )}{\partial  \varphi_{\theta_u}}  \\ 
&=   \frac{\partial (\text{exp}[-\hat p \frac{1}{\sqrt{\bm{D}}} (\bm{D} - \bm{S}) \frac{1}{\sqrt{\bm{D}}}]  )}{\partial  \varphi_{\theta_u}}
\end{align*}
which is a composition of differentiable operations. {\color{black}Algorithm \ref{alg:mainalgo} summarises the learning algorithm; the proposed architecture is presented in Figure \ref{img:architecture}.}

\begin{algorithm}[h!]
\caption{CDC}\label{alg:mainalgo}
\small
{\color{black}
	\begin{algorithmic}[1]
		\State Inizialise actor ($ \bm{\mu}_{\theta_1}, \dots,  \bm{\mu}_{\theta_N} $) and critic networks ($ Q_{\theta_1}, \dots, Q_{\theta_N} $)
		\State Inizialise actor target networks ($ \bm{\mu'}_{\theta_1}, \dots,  \bm{\mu'}_{\theta_N} $) and critic target networks ($ Q'_{\theta_1}, \dots, Q'_{\theta_N} $)
		\State Inizialise replay buffer $ \mathcal{D} $
		\For{episode = 1 to E}	
		\State Reset environment, $ \bm{o}^1 = \bm{o}^{1}_1, \dots, \bm{o}^{1}_N $
		\For{t = 1 to T}
		%				\State Every $ i^{th} $ agent receives its observation $ \bm{o}_i^t $ 
		\State Generate $ \bm{C}^t $ (Eq. \ref{eq:communicationtensor}) and $ \bm{S}^t $ (Eq. \ref{eq:strengthmatrix})
		\For{{\color{black}$p$} = 1 to $ P $}
		\State Compute Heat Kernel $ H(p)^t $ (Eq. \ref{eq:heatequation})
		\EndFor 	
		\State Build $ \bm{H}^t $ with stable Heat Kernel values (Eq. \ref{eq:threshold} )
		\For{agent i = 1 to N}
		\State Produce agent's message $ \bm{m}^t_i $ (Eq. \ref{eq:message})
		\State Select action $ a^t_i = \bm{\mu}_{\theta_i}(\bm{m}^t_i) $
		\EndFor
		%				\State Execute $ \bm{a}^t=(a^t_1, \dots, a^t_N) $, observe $ r $ and $(\bm{o}^{t+1}, \dots, \bm{o}^{t+1}_N) $
		\State Execute $ \bm{a}^t=(a^t_1, \dots, a^t_N) $, observe $ r $ and $\bm{o}^{t+1}$
		\State Store transaction $(\bm{o}^{t},\bm{a}^t,r,\bm{o}^{t+1})$ in $ \mathcal{D} $		
		\EndFor % For t
		\For{agent i = 1 to $ N $}
		\State Sample minibatch $ \Theta $ of $ B $ transactions $(\bm{o},\bm{a},r,\bm{o'})$
		\State Update critic by minimizing:\\
		\State $ L(\theta_i) = \frac{1}{B} \sum_{( \bm{o},\bm{a},r,\bm{o'}) \in  \Theta} (y -Q(\bm{o}, \bm{a}))^2$,
		\State where $ y = r_i + \gamma Q(\bm{o}', \bm{a}') \lvert_{a'_k =  \bm{\mu'}_{\theta_k}(\bm{m}'_k) }  $
		%			\State in which $ \bm{m}'_k $ is computed through Equations \ref{eq:communicationtensor},\ref{eq:strengthmatrix}, \ref{eq:heatequation},\ref{eq:threshold}, \ref{eq:message} 
		\State in which $ \bm{m}'_k $ is global message computed using target networks
		\State Update actor according to the policy gradient:
		\small
		%			\State $\nabla_{\theta_i} J \approx \frac{1}{B} \sum_{( \bm{o},\mathbf{a},r,\bm{o'}) \in \Theta}  \Big(\nabla_{\theta_i}\bm{\mu}_{\theta_i}(\bm{m}_i) \nabla_{a_i}Q^{\bm{\mu}_{\theta_i}}(\bm{o}, \bm{a})|_{a_i=\bm{\mu}_{\theta_i}(\bm{m}_i)} \Big) 	$
		\State \hspace{-1cm} $\nabla_{\theta_i} J \approx \frac{1}{B} \sum  \Big(\nabla_{\theta_i}\bm{\mu}_{\theta_i}(\bm{m}_i) \nabla_{a_i}Q^{\bm{\mu}_{\theta_i}}(\bm{o}, \bm{a})\lvert_{a_i=\bm{\mu}_{\theta_i}(\bm{m}_i)} \Big) 	$							
		\EndFor			
		\State Update target networks:
		\State $\theta^{'}_i = \tau \theta_i + (1-\tau)\theta^{'}_i$		
		\EndFor
	\end{algorithmic}
}
\end{algorithm}	
\normalsize

%\begin{equation}
% 	\frac{\partial \varphi(\bm{m}^t_u)}{\partial{\theta_u}}   =  \frac{\partial  \varphi \Big( \sum\limits_{v \in V} \bm{H}^t_{u,v}\bm{c}^t_{u,v} \Big)}{\partial  \varphi_{\theta_u}} 
%\end{equation}
%
%\begin{equation}
%\frac{\partial  \varphi \Big( \sum\limits_{v \in V} \bm{H}^t_{u,v}\bm{c}^t_{u,v} \Big)}{\partial  \varphi_{\theta^p_u}}   = \sum\limits_{v \in V}  \frac{\partial  \varphi \Big( \bm{H}^t_{u,v} \bm{c}^t_{u,v} \Big)}{\partial  \varphi_{\theta^p_u}}
%\end{equation}
%
%\begin{equation}
%\sum\limits_{v \in V}  \frac{\partial  \varphi \Big( \bm{H}^t_{u,v} \bm{c}^t_{u,v} \Big)}{\partial  \varphi_{\theta^p_u}}  = \sum\limits_{v \in V}  \frac{\partial  \varphi \Big( \bm{H}^t_{u,v}  \Big)}{\partial  \varphi_{\theta^p_u}} \bm{c}^t_{u,v} + \bm{H}^t_{u,v} \frac{\partial  \varphi \Big(    \bm{c}^t_{u,v} \Big)}{\partial  \varphi_{\theta^p_u}}
%\end{equation}

%\begin{equation}
% \frac{\partial  \varphi ( \bm{H}^t_{u,v}  )}{\partial  \varphi_{\theta^p_u}} = \frac{\partial (\text{exp}[-p \mathcal{\hat{L}}]  )}{\partial  \varphi_{\theta^p_u}}  
%\end{equation}
%
%\begin{equation}
%\frac{\partial  \varphi ( \bm{H}^t_{u,v}  )}{\partial  \varphi_{\theta^p_u}} = \frac{\partial (\text{exp}[-p \frac{1}{\sqrt{\bm{D}}}\mathcal{L} \frac{1}{\sqrt{\bm{D}}}]  )}{\partial  \varphi_{\theta^p_u}}  
%\end{equation}

%where it can be noted that the centralised critic (green) receiving observations and actions to provide feedback to agents is used only during training, and that the graph-driven communication process is also a key component of the proposed model during execution.

\section{Experimental results} \label{sec:experiments}

\subsection{Environments}\label{subsec:environments}

The performance of CDC has been assessed in four different environments. Three of them are commonly used swarm robotic benchmarks: \emph{Navigation Control}, \emph{Formation Control} and \emph{Line Control} \cite{mesbahi2010graph,balch1998behavior,agarwal2019learning}. A fourth one, \emph{Pack Control}, has been added to study a more challenging task. All the environments have been tested using the Multi-Agent Particle Environment \cite{lowe2017multi,mordatch2017emergence}, which allows agents to move around in two-dimensional spaces with discretised action spaces.
{\color{black} In \emph{Navigation Control} there are $ N $ agents and $ N $ fixed landmarks. The agents must move closer to all landmarks whilst avoiding collisions. Landmarks are not assigned to particular agents, and the agents are rewarded for minimizing the distances between their positions and the landmarks' positions.
Each agent can observe the position of all the landmarks and other agents. In \emph{Formation Control} there are $ N $ agents and only one landmark. In this scenario, the agents must navigate in order to form a polygonal geometric shape, whose shape is defined by the $N$ agents, and centred around the landmark. The agents' objective is to minimize the distances between their locations and the positions required to form the expected shape. Each agent can observe the landmark only. \emph{Line Control} is very similar to \emph{Formation Control} with the difference that the agents must navigate in order to position themselves along the straight line connecting the two landmarks. Finally in \emph{Dynamic Pack Control}
there are $ N $ agents, of which two are leaders and $ N-2$ are members, and one landmark. The objective of this task is to simulate a pack behaviour, where agents have to navigate to reach the landmark. Once a landmark is occupied, it moves to a different location. The landmark location is accessible only to the leaders, while the members are blind, i.e. they can only see their current location. Typical agent configurations arising from each environment we use here are reported in Figure \ref{fig:environments}.

\begin{figure}[H] 
	\centering
	\begin{minipage}[b]{0.40\linewidth}
		\centering
		\includegraphics[width=0.9\linewidth]{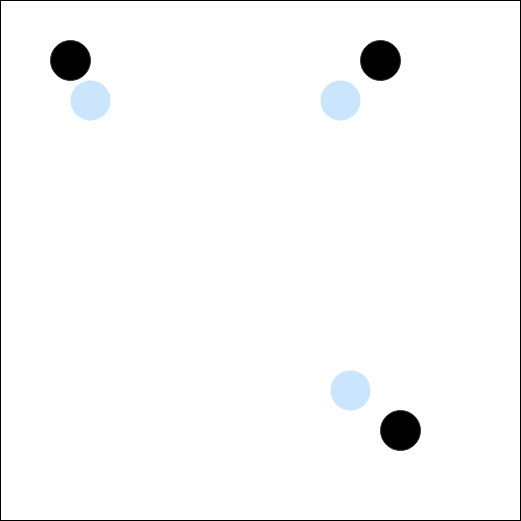} 
		\vspace{-1mm}
		\caption*{\textmd{(a) Navigation Control $ N = 3 $}}
	\end{minipage}%% 
	\begin{minipage}[b]{0.40\linewidth}
		\centering
		\includegraphics[width=0.9\linewidth]{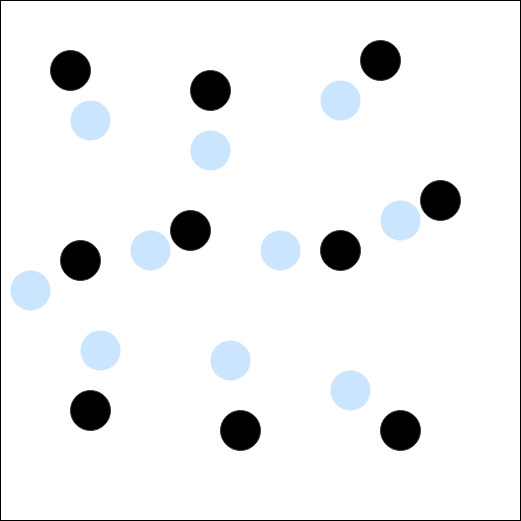} 
		\vspace{-1mm}
		\caption*{\textmd{(b) Navigation Control $ N = 10 $}}
	\end{minipage}%% 
	
	%%%%%%%%%%%%%%%%%%%%%%%
	
	\begin{minipage}[b]{0.40 \linewidth}
		\centering
		\includegraphics[width=0.9\linewidth]{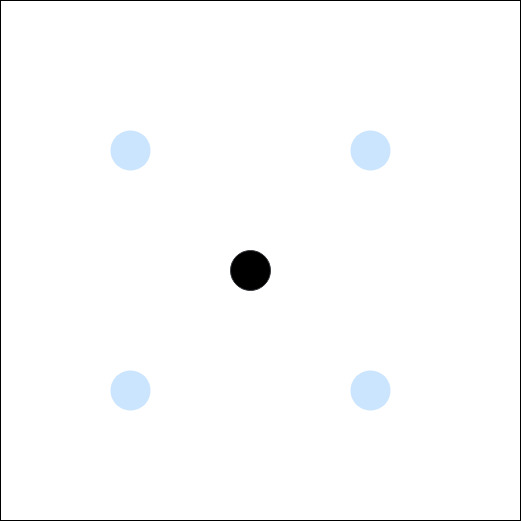} 
		\vspace{-1mm}
		\caption*{\textmd{(c) Formation Control $ N = 4 $}}
	\end{minipage} 
	\begin{minipage}[b]{0.40 \linewidth}
		\centering
		\includegraphics[width=0.9\linewidth]{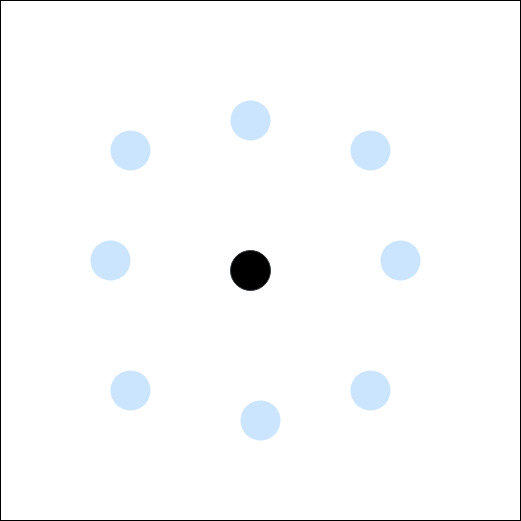} 
		\vspace{-1mm}
		\caption*{\textmd{(d) Formation Control $ N = 10 $}}
	\end{minipage}
	\vspace{1mm}

	%%%%%%%%%%%%%%%%%%%%%%%
	\begin{minipage}[b]{0.40\linewidth}
		\centering
		\includegraphics[width=.9\linewidth]{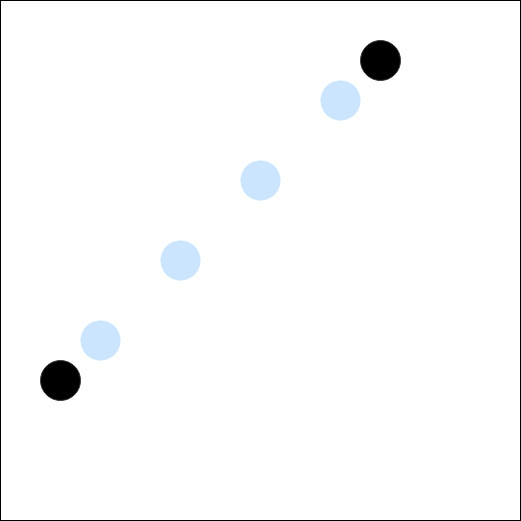} 
		\vspace{-1mm}
		\caption*{\textmd{(e) Line Control $ N = 4 $}}
	\end{minipage}%% 
	\begin{minipage}[b]{0.40\linewidth}
		\centering
		\includegraphics[width=.9\linewidth]{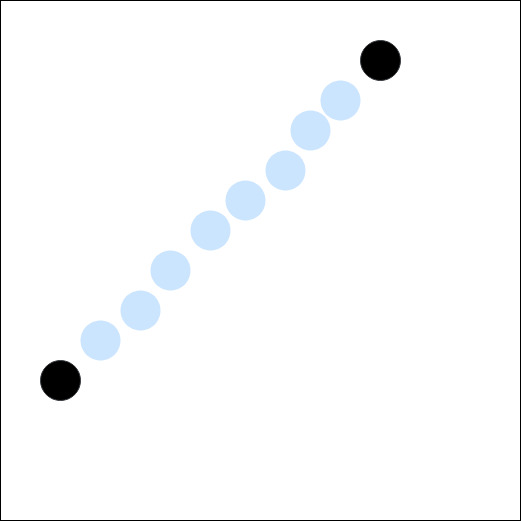} 
		\vspace{-1mm}
		\caption*{\textmd{(f) Line Control $ N = 10 $}}
	\end{minipage}%% 
	
	\vspace{1mm}
	
	%%%%%%%%%%%%%%%%%%%%%%%
	\begin{minipage}[b]{0.40 \linewidth}
		\centering
		\includegraphics[width=.9\linewidth]{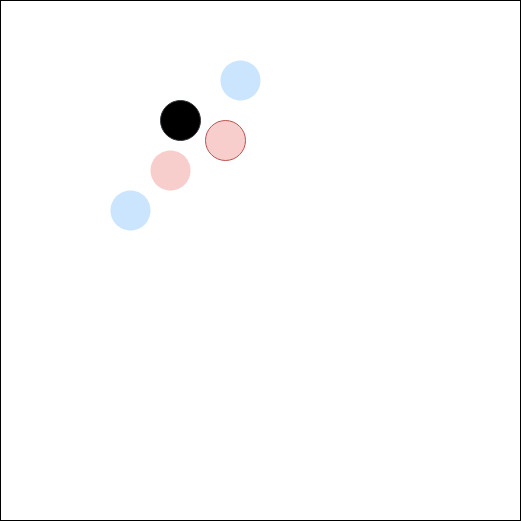} 
		\vspace{-1mm}
		\caption*{\textmd{(g) Pack Control $ N=4 $}}    
	\end{minipage} 
	\begin{minipage}[b]{00.40 \linewidth}
		\centering
		\includegraphics[width=.9\linewidth]{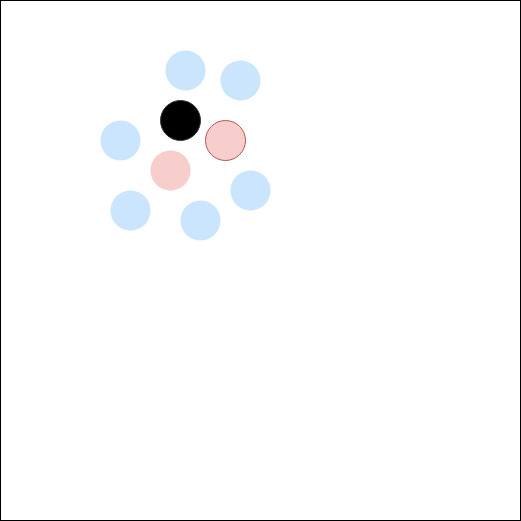} 
		\vspace{-1mm}
		\caption*{\textmd{(h) Pack Control $ N=8 $}}
	\end{minipage} 
	%%%%%%%%%%%%%%%%%%%%%%%	
	\caption{\textmd{Typical agent configurations for all our environments.}}\label{fig:environments}
	% Black shperes stand for landmarks. Blue shperes correspond to agents in (a), (b) and (c). In Pack Control (d) the red spheres indicates the leader agents and the blue ones the members.}}
	%	\label{fig:environments} 
\end{figure}
} % end color_red
%A brieft description of each environment is in order (more details are given in Supplementary Material, Section \ref{sec:supp_envs}). 
%\paragraph*{Navigation Control.}
%There are $ N $ agents and $ N $ fixed landmarks. The agents must move closer to all landmarks whilst avoiding collisions.
%\paragraph*{Formation Control.} There are $ N $ agents and only one landmark. The agents must navigate in order to form a polygonal geometric shape centred around the landmark.
%\paragraph*{Line Control.} There are $ N $ agents and two landmarks. The agents must navigate in order to position themselves along the straight line connecting the two landmarks. 
%\paragraph*{Dynamic Pack Control.} There are $ N $ agents, of which two are leaders and $ N-2$ are members, and one landmark, which is visible to leaders only. All agents needs to reach the landmark, that, once occupied, moves to a different location. 

%\vspace{1mm}

For each environment we have tested two versions with different number of agents: a {\it basic} one focusing on solving the designed task when $3-4$ agents are involved, and a {\it scalable} one to show the ability to succeed with $8-10$ agents. The performance of competing MADRL algorithms has been assessed using a number of metrics: the \emph{reward}, which quantifies how well a task has been solved (the higher the better); the \emph{distance}, which indicates the amount of navigation carried out by the agents to solve the task (the lower the better); the number of \emph{collisions}, which shows the ability to avoid collisions (the lower the better); the \emph{time} required to solve the task (the lower the better); the \emph{success rate}, defined as the number of times an algorithm has solved a task over the total number of attempts; and \emph{caught targets}, which refers to the number of landmarks that the pack managed to reach. Illustrative videos showing CDC in action on the above environments can be found online \footnote{\url{https://youtu.be/H9kMtrnvRCQ}}.

\subsection{Implementation details and experimental setup}

For our experiments, we use neural networks with two hidden layers ($ 64 $ each) to implement the graph generation modules (Eq. \ref{eq:strengthmatrix}, \ref{eq:communicationtensor}) and the action selector in Eq. \ref{eq:actionselector}. The RNN described in Equation \ref{eq:rnncritic} is implemented as a long-short term memory (LSTM) network \cite{schmidhuber1996general} with $ 64 $ units for the hidden state.

We use the Adam optimizer \cite{kingma2014adam} with a learning rate of $ 10^{-3} $ for critic and $ 10^{-4} $ for policies. 
Similarly to \cite{agarwal2019learning,wang2019stmarl}, we set $ \theta_1 = \theta_2 = \dots = \theta_N $ in order to make the model invariant to the number of agents.
The reward discount factor is set to $ 0.95 $, the size of the replay buffer to $ 10^{6} $, and the batch size to $ 1,024 $. At each iteration, we calculate the heat kernel over a finite grid of $P=300$ time points, with a threshold for getting stable values set to $s=0.05$. This value has been determined experimentally (see Table \ref{tab:supp_thresholdresults}).
The number of time steps for episode, $ T $, is set to $50$ for all the environments, except for Navigation Control where is set to $ 25 $. For Formation Control, Line Control and Pack Control the number $ E $ of episodes is set to is set to $ 50,000 $ for the basic versions ($ 30,000$ for scalable versions), while for Navigation Control is set to $ 100,000 $ ($ 30,000 $ for scalable versions).

All network parameters are updated every time $100$ new samples are added to the replay buffer. Soft updates with target networks use $ \tau = 0.01 $. 
We adopt the low-variance gradient estimator Gumbel-Softmax for discrete actions in order to allow the back-propagation to work properly with categorical variable, which can truncate the gradient's flow. 
All the presented results are produced by running every experiment $5$ times with different seeds ($ 1 $,$ 2001 $,$ 4001 $,$ 6001 $,$ 8001 $) in order to avoid that a particular choice of the seed can significantly condition the final performance.
Python 3.6.6 \cite{van1995python} with PyTorch 0.4.1 \cite{paszke2017automatic} is used as framework for machine learning and automatic differentiable computing. NetworkX 2.2 \cite{hagberg2008exploring} has been used for graph analysis.
Computations were mainly performed using Intel(R) Xeon(R) CPU E5-2650 v3 at 2.30GHz as CPU and GeForce GTX TITAN X as GPU. With this configuration, the proposed CDC in average took approximately 8.3 hours to complete a training procedure on environments with four agents involved.

\subsection{Main results} \label{subsec:main_results}
\small
\begin{table}[t!]
\begin{center}
	\scalebox{0.7}{
		{\color{black}
			\begin{tabular}{l|lll|lll}
				& \multicolumn{3}{c|}{Navigation Control $ N = 3 $	} & \multicolumn{3}{c}{Navigation Control $ N = 10 $	} \\
				& Reward              & \# collisions             & Distance      & Reward                  & \# collisions         & Distance \\
				\hline
				DDPG	& $ -57.3 \pm 9.94  $ & $  1.24 \pm 0.39  $ &  $ 4.09 \pm 6.92  $ & $ -115.93 \pm 21.26  $  & $ 8.83 \pm 6.41  $    & $ 3.6 \pm 0.85  $  \\
				MADDPG	& $ -45.23 \pm 6.59 $ & $ 0.77 \pm 0.24   $ &  $ 3.16 \pm 5.74  $ & $ -112.17 \pm 13.23   $ & $ 12.29 \pm 7.45  $   & $ 3.44 \pm 0.53   $  \\
				CommNet	& $ -48.95 \pm 6.25  $ & $ 0.92 \pm 0.24  $ &  $ 3.49 \pm 5.09  $ & $ -104.49 \pm 10.45   $ & $ 12.21 \pm 6.87  $   & $ 3.14 \pm 0.41  $  \\
				MAAC	& $ -43.18 \pm 6.44  $ & $ 0.71 \pm 0.24  $ &  $ 1.46 \pm 2.97  $ & $ -107.38 \pm 11.81  $  & $ 9.04 \pm 6.46  $    & $ 3.26 \pm 0.46  $  \\
				ST-MARL	& $ -55.36 \pm 8.17  $ & $ 1.54 \pm 3.56  $ &  $1.2 \pm 0.33   $  & $ -110.69 \pm 15.75  $  & $ 32.73 \pm 32.77   $ & $ 3.27 \pm 0.57  $  \\
				When2Com	& $ -40.7 \pm (5.33) $ & $ 0.61 \pm (0.21) $   &  $ 1.06 \pm (3.26) $   & $ -112.51 \pm (14.48) $ & $ 13.68 \pm (11.29) $	& $ 3.45 \pm (0.57) $   \\
				TarMAC	    & $ -44.9 \pm (6.22) $ & $ 0.77 \pm (0.24) $    & $ 2.14 \pm (4.36) $	  & $ -110.67 \pm (13.76) $ & $ 9.81 \pm (7.66) $	 & $ 3.39 \pm (0.54) $  \\
				IS	    & $ -42.6 \pm (6.70) $ & $ 0.70 \pm (0.29) $    & $ 1.22 \pm (3.56) $	  & $ -111.67 \pm (9.18) $ & $ 12.28 \pm (7.27) $	 & $ 3.39 \pm (0.68) $  \\
				CDC   & $ \bm{-39.16 \pm 4.77}  $ & \bm{$ 0.56 \pm 0.19  $} &  $ \bm{0.4 \pm 1.66 } $ & $ \bm{-102.68 \pm 10.1 } $  & $ \bm{9.03 \pm 9.36 } $ & $ \bm{3.06 \pm 0.4  }  $  \\
				\hline
				& \multicolumn{3}{c|}{Formation Control $ N = 4 $	} & \multicolumn{3}{c}{Formation Control $ N = 10 $	} \\
				& Reward               & Time            & Success Rate    & Reward               & Time           & Success Rate \\
				\hline			
				DDPG	& $ -39.43\pm12.37  $ & $ 50\pm0.0 $   &  $ 0 \pm 0.0  $ & $ -49.27\pm6.11  $  & $ 50\pm0.0  $ & $ 0\pm0.0  $  \\
				MADDPG	& $ -19.86\pm6.04  $  & $50\pm0.0   $  &  $ 0 \pm 0.0  $ & $ -20.65\pm7.11  $  & $ 50\pm0.0  $ & $ 0 \pm 0.0   $  \\
				CommNet	& $  -7.77 \pm 2.06   $ & $ 45.8\pm10.19  $ &  $ 0.18 \pm 0.38   $ & $ -10.22\pm1.03  $  & $ 48.89\pm5.5  $ & $ 0.04\pm0.2  $  \\
				MAAC	& $  -5.77 \pm 1.53 $ & $ 26.66\pm17.2  $ &  $ 0.66\pm0.47  $ & $ -9.63\pm1.35  $  & $ 50\pm0.0  $ & $ 0\pm0.0  $  \\
				ST-MARL	& $ -20.24\pm3.0   $  & $ 50\pm0.0  $     &  $ 0\pm0.0  $              & $ -19.81\pm5.74  $ & $ 50\pm0.0  $ & $ 0\pm0.0  $  \\
				When2Com	& $-17.00-\pm(4.16)$	 &  $ 48.21\pm(10.11) $   &$0.12\pm(0.31)$	   &$-18.49\pm(1.23)$ &$48.72\pm(0.9)$ & $0.01\pm(0.1)$	 \\
				TarMAC	    & $-14.25\pm(2.58)$  &  $ 47.35\pm(12.87) $   & $0.13\pm(0.45)$	  &$-19.06\pm(1.23)$ &$49.44\pm(5.6)$ &$0.01\pm(0.1)$	 \\
				IS	    & $-18.72\pm(3.43)$	 &  $ 49.79\pm(9.96) $   &$0.1\pm(0.41)$	   & $ -18.30\pm4.36  $  & $ 50\pm0.0  $ & $ 0 \pm 0.0   $ 	 \\
				CDC    & $ \bm{-4.22 \pm 1.46  }$ & $ \bm{11.82\pm5.49 } $ &  $ \bm{0.99\pm0.12 } $ & $ \bm{-7.51\pm1.06 } $  & $ \bm{15.21\pm9.23 } $ & $ \bm{0.99\pm0.1 } $   \\
				\hline
				& \multicolumn{3}{c|}{Line Control $ N = 4 $} & \multicolumn{3}{c}{Line Control $ N = 10 $	} \\
				& Reward               & Time               	& Success Rate   	 & Reward               & Time           		& Success Rate \\
				\hline			
				DDPG	& $-33.45\pm10.58   $ 	& $ 49.99\pm0.22  $    &  $ 0\pm0.0  $     & $ -68.19\pm10.2  $  & $ 50\pm0.0  $ 		&  $0\pm0.0 $  \\
				MADDPG	& $ -18.75\pm2.32  $ 	& $ 47.32\pm9.14  $ 	&  $ 0.08\pm0.27  $ & $ -12.69\pm2.11  $  & $ 48.48\pm7.12  $ 	& $ 0.04\pm0.21  $  \\
				CommNet	& $ -10.99\pm2.24  $ & $ 46.97\pm8.93  $ &  $ 0.12\pm0.33  $ & $ -9.58\pm1.28  $  & $ 37.73\pm14.85  $ & $ 0.47\pm0.5  $  \\
				MAAC	& $ -7.38\pm2.09 	 $ & $ 17.08\pm12.17  $ &  $ 0.89\pm0.32   $ & $ -8.58\pm1.52  $  & $ 22.55\pm16.09  $ & $ 0.76\pm0.43  $ \\
				ST-MARL	& $ -23.87\pm7.77  $  & $ 50\pm0.0  $      &  $ 0\pm0.0  $       & $ -19.24\pm6.26   $   & $ 50\pm0.0  $      & $ 0\pm0.0  $  \\
				When2Com	&	$-16.45\pm(3.01)$  & $46\pm(0.0)$   & $0.11\pm(0.3)$	  & $-10.1\pm(2.8)$	 &  $ 49.55\pm4.24  $ & $0.01\pm(0.12)$\\
				TarMAC	    & $-17.75\pm(4.24)$	 & $47.00\pm(0.0)$   & $0.09\pm(0.31)$	  & $-11.83\pm(1.63)$	 &  $ 49.91\pm1.12  $  & $0.01\pm(0.09)$\\
				IS	    & $-16.11\pm(4.24)$	 & $45.20\pm(0.0)$   & $0.10\pm(0.15)$	  & $-11.90\pm(1.52)$	 &  $ 49.84\pm1.15  $  & $0.01\pm(0.03)$\\
				CDC    & $\bm{-5.97\pm1.73 }  $ & $ \bm{10.42\pm5.58 } $ &  $ \bm{0.98\pm0.13 } $ & $ \bm{-7.96\pm1.19 } $  & $\bm{15.06\pm12.02 }  $ & $ \bm{0.91\pm0.29 } $ \\
				\hline
				& \multicolumn{3}{c|}{Dynamic Pack Control $ N = 4 $	} & \multicolumn{3}{c}{Dynamic Pack Control $ N = 8 $	} \\
				& Reward               & Distance               & Targets caught   			& Reward               & Distance               & Targets caught  \\	
				\hline		
				DDPG	& $ -224.77\pm87.65  $ & $ 3.52\pm1.67  $ 	&  $ 0\pm0.0  $ 		& $ -279.67\pm70.18  $  	& $ 4.58\pm1.4  $ 	& $ 0\pm0.0  $  \\
				MADDPG	& $ -116.15\pm71.37  $ & $ 1.46\pm0.72  $ 	&  $ 0.2\pm0.13  $ 	& $ -110.86\pm28.66  $ 	& $ 1.22\pm0.28  $ &  $ 0.0\pm0.05  $ \\
				CommNet	& $ 293.35\pm446.89  $ & $ 1.11\pm0.12  $ &  $ 0.81\pm0.89  $ & $ -76.18\pm138.73  $  & $ 1.13\pm0.25  $ & $ 0.07\pm0.28  $  \\
				MAAC	& $ -95.29\pm61.65  $ & $ 1.25\pm0.21  $ &  $ 0.01\pm0.12  $ & $ -105.15\pm46.42  $  & $ 1.15\pm0.28 $  & $ 0.01\pm0.09  $ \\
				ST-MARL	& $ -107.02\pm71.84  $ & $ 1.26\pm0.3  $ &  $ 0.02\pm0.14   $ & $ -123.91\pm16.89  $  & $ 1.42\pm0.36  $ & $ 0\pm0.0   $  \\
				When2Com	& $-108.47\pm(73.58)$ & $1.32\pm(0.33)$	   & $ 0.02\pm(0.14)  $  & $-111.47\pm(73.58)$ & $1.32\pm(0.33)$	   & $	0.02\pm(0.14)$ \\
				TarMAC	    & $50.47\pm(73.58)$ & $ 1.20\pm0.21  $   & $ 0.3\pm0.55  $  & $ -78.18\pm42.5  $  & $ 1.18\pm0.76  $ & $ 0.05\pm0.21  $ \\
				IS	& $ 235.74\pm446.89  $ & $ 1.06\pm0.35  $ &  $ 0.80\pm0.63  $ & $ 50.19\pm310.44  $  & $ {1.10\pm0.29 } $ & $ {0.34\pm0.98 } $  \\
				CDC    & $ \bm{369.5\pm463.92 } $ & $ \bm{1.09\pm0.1 } $ &  $ \bm{0.96\pm0.93 } $ & $ \bm{58.03\pm279.05 } $  & $ \bm{1.12\pm0.14 } $ & $ \bm{0.35\pm0.56 } $  \\						
				\hline
			\end{tabular}
		}
	}
	\vspace{1mm}
	\caption{{\color{black} \textmd{Comparison of DDPG, MADDPG, CommNet, MAAC, ST-MARL, When2Com, TarMAC, IS and CDC on all environments. $N$ is the number of agents. Results are averaged over five different seeds.}}}
	\label{tab:mainresults}
\end{center}
\end{table}
\normalsize

We have compared CDC against several different baselines, each one representing a different way to approach the MA coordination problem: independent DDPG \cite{silver2014deterministic,lillicrapHPHETS15}, MADDPG \cite{lowe2017multi}, CommNet \cite{sukhbaatar2016learning}, MAAC \cite{iqbal2018actor}, ST-MARL \cite{wang2019stmarl}, When2Com \cite{liu2020when2com} and TarMAC \cite{das2018tarmac} {\color{black} and Intention Sharing (IS\footnote{{\color{black}In our implementation, the number of steps to be predicted is set to one, i.e. each agent predicts the next step of every other agent. In the original paper, this is the equivalent to IS(H=1). In addition, in order to maintain a fair comparison with the other baselines, a  message at time $ t $ is used to generate the next actions, i.e. we do not rely on previously generated messages.}}) \cite{kim2020communication}}. Independent DDPG provides the simplest baseline in that each agent works independently to solve the task. In MADDPG each agent has its own critic with access to combined observations and actions from all agents during learning. CommNet implements an explicit form of communication; the policies are implemented through a large neural network with some components of the networks shared across all the agents and others agent-specific. At every time-step each agent's action depends on the local observation, and on the average of all other policies (neural network hidden states), used as messages. MAAC is a state-of-the art method in which an attention mechanism guides the critics to select the information to be shared with the actors. ST-MARL uses a graph neural network to capture the spatio-temporal dependency of the observations and facilitate cooperation. Unlike our approach, the graph edges here represents the time-depending agents' relationships, and capture the spatial and temporal dependencies amongst agents.  When2Com utilises an attentional model to compute pairwise similarities between the agents' observation encodings, which results in a fully connected graph that is subsequently sparsified by a thresholding operation. Afterwards, each agent uses the remaining similarities scores to weight its neighbor observations before producing its action. TarMac is a framework where the agents broadcast their messages and then select whom to communicate to by aggregating the received communications together through an attention mechanism. 
{\color{black} In IS \cite{kim2020communication} the agents generate their future intentions by simulating their trajectory and then an attention model aggregate this information together to share it with the others.}
Differently from the methods above, CDC utilises graph structures to support the formation of communication connectivities and then use the heat kernel, as an alternative form of attention mechanism, to allow to each agent to aggregate the messages coming from the others.
\begingroup
\setlength{\tabcolsep}{10pt} % Default value: 6pt
\renewcommand{\arraystretch}{1.5} % Default value: 1
\begin{table}[h]
\begin{center}
	\scalebox{0.7}{
		{\color{black}
			\begin{tabular}{|C{1.5cm}|C{2.5cm}|C{3.5cm}|C{2.5cm}|C{3cm}|}
				\hline
				&  Type of communication   &  How information is aggregated  &  Has a graph-based architecture   & Is communication delayed                        \\ 
				\hline
				%					------------------------------------------------------------------------------	
				DDPG	& NA & NA  & No & NA     
				\\
				\hline
				%					------------------------------------------------------------------------------	
				MADDPG
				& Implicit &   
				\centered{
					Observation and \\ action concatenation 
				} 
				& No  & Yes    
				\\
				\hline
				%					--------------------------------------------------------------------------
				CommNet	& Explicit & \centered{
					Sharing neural-networks \\ hidden states
				}  &  No & No   \\
				\hline
				%					--------------------------------------------------------------------------
				MAAC	&   Implicit   & Attention  &  No & Yes     \\ \hline
				%					--------------------------------------------------------------------------
				ST-MARL	& Implicit & RNN + Attention  &  Yes  & No    \\ \hline
				%					--------------------------------------------------------------------------
				When2Com	& Implicit & Attention    &  Yes  & No 	   \\\hline
				%					--------------------------------------------------------------------------
				TarMAC	    & Explicit  & Attention   & No	  &  Yes	  \\ \hline
				%					--------------------------------------------------------------------------
				IS	    & Explicit   &  Attention    & No	  & No  \\ \hline
				%					--------------------------------------------------------------------------
				CDC   & Explicit & Heat Kernel & Yes  & No  \\ \hline
				%					--------------------------------------------------------------------------
			\end{tabular}
		}
	}
	\vspace{1mm}
	\caption{{\color{black} \textmd{A comparative summary of various MARL algorithms according to how communication is implemented.}}}
	\label{tab:methods_comparison}
\end{center}
\end{table}
\normalsize
\endgroup

{\color{black}
In Table \ref{tab:methods_comparison} we provide a summary of selected features for each MADRL algorithm used in this work. First, we have indicated whether the communication is implicit or explicit. {\color{black} The former refers to the ability to share information without sending explicit messages, i.e. communication is inherited from a certain behaviour rather than being deliberately shared \cite{breazeal2005effects}; studies have shown that this approach is used by both animals and humans \cite{mech2007wolves,quick2012bottlenose,schaller2009serengeti} and has discussed in a number of multi-agent reinforcement learning works \cite{das2018tarmac, montesello1998implicit,li2020deep,grupen2022multi,sukhbaatar2016learning,singh2018learning,peng2017multiagent}. Explicit communication assumes the existence of a specific mechanism deliberatively introduced to share information within the system; this is considered to be the most common form of human communication \cite{gildert2018need,haakansson2013communication} and has also been widely explored in the context of reinforcement learning \cite{pesce2019improving,kim2020communication,liu2020when2com,foerster2016learning}.}
This categorization can help interpret the performance achieved in certain environments, such as Dynamic Pack Control, where explicit communication is more beneficial.

We also report on how the information is aggregated amongst agents, whether the algorithm relies on a graph-based architecture, and whether the communication content is delayed, i.e. it only utilised in the future but does not affect the current actions. For example, in TarMac, each message is broadcasted and utilised by the agents in the next step, while in MAAC and MADDPG the communication happens through the critics and affect future actions once the policy parameters get updated.

} % end color_red
Table \ref{tab:mainresults} summarises the experimental results obtained from all algorithms across all the environments. The metric values are obtained by executing the best model (chosen according to the best average reward returned during training) for an additional $ 100 $ episodes. We repeated each experiment using $5$ different seeds, and each entry of Table \ref{tab:mainresults} is an average over $ 500 $ values.

%	\begin{figure}
%		\begin{minipage}[b]{1\linewidth}
%			\centering
%			\includegraphics[width=1\linewidth]{img_lcurve_seeds_simple_formation_po10.jpg} 
%		\end{minipage} 
%		\caption*{\textmd{(a) Formation Control}}
%		\begin{minipage}[b]{1\linewidth}
%			\centering
%			\includegraphics[width=1\linewidth]{img_lcurve_seeds_simple_spread_pack_leader_8_2.jpg} 
%		\end{minipage} 
%		\caption*{\textmd{(b) Dynamic Pack Control}}
%		\caption{\textmd{Learning curves for all the comparison algorithms on Formation Control (a) and Dynamic Pack Control (b). The x-axis the number of episodes and the y-axis the achieved rewards. All the results are averaged over five different runs.}}
%		\label{fig:learningcurves} 
%	\end{figure}
\begin{figure}[H] 
\begin{minipage}[b]{0.5\linewidth}
	\centering
	\includegraphics[width=1\linewidth]{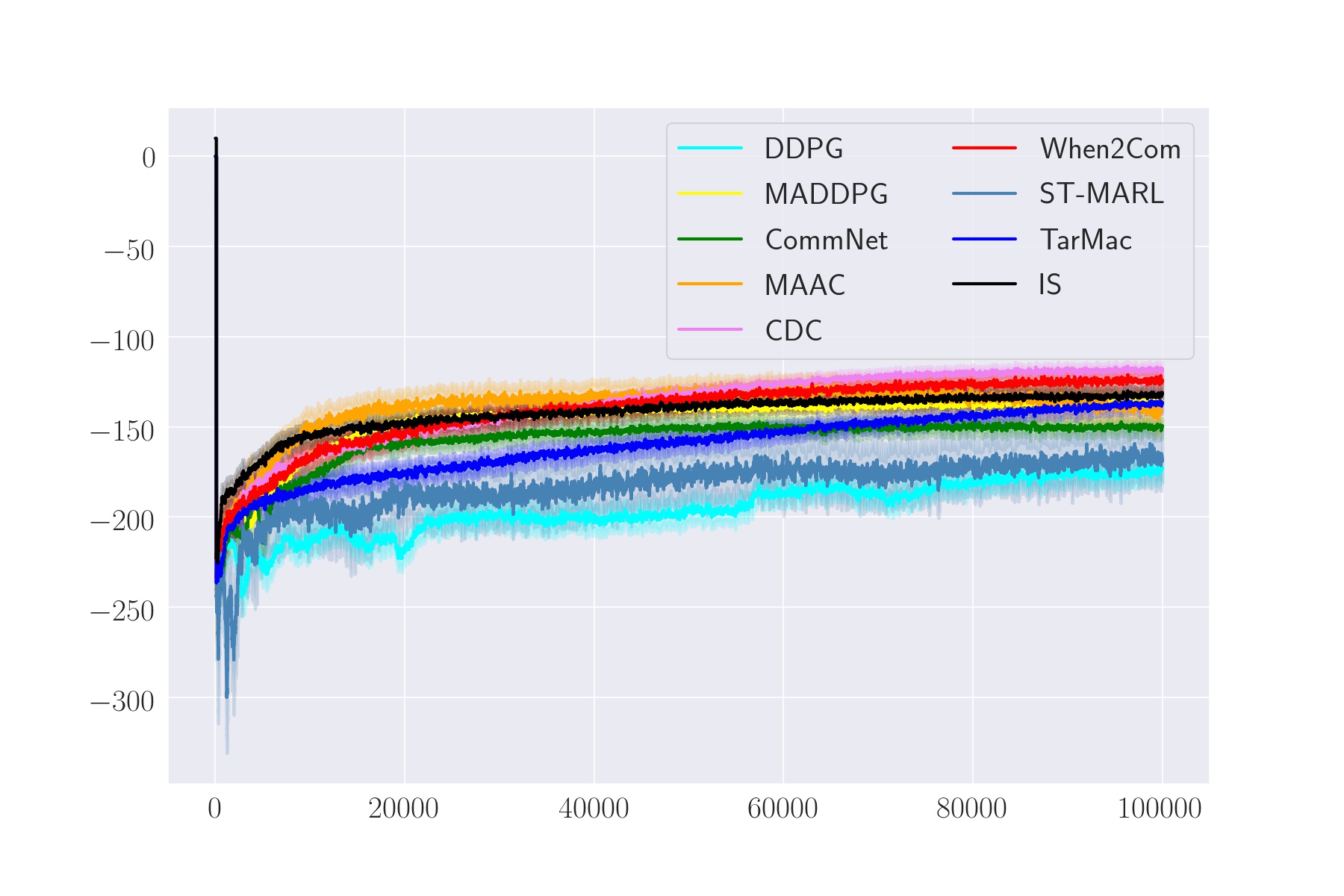} 
\end{minipage}%% 
\begin{minipage}[b]{0.5\linewidth}
	\centering
	\includegraphics[width=1\linewidth]{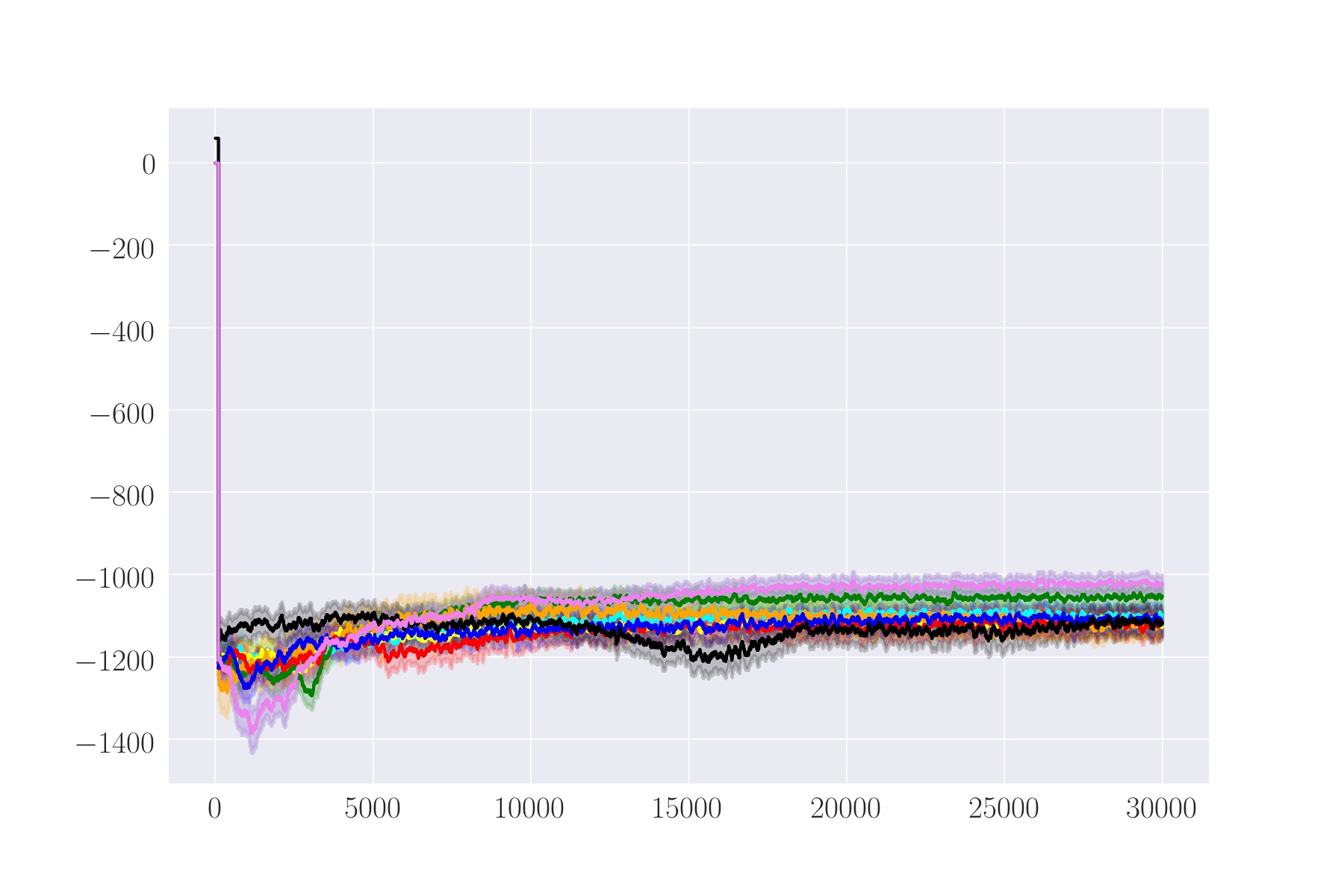} 
\end{minipage}%% % 
\caption*{\hspace{0.5cm}\textmd{Navigation Control $ N = 3 $}\hspace{2.5cm}\textmd{Navigation Control $ N = 10 $}}
%%%%%%%%%%%%%%%%%%%%%%
\begin{minipage}[b]{0.5\linewidth}
	\centering
	\includegraphics[width=1\linewidth]{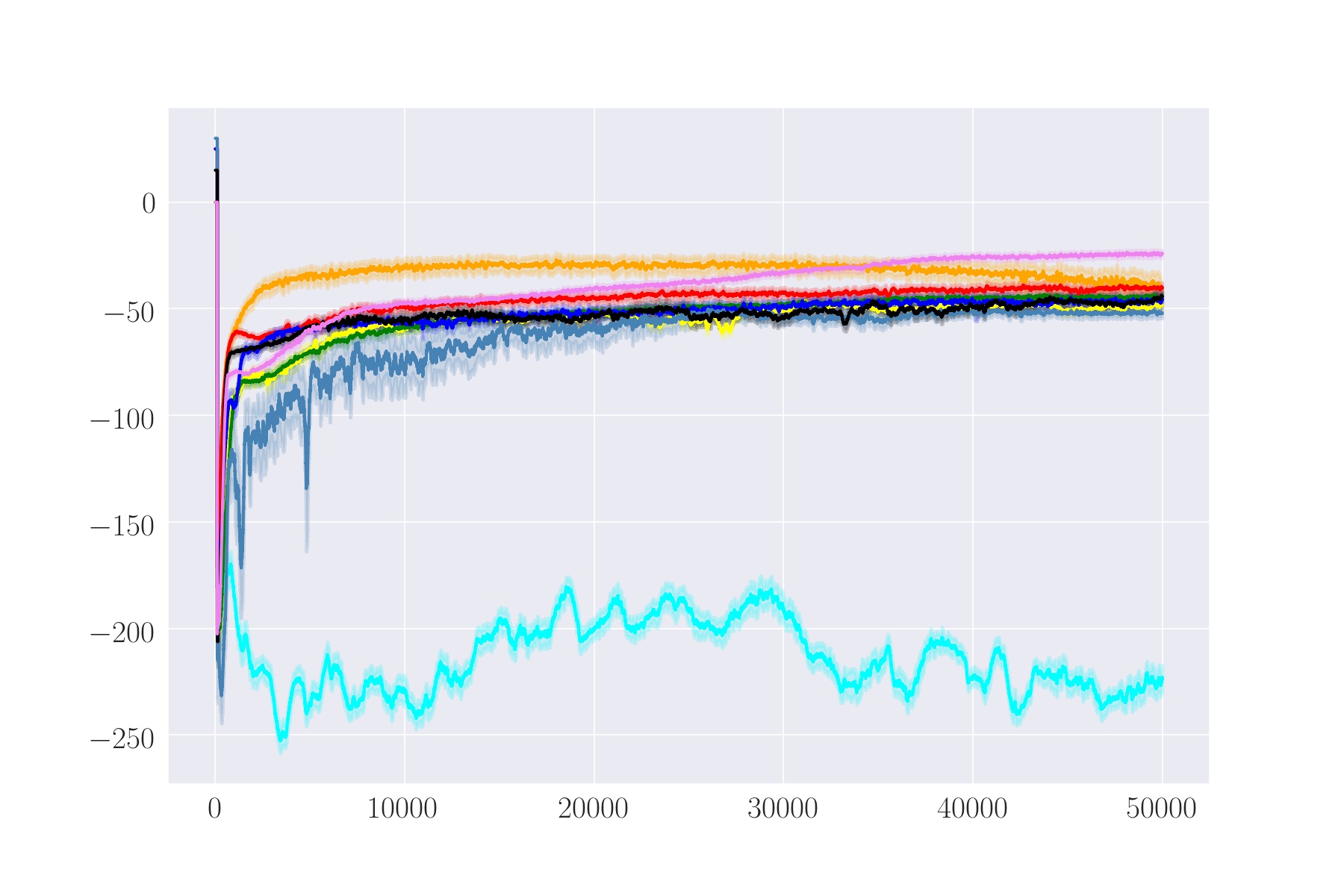} 
\end{minipage}%% 
\begin{minipage}[b]{0.5\linewidth}
	\centering
	\includegraphics[width=1\linewidth]{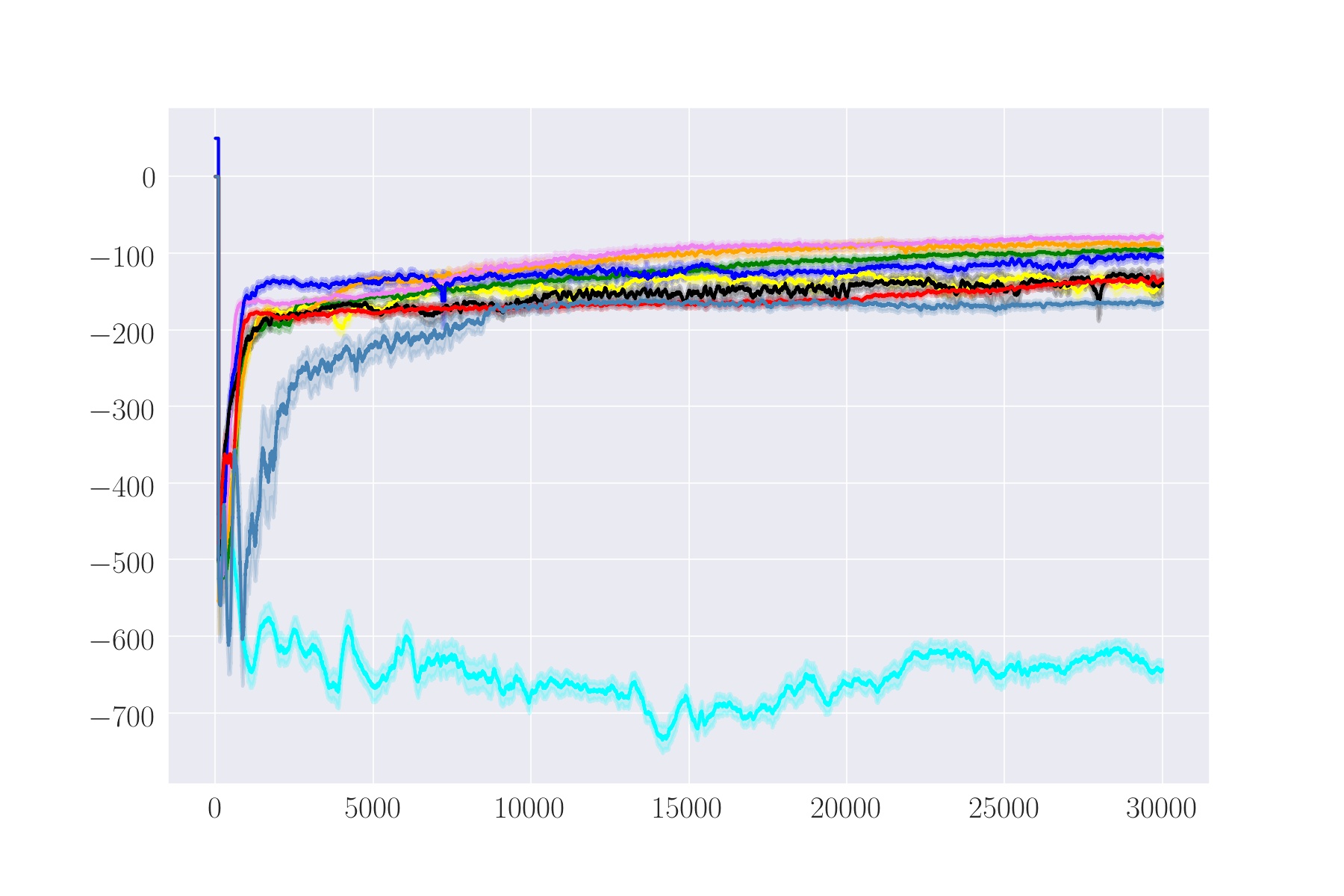} 
\end{minipage}%% % 
\caption*{\hspace{0.5cm}\textmd{Line Control $ N = 4 $}\hspace{2.5cm}\textmd{Line Control $ N = 10 $}}	
%		\caption{\textmd{Learning curves for $6$ competing algorithms assessed on Navigation Control and Line Control. Horizontal axes report the number of episodes, while vertical axes the achieved rewards. Results are averaged over five different runs.}}
%		\label{fig:supp_learning_curves1} 
%	\end{figure*}
%	
%	\begin{figure}[H] 
\begin{minipage}[b]{0.5\linewidth}
	\centering
	\includegraphics[width=1\linewidth]{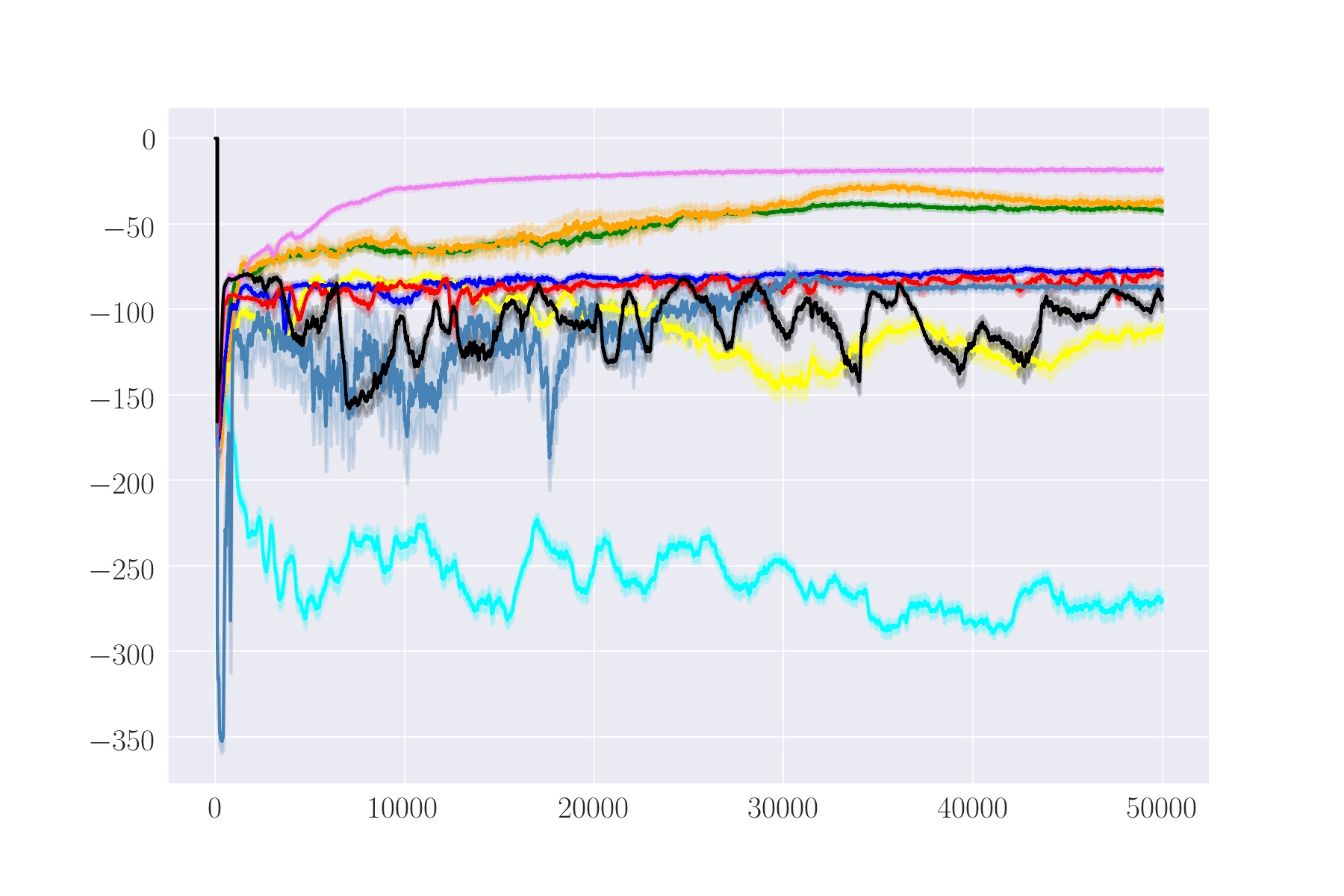} 
\end{minipage}%% 
\begin{minipage}[b]{0.5\linewidth}
	\centering
	\includegraphics[width=1\linewidth]{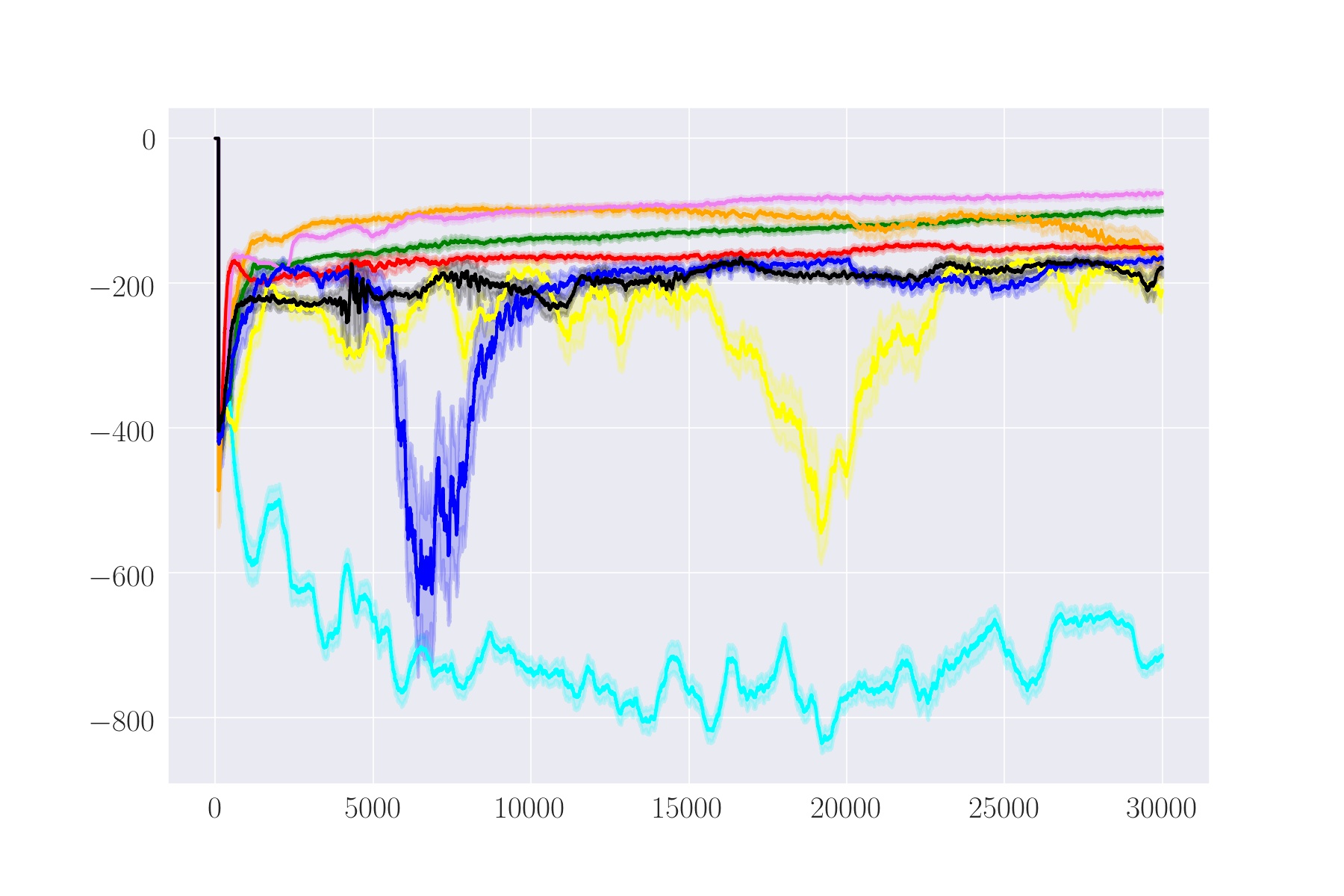} 
\end{minipage}%% % 
\caption*{\hspace{0.5cm}\textmd{Formation Control $ N = 4 $}\hspace{2.5cm}\textmd{Formation Control $ N = 10 $}}
%%%%%%%%%%%%%%%%%%%%%%
\begin{minipage}[b]{0.5\linewidth}
	\centering
	\includegraphics[width=1\linewidth]{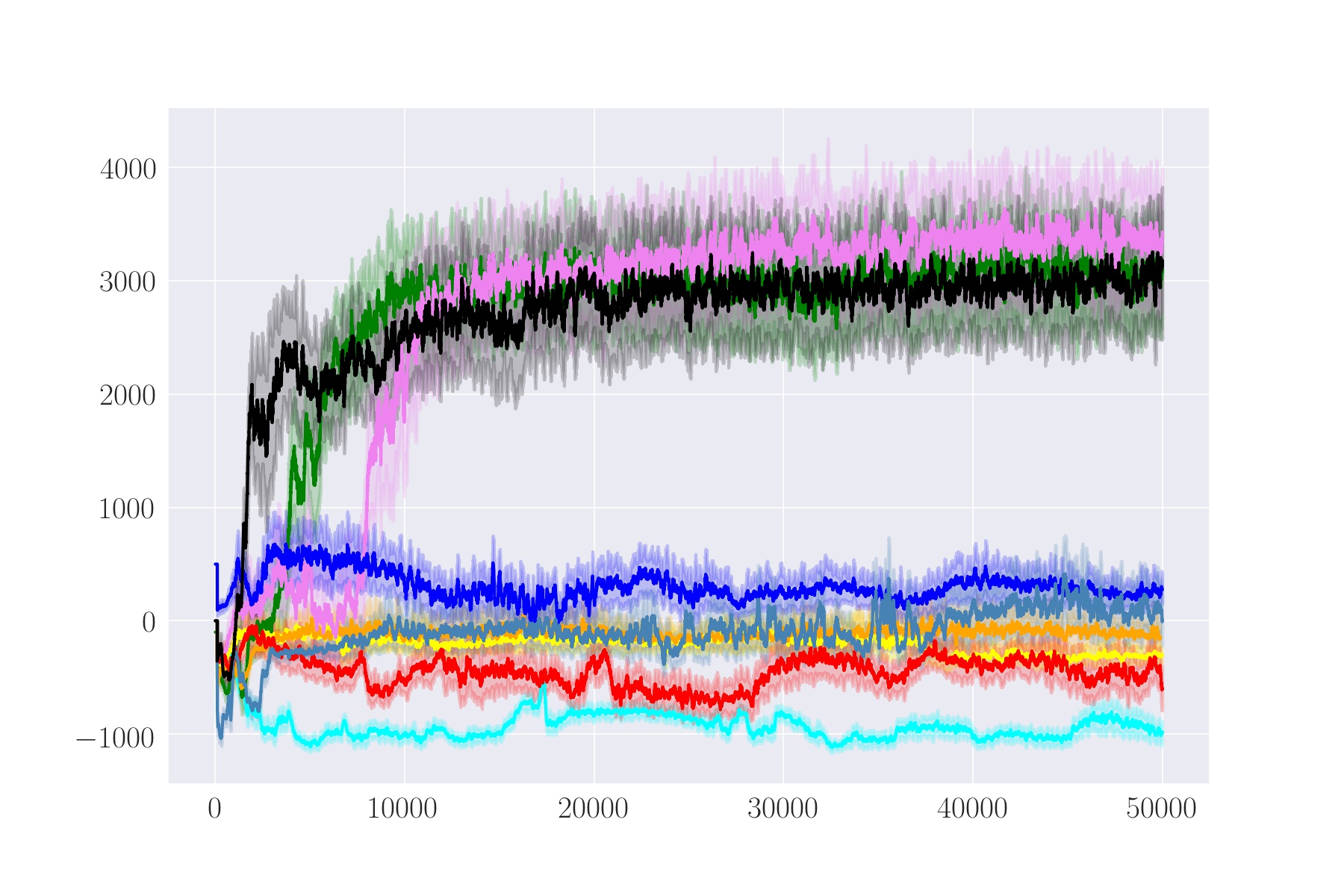} 
\end{minipage}%% 
\begin{minipage}[b]{0.5\linewidth}
	\centering
	\includegraphics[width=1\linewidth]{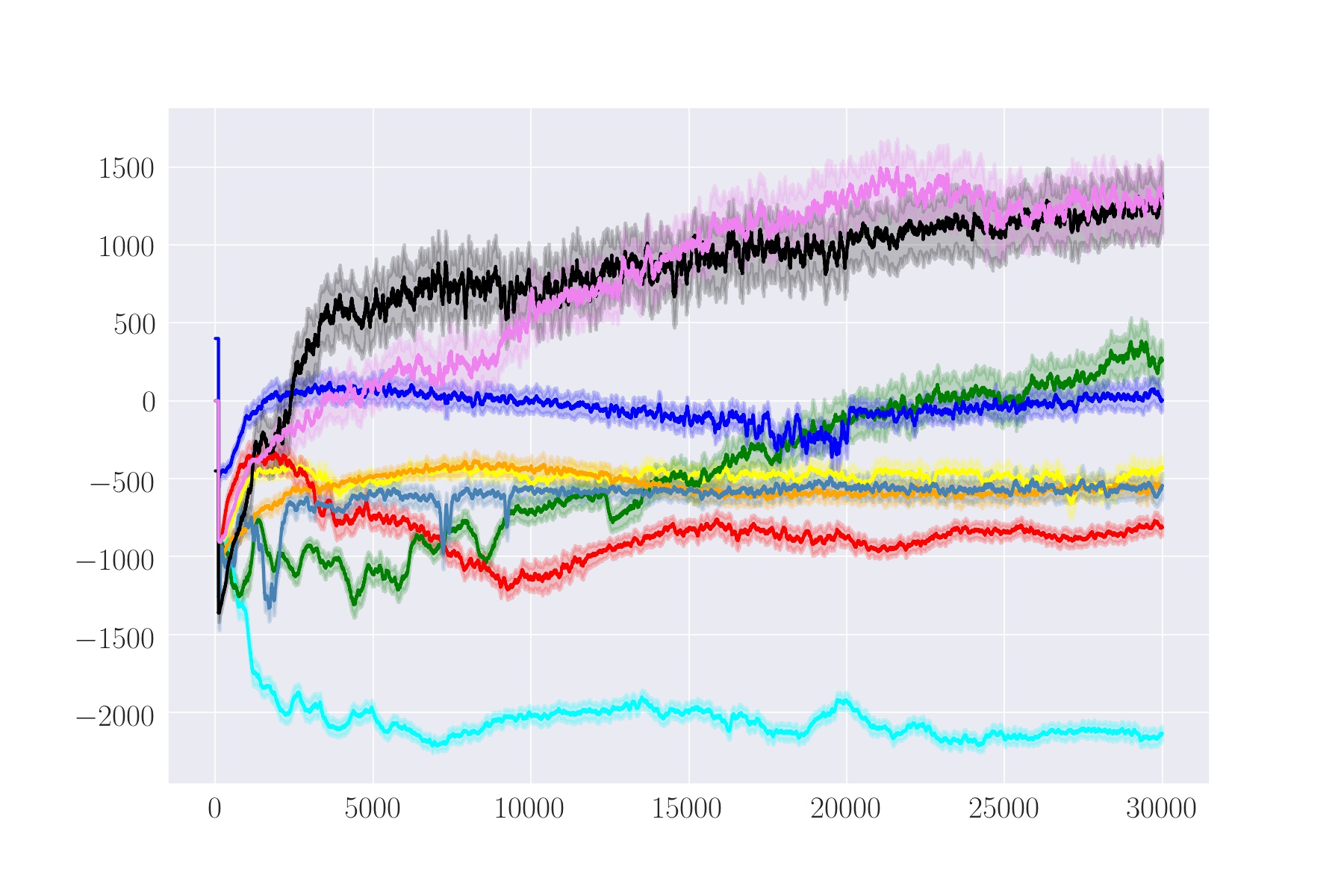} 
\end{minipage}%% % 
\caption*{\hspace{0.5cm}\textmd{Dynamic Pack Control $ N = 4 $}\hspace{1.7cm}\textmd{Dynamic Pack Control $ N = 8 $}}	
\caption{\textmd{{\color{black}Learning curves for $9$ competing algorithms assessed on Navigation Control, Line Control, Formation Control and Dynamic Pack Control. Horizontal axes report the number of episodes, while vertical axes the achieved rewards. Results are averaged over five different runs.}}}
\label{fig:learningcurves} 
\end{figure}
It can be noted that CDC outperforms all the competitors on all four environments on all the metrics. In Navigation Control ($ N=3 $), the task is solved by minimizing the overall distance travelled and the number of collisions, with an improvement over MAAC. In Formation Control ($ N=4 $), the best performance is also achieved by CDC, which always succeeded in half of time compared to MAAC. 

When the number of agents is increased, and the level of difficulty is significantly higher, all the baselines fail to complete the task whilst CDC still maintains excellent performance with a success rate of $ 0.99 $. In Line Control, both scenarios ($ N=4 $ and $ N=10 $) are efficiently solved by CDC with higher success rate and less time compared to MAAC, while all other algorithms fail. For Dynamic Pack Control, amongst the competitors, only CommNet does not fail. In this environment, only the leaders can see the point of interest, hence the other agents must learn how to communicate with them.  In this case, CDC also outperforms CommNet on both the number of targets that are being caught and travelled distance. Overall, it can be noted that the gains in performance achieved by CDC, compared to other methods, significantly increase when increasing the number of agents.

Learning curves for {\color{black} all the environments}, averaged over five runs, are shown in Figure \ref{fig:learningcurves}. Here it can be noticed that CDC reaches the highest reward overall. The Dynamic Pack Control task is particularly interesting  as only {\color{black}three} methods are capable of solving it, {\color{black}CommNet, IS and CDC}, and all of them implement explicit communication mechanisms. 
%With only $4$ agents, the reward curves of CommNet, MAAC, When2Com, TarMac and CDC tend to be approximately similar, but increasing the numbers of agents to $8$ highlights the remarkable benefits introduced by CDC.
The high variance associated with CDC and CommNet in Dynamic Pack Control can be explained by the fact that, when a landmark is reached by all the agents, the environment returns a higher reward. These are the only two methods capable of solving the task, and lower variance is associated to other methods that perform poorly.
The performance of CDC when varying the number of agents at execution time is investigated (see Appendix, Section \ref{sec:supp_varna}).

%\begin{figure}[t!] 
%	%	\begin{minipage}[b]{0.9\linewidth}
%	%		\centering
%	%		\includegraphics[width=1\linewidth]{img_lcurve_seeds_simple_formation_po} 
%	%	\end{minipage}%% 
%	%	\vspace{-5mm}
%	%	\caption*{\textmd{Formation Control $ N = 4 $}}
%	\begin{minipage}[b]{0.9 \linewidth}
%		\centering
%		\includegraphics[width=1\linewidth]{img_lcurve_seeds_simple_formation_po10} 
%	\end{minipage} 
%	\vspace{-5mm}
%	\caption*{\textmd{Formation Control $ N = 10 $}}
%	%%%%%%%%%%%%%%%%%%%%%%%
%	%	\begin{minipage}[b]{0.9\linewidth}
%	%		\centering
%	%		\includegraphics[width=1\linewidth]{img_lcurve_seeds_simple_spread_pack_leader_4_2} 
%	%	\end{minipage}%% 
%	%	\vspace{-5mm}
%	%	\caption*{\textmd{Pack Control $ N = 4 $}}
%	\begin{minipage}[b]{0.9 \linewidth}
%		\centering
%		\includegraphics[width=1\linewidth]{img_lcurve_seeds_simple_spread_pack_leader_8_2} 
%	\end{minipage} 
%	\vspace{-5mm}
%	\caption*{\textmd{Dynamic Pack Control $ N = 8 $}}
%	\caption{\textmd{Learning curves for comparison algorithms on Formation Control and Dynamic Pack Control. On the x-axis the number of episodes and on y-axis the achieved rewards. Results are averaged over five different runs.}}
%	%			X-axes report the number of episodes, while vertical axes the achieved rewards. }}
%	\label{fig:learningcurves} 
%\end{figure}

\subsection{Communication analysis}

\begin{figure}[H] 
%	\begin{minipage}[b]{0.95\linewidth}
%		\centering
%		\includegraphics[width=0.95\linewidth]{img_time}
%	\end{minipage}
\begin{minipage}[b]{0.95\linewidth}
	\centering
	\begin{minipage}[b]{0.20\linewidth}
		\centering
		\includegraphics[width=0.9\linewidth,fbox]{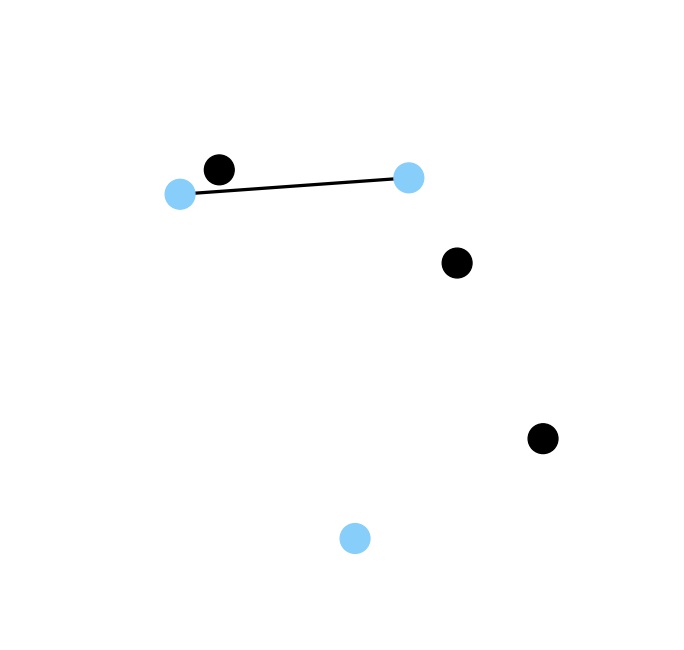} 
	\end{minipage}%% 	\begin{minipage}[b]{0.20\linewidth}
	\begin{minipage}[b]{0.20\linewidth}
		\centering
		\includegraphics[width=0.9\linewidth,fbox]{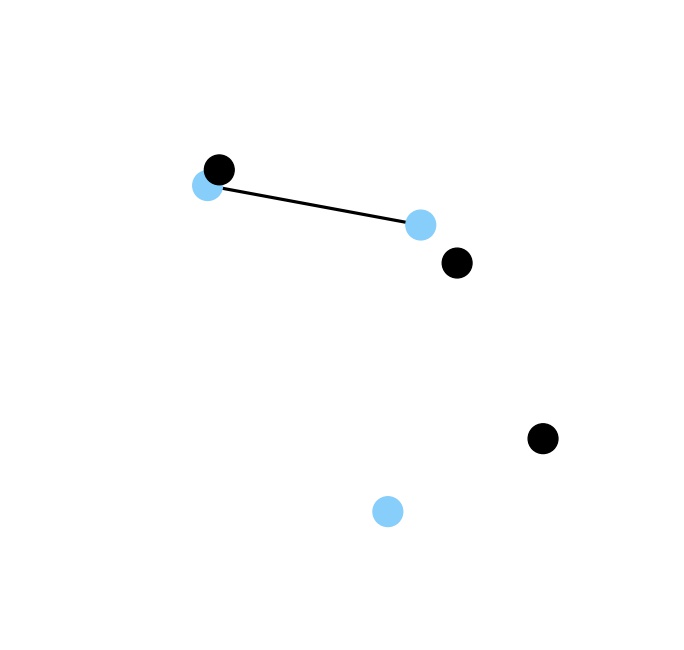} 
	\end{minipage}%% 
	\begin{minipage}[b]{0.20\linewidth}
		\centering
		\includegraphics[width=0.9\linewidth,fbox]{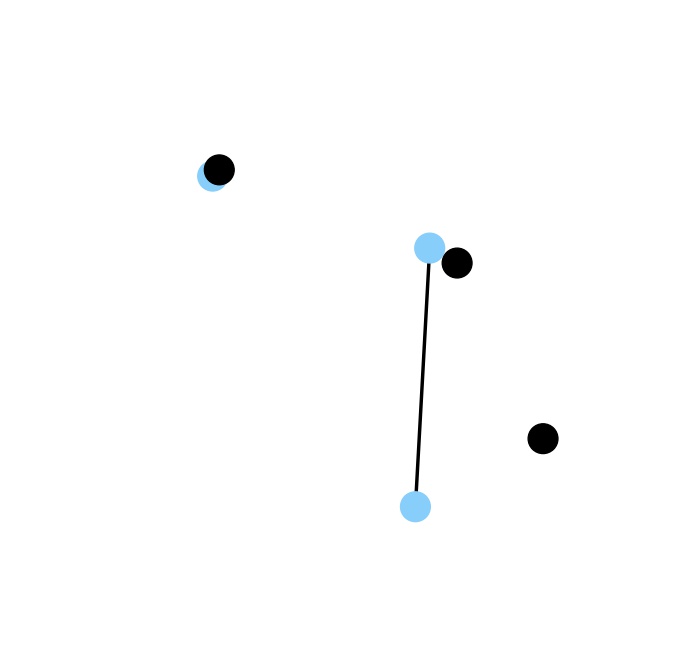} 
	\end{minipage}%% 
	\begin{minipage}[b]{0.20\linewidth}
		\centering
		\includegraphics[width=0.9\linewidth,fbox]{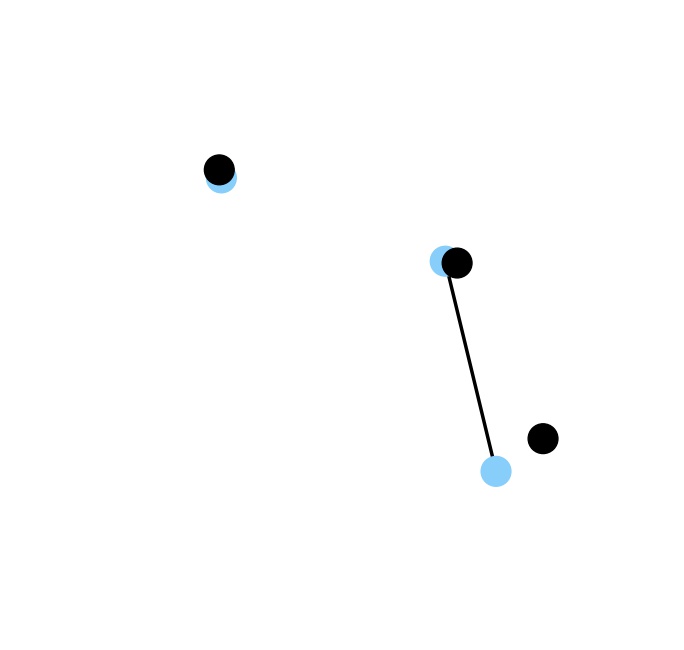} 
	\end{minipage}%% 
	\begin{minipage}[b]{0.20\linewidth}
		\centering
		\includegraphics[width=0.9\linewidth,fbox]{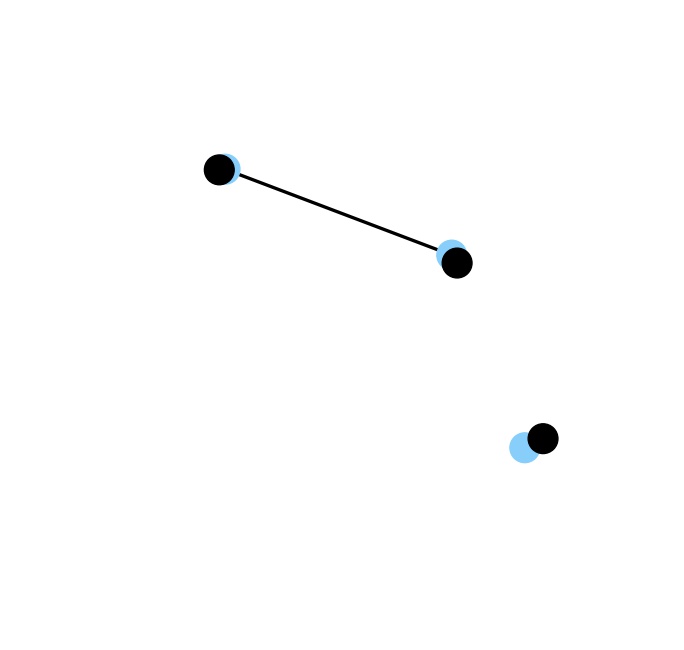} 
	\end{minipage}%% 
	\vspace{-1mm}
	\caption*{\textmd{(a) Navigation Control $ N=3 $}}
	%%%%%%%%%%%%%%%%%%%%%%%%%%%%%%%%%%%
	\begin{minipage}[b]{0.20\linewidth}
		\centering
		\includegraphics[width=0.9\linewidth,fbox]{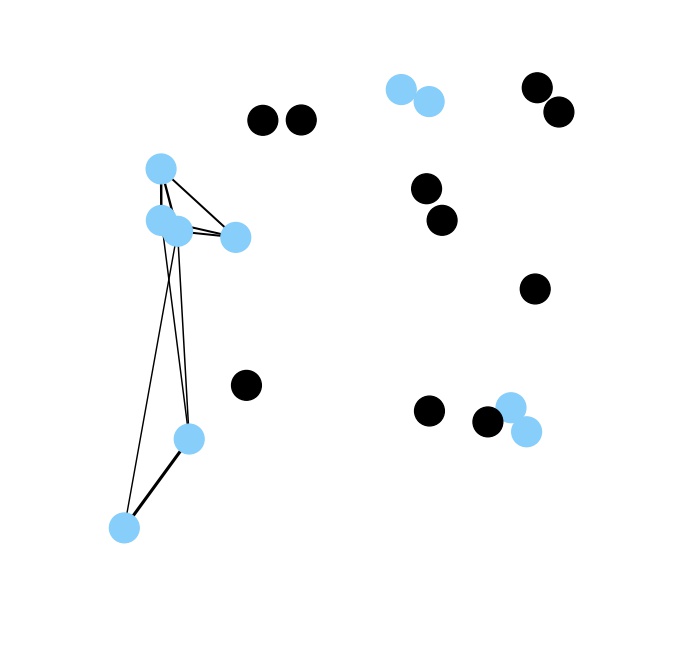} 
	\end{minipage}%% 	\begin{minipage}[b]{0.20\linewidth}
	\begin{minipage}[b]{0.20\linewidth}
		\centering
		\includegraphics[width=0.9\linewidth,fbox]{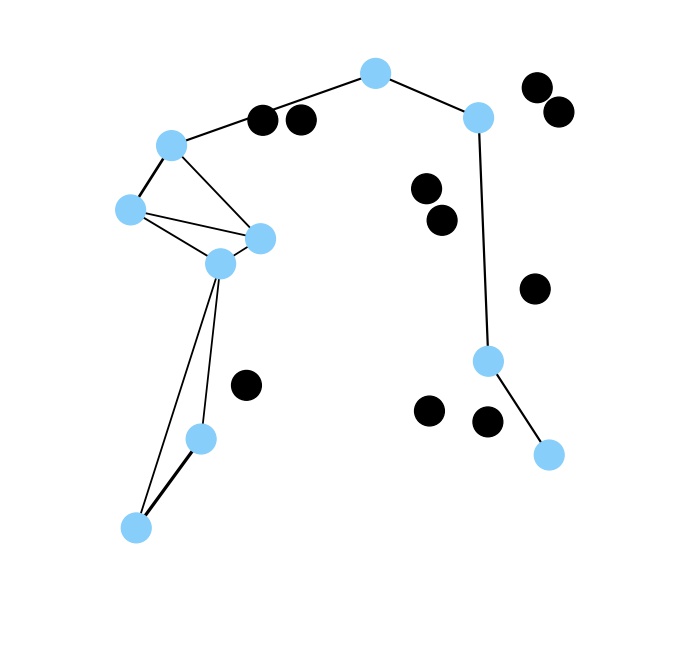} 
	\end{minipage}%% 
	\begin{minipage}[b]{0.20\linewidth}
		\centering
		\includegraphics[width=0.9\linewidth,fbox]{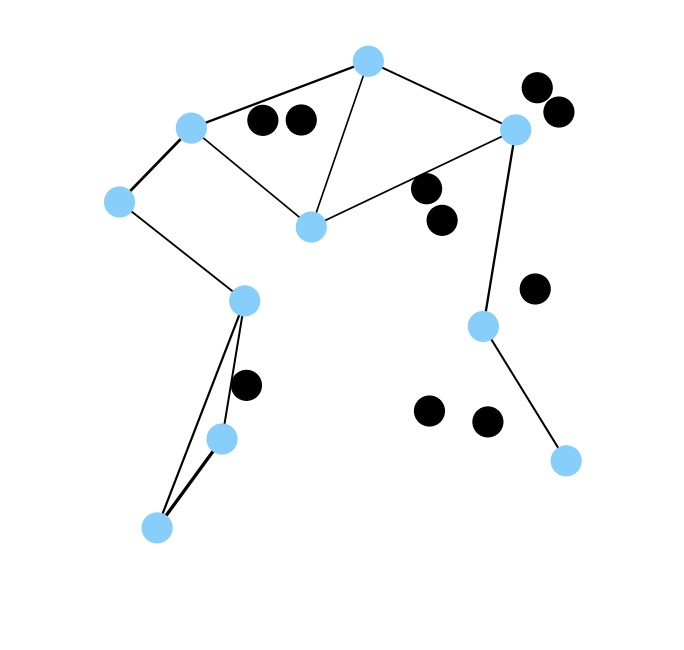} 
	\end{minipage}%% 
	\begin{minipage}[b]{0.20\linewidth}
		\centering
		\includegraphics[width=0.9\linewidth,fbox]{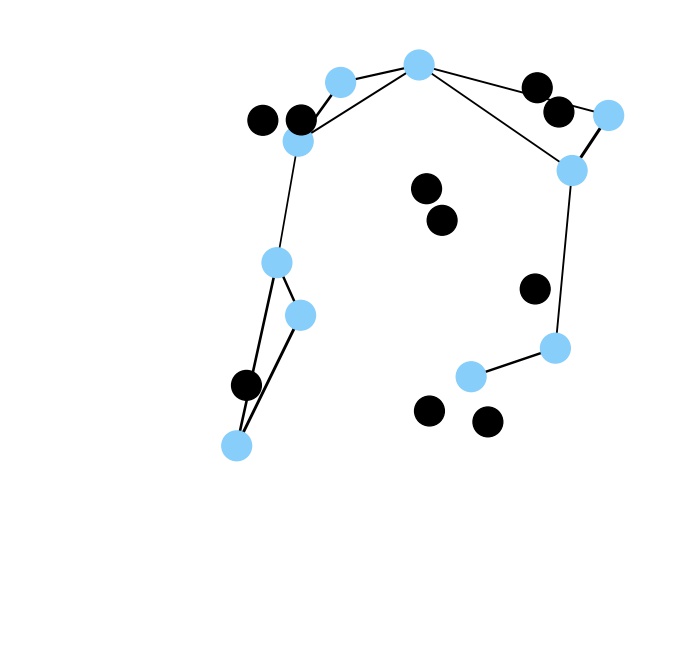} 
	\end{minipage}%% 
	\begin{minipage}[b]{0.20\linewidth}
		\centering
		\includegraphics[width=0.9\linewidth,fbox]{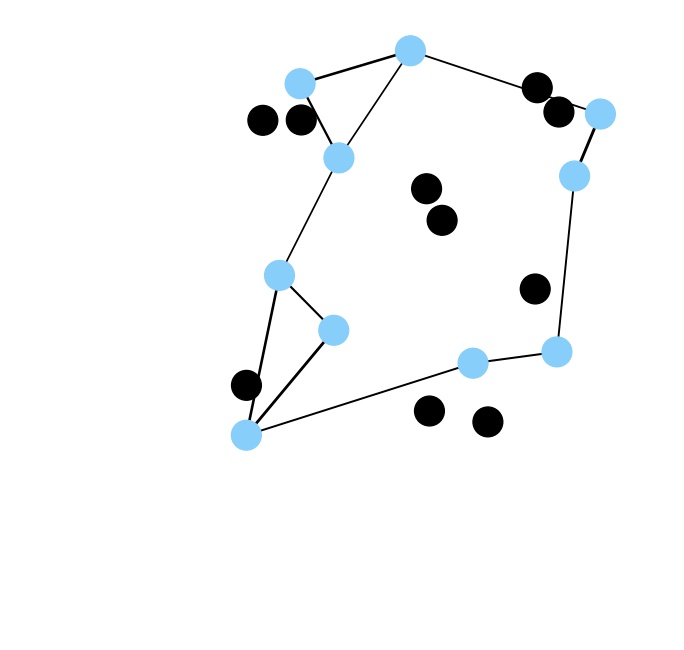} 
	\end{minipage}%% 
	\vspace{-1mm}
	\caption*{\textmd{(b) Navigation Control $ N=10 $}}
	%%%%%%%%%%%%%%%%%%%%%%%%%%%%%%%%%%%
	\begin{minipage}[b]{0.20\linewidth}
		\centering
		\includegraphics[width=0.9\linewidth,fbox]{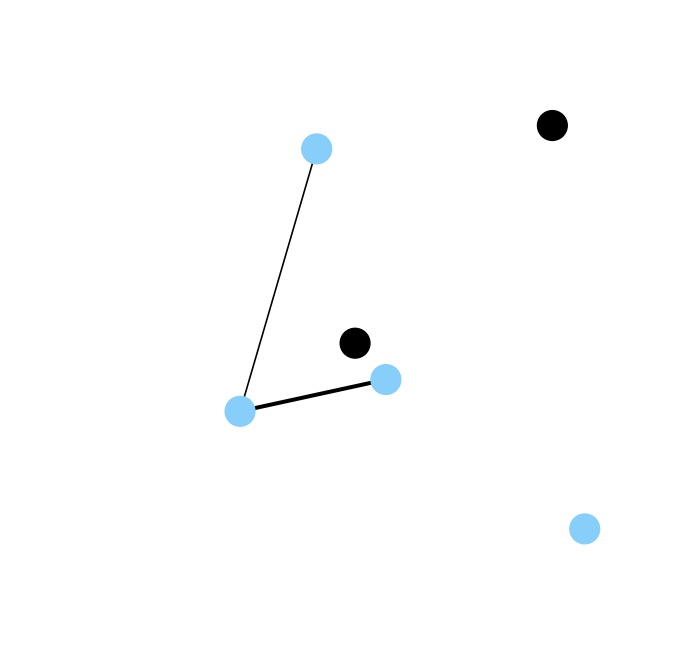} 
	\end{minipage}%% 	\begin{minipage}[b]{0.20\linewidth}
	\begin{minipage}[b]{0.20\linewidth}
		\centering
		\includegraphics[width=0.9\linewidth,fbox]{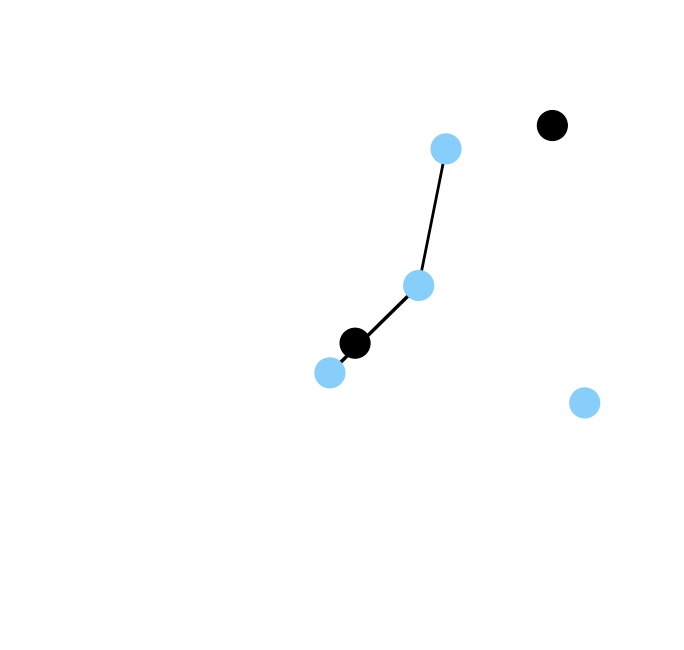} 
	\end{minipage}%% 
	\begin{minipage}[b]{0.20\linewidth}
		\centering
		\includegraphics[width=0.9\linewidth,fbox]{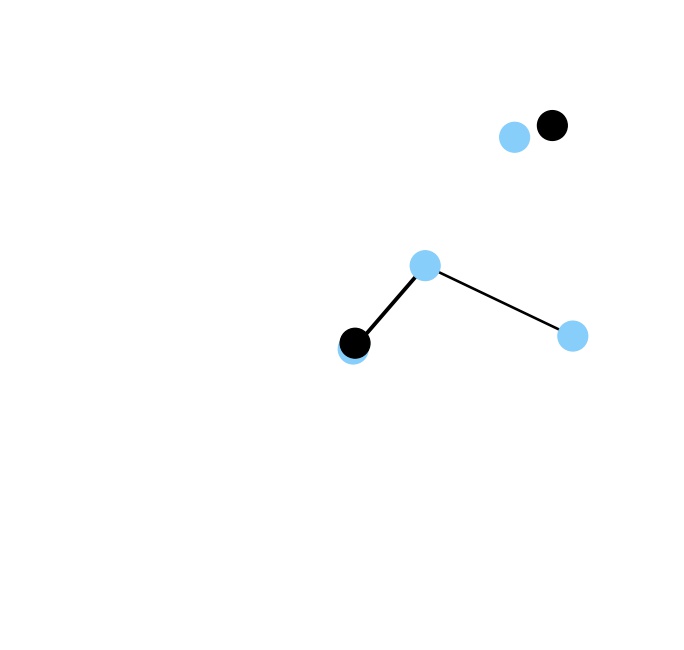} 
	\end{minipage}%% 
	\begin{minipage}[b]{0.20\linewidth}
		\centering
		\includegraphics[width=0.9\linewidth,fbox]{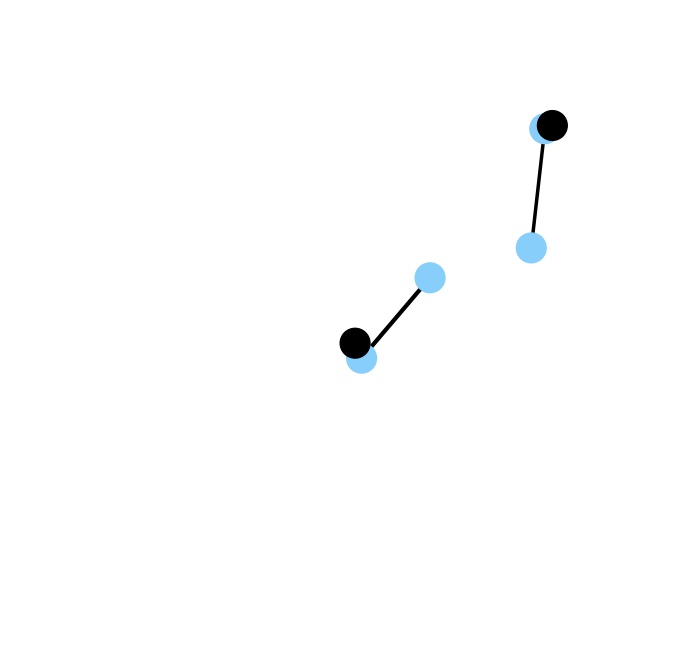} 
	\end{minipage}%% 
	\begin{minipage}[b]{0.20\linewidth}
		\centering
		\includegraphics[width=0.9\linewidth,fbox]{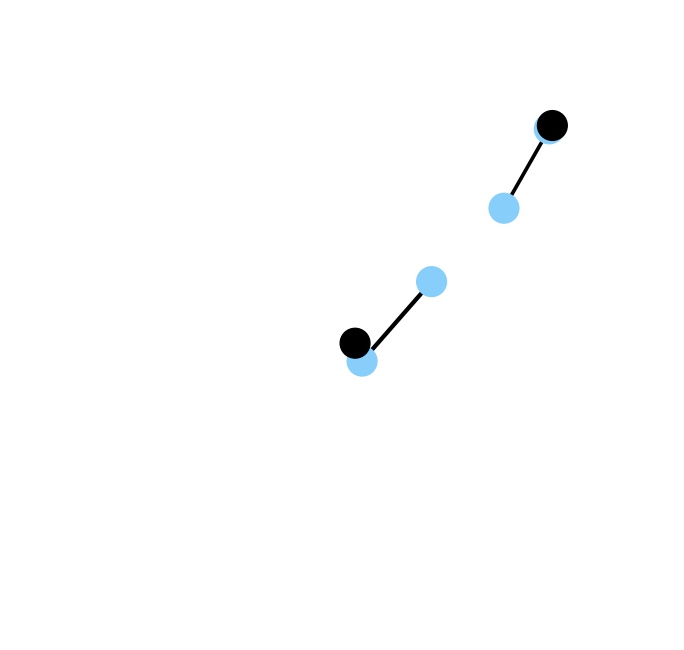} 
	\end{minipage}%% 
	\vspace{-1mm}
	\caption*{\textmd{(c) Line Control $ N=4 $}}
	%%%%%%%%%%%%%%%%%%%%%%%%%%%%%%%%%%%
	\begin{minipage}[b]{0.20\linewidth}
		\centering
		\includegraphics[width=0.9\linewidth,fbox]{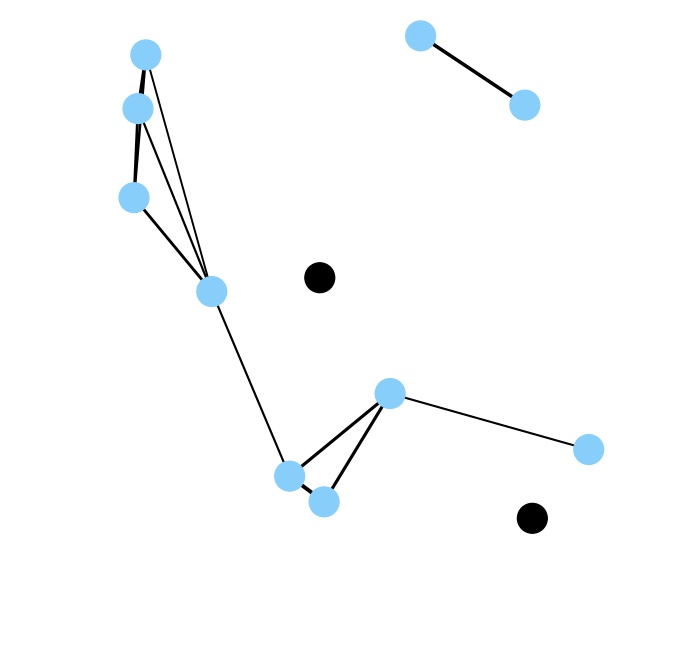} 
	\end{minipage}%% 	\begin{minipage}[b]{0.20\linewidth}
	\begin{minipage}[b]{0.20\linewidth}
		\centering
		\includegraphics[width=0.9\linewidth,fbox]{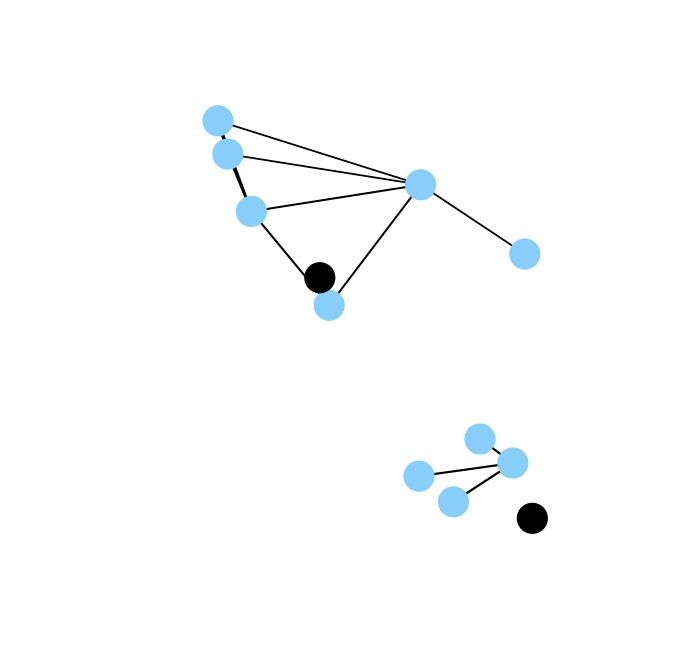} 
	\end{minipage}%% 
	\begin{minipage}[b]{0.20\linewidth}
		\centering
		\includegraphics[width=0.9\linewidth,fbox]{img_ga_simple_line_po10_episode4_15} 
	\end{minipage}%% 
	\begin{minipage}[b]{0.20\linewidth}
		\centering
		\includegraphics[width=0.9\linewidth,fbox]{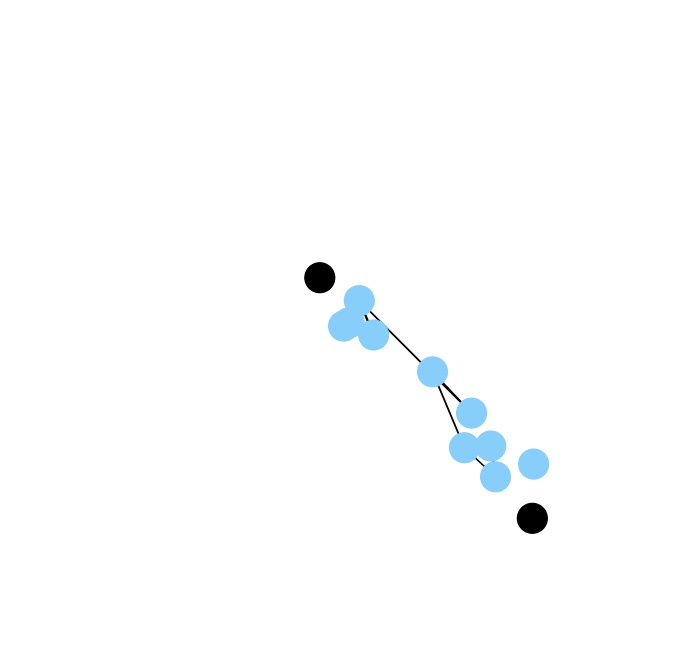} 
	\end{minipage}%% 
	\begin{minipage}[b]{0.20\linewidth}
		\centering
		\includegraphics[width=0.9\linewidth,fbox]{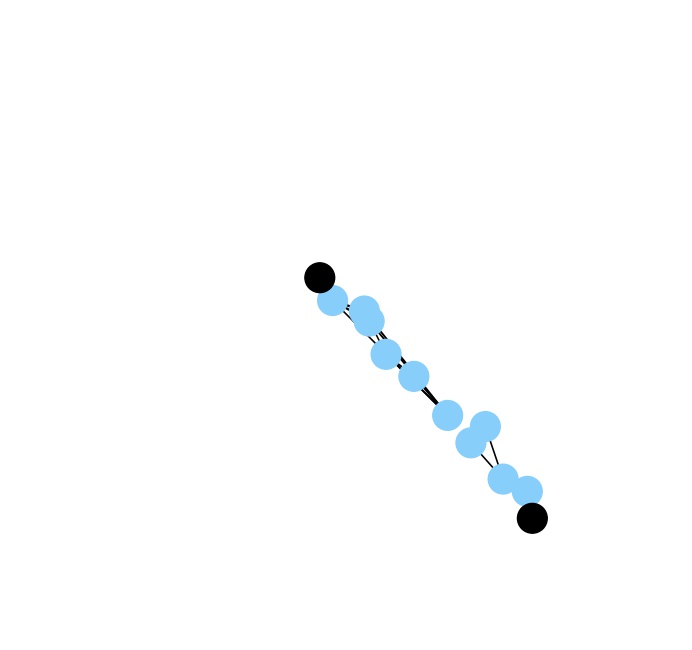} 
	\end{minipage}%% 
	\vspace{-1mm}
	\caption*{\textmd{(d) Line Control $ N=10 $}}
\end{minipage}
\vspace{-1.5mm}
\caption{\textmd{{\color{black}Examples of communication networks $ G^t $ evolving over different episode time-steps on Navigation Control and Line Control. Black circles represent landmarks; agents are represented in blue. Connections indicate the heat kernel connectivity weights generated by CDC.}}}
\label{fig:hkenvironments1} 
\end{figure}

\begin{figure}[H] 
\begin{minipage}[b]{0.20\linewidth}
	\centering
	\includegraphics[width=0.9\linewidth,fbox]{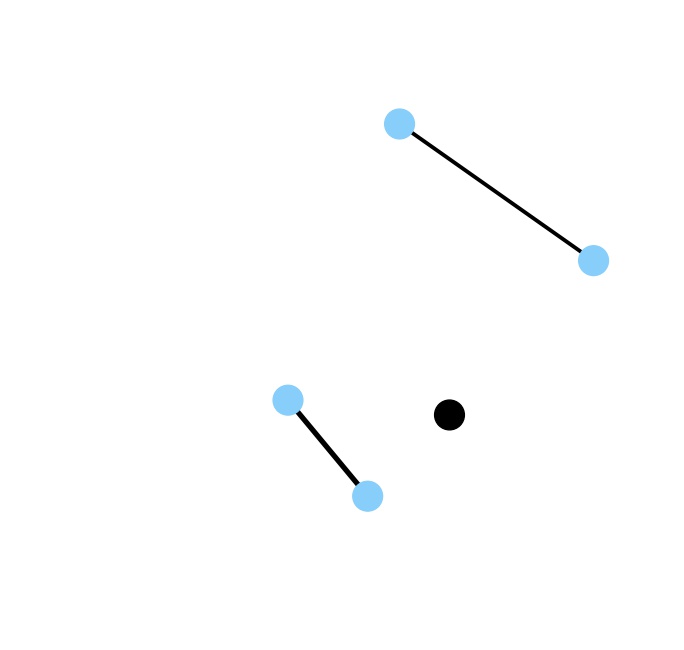} 
\end{minipage}%% 	\begin{minipage}[b]{0.20\linewidth}
\begin{minipage}[b]{0.20\linewidth}
	\centering
	\includegraphics[width=0.9\linewidth,fbox]{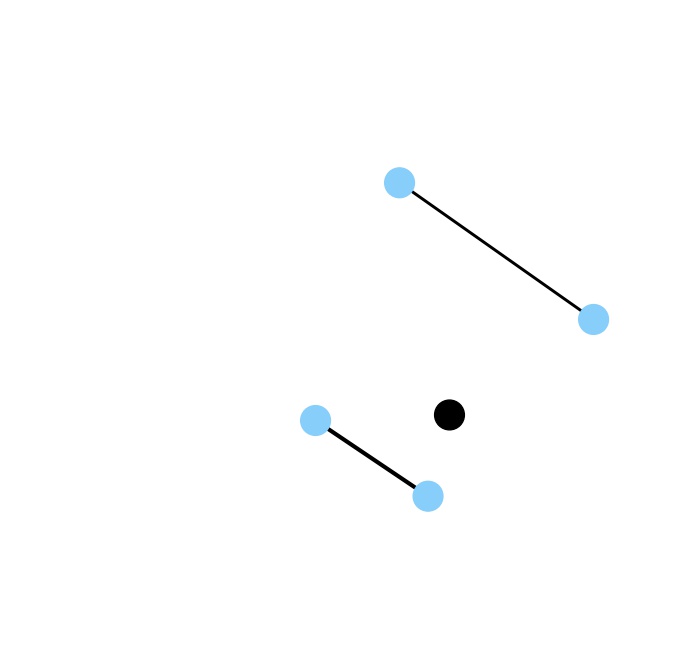} 
\end{minipage}%% 
\begin{minipage}[b]{0.20\linewidth}
	\centering
	\includegraphics[width=0.9\linewidth,fbox]{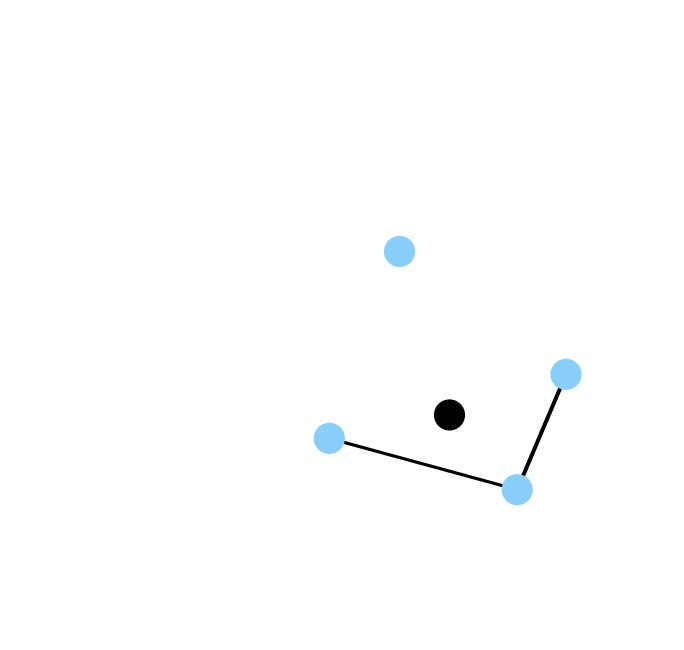} 
\end{minipage}%% 
\begin{minipage}[b]{0.20\linewidth}
	\centering
	\includegraphics[width=0.9\linewidth,fbox]{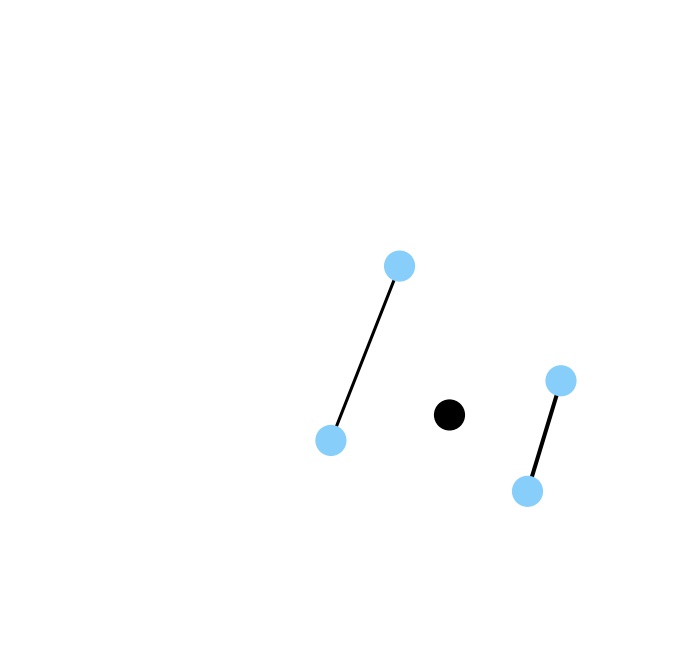} 
\end{minipage}%% 
\begin{minipage}[b]{0.20\linewidth}
	\centering
	\includegraphics[width=0.9\linewidth,fbox]{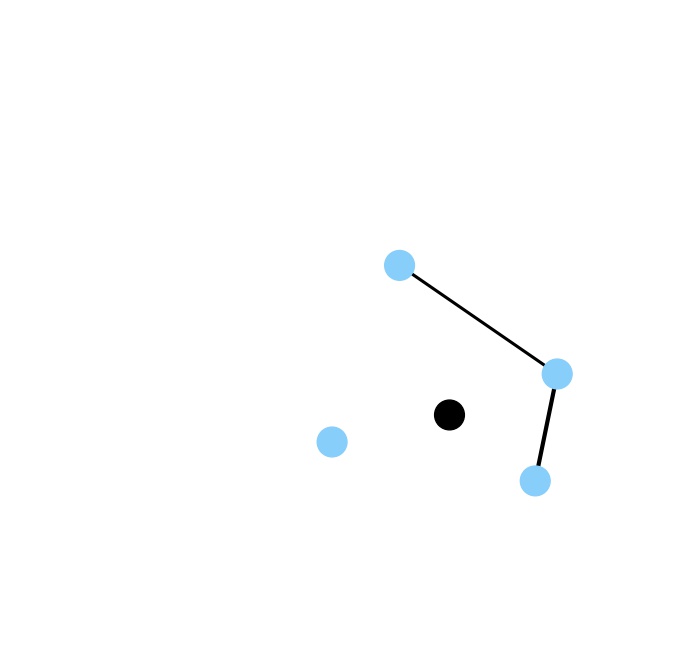} 
\end{minipage}%% 
\vspace{-1mm}
\caption*{\textmd{(a) Formation Control $ N=4 $}}
\begin{minipage}[b]{0.20\linewidth}
	\centering
	\includegraphics[width=0.9\linewidth,fbox]{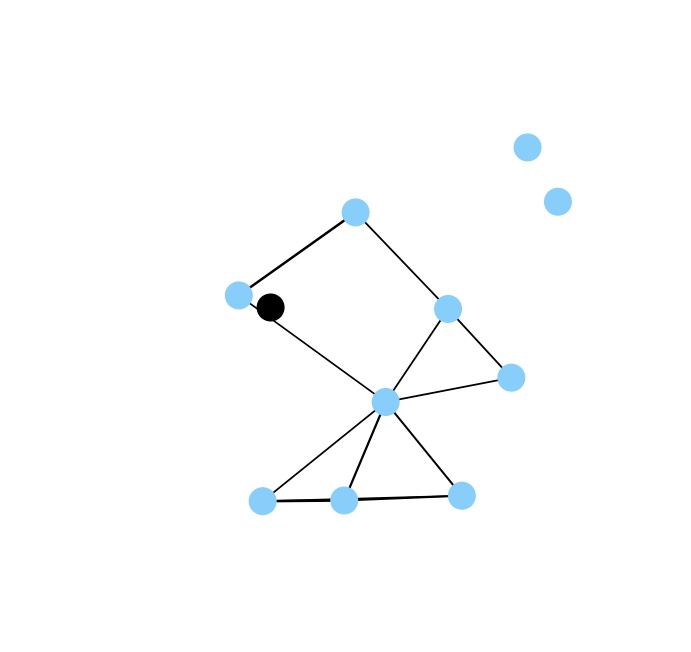} 
\end{minipage}%% 	\begin{minipage}[b]{0.20\linewidth}
\begin{minipage}[b]{0.20\linewidth}
	\centering
	\includegraphics[width=0.9\linewidth,fbox]{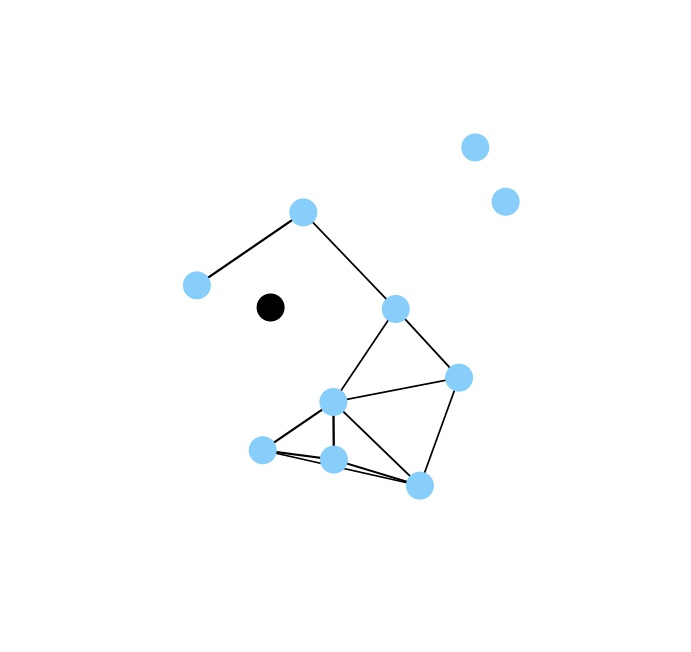} 
\end{minipage}%% 
\begin{minipage}[b]{0.20\linewidth}
	\centering
	\includegraphics[width=0.9\linewidth,fbox]{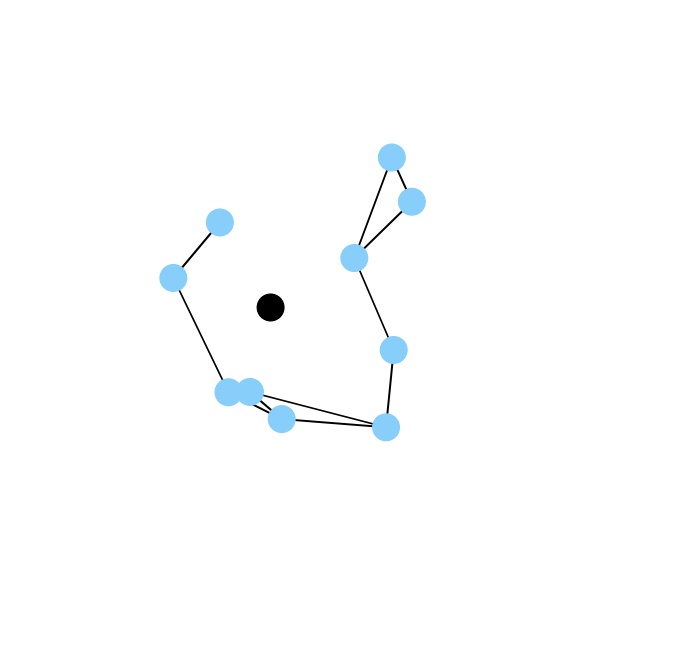} 
\end{minipage}%% 
\begin{minipage}[b]{0.20\linewidth}
	\centering
	\includegraphics[width=0.9\linewidth,fbox]{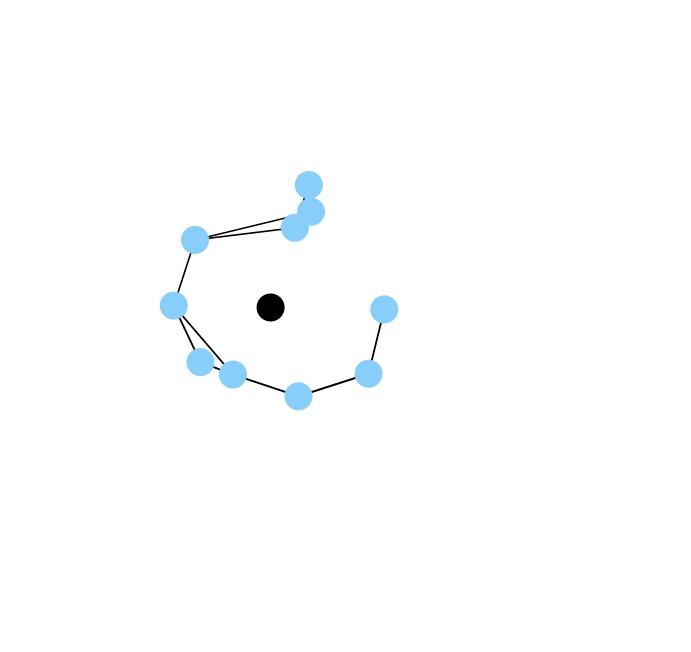} 
\end{minipage}%% 
\begin{minipage}[b]{0.20\linewidth}
	\centering
	\includegraphics[width=0.9\linewidth,fbox]{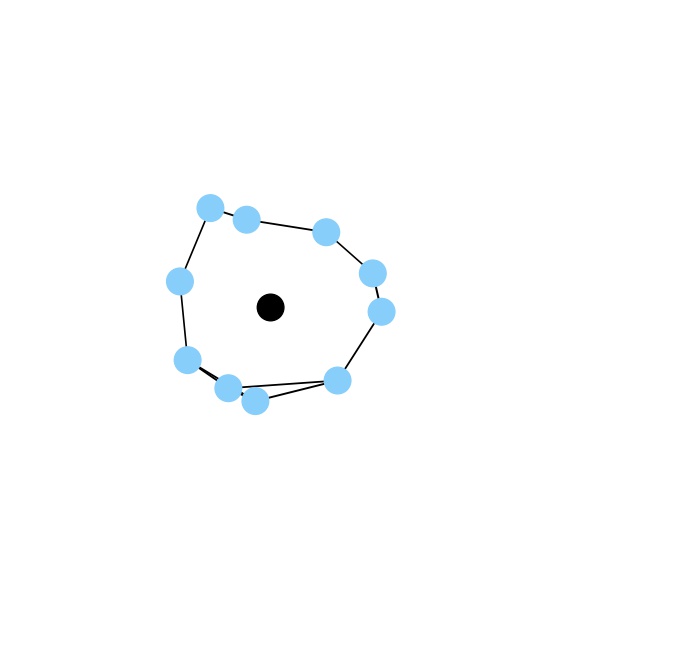} 
\end{minipage}%% 
\vspace{-1mm}
\caption*{\textmd{(b) Formation Control $ N=10 $}}
%%%%%%%%%%%%%%%%%%%%%%%%%%%%%%%%%%%
\begin{minipage}[b]{0.20\linewidth}
	\centering
	\includegraphics[width=0.9\linewidth,fbox]{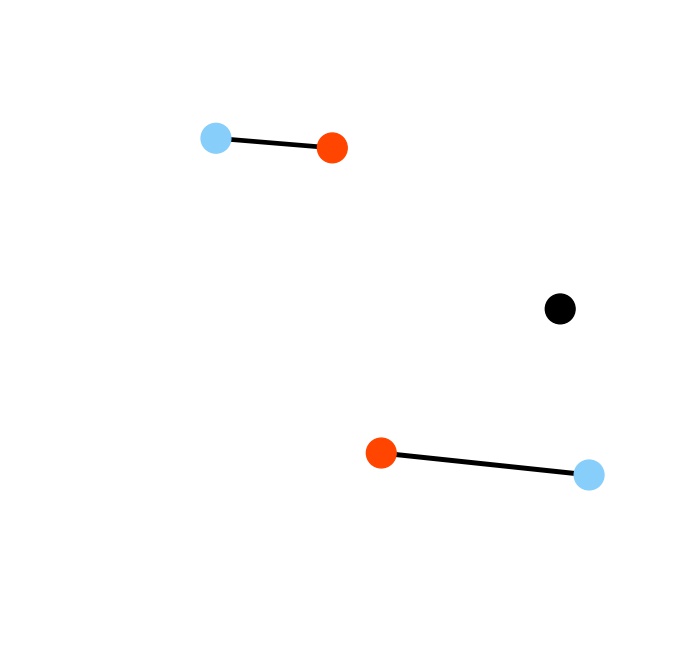} 
\end{minipage}%% 	\begin{minipage}[b]{0.20\linewidth}
\begin{minipage}[b]{0.20\linewidth}
	\centering
	\includegraphics[width=0.9\linewidth,fbox]{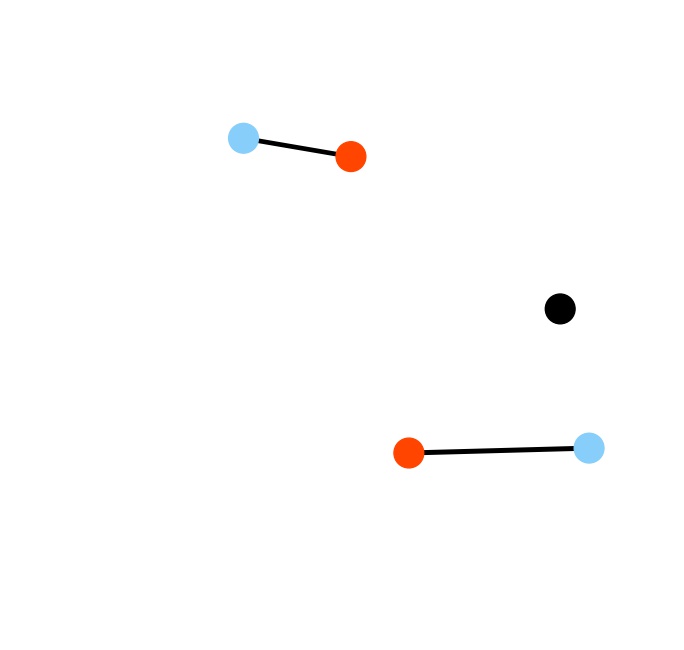} 
\end{minipage}%% 
\begin{minipage}[b]{0.20\linewidth}
	\centering
	\includegraphics[width=0.9\linewidth,fbox]{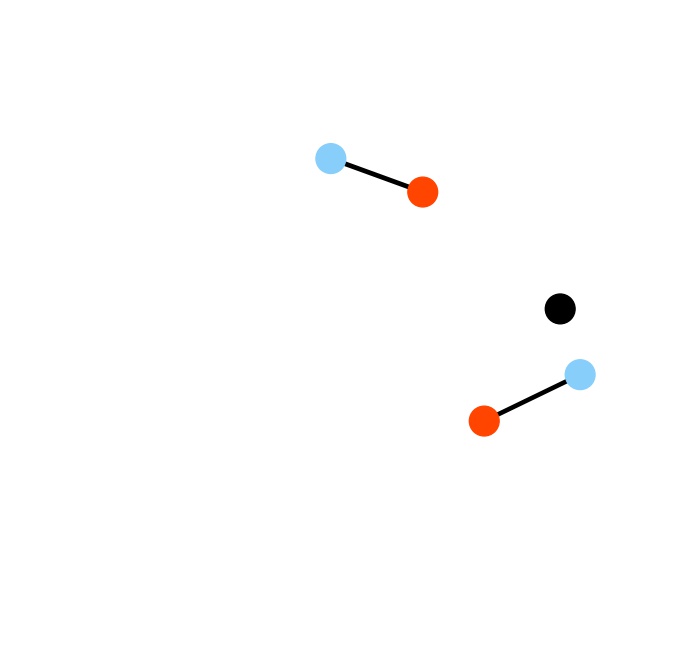} 
\end{minipage}%% 
\begin{minipage}[b]{0.20\linewidth}
	\centering
	\includegraphics[width=0.9\linewidth,fbox]{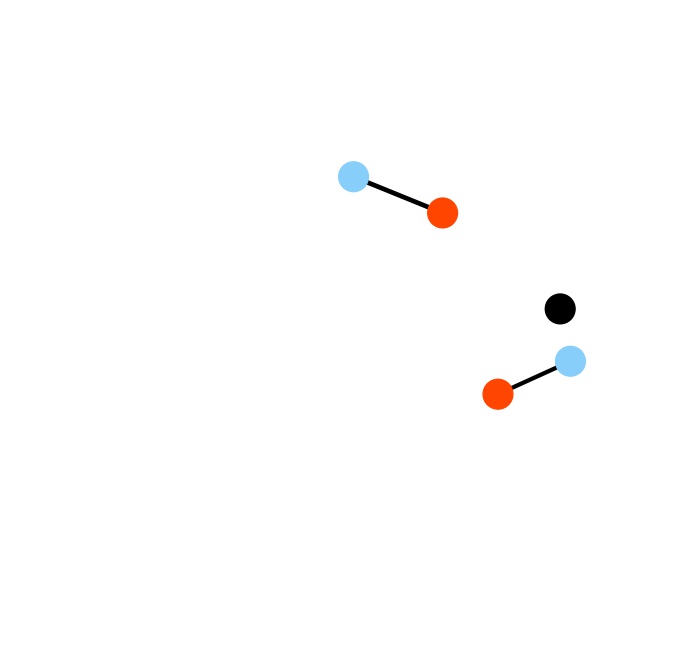} 
\end{minipage}%% 
\begin{minipage}[b]{0.20\linewidth}
	\centering
	\includegraphics[width=0.9\linewidth,fbox]{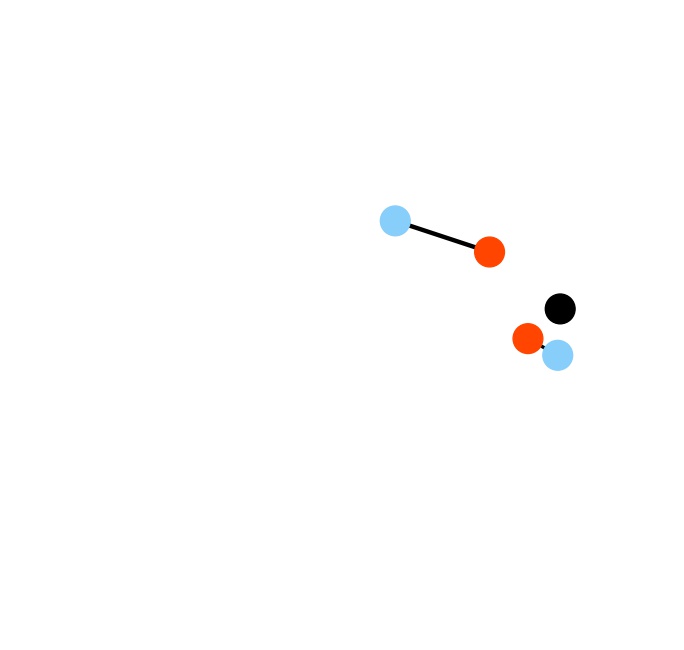} 
\end{minipage}%%
\vspace{-1mm}
\caption*{\textmd{(c) Dynamical Pack Control $ N=4 $}}
%%%%%%%%%%%%%%%%%%%%%%%%%%%%%%%%%%%
\begin{minipage}[b]{0.20\linewidth}
	\centering
	\includegraphics[width=0.9\linewidth,fbox]{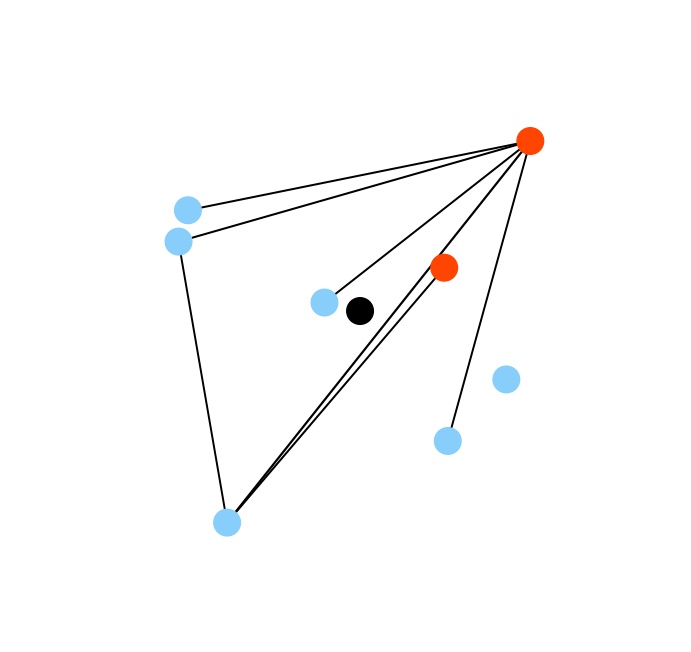} 
\end{minipage}%% 	\begin{minipage}[b]{0.20\linewidth}
\begin{minipage}[b]{0.20\linewidth}
	\centering
	\includegraphics[width=0.9\linewidth,fbox]{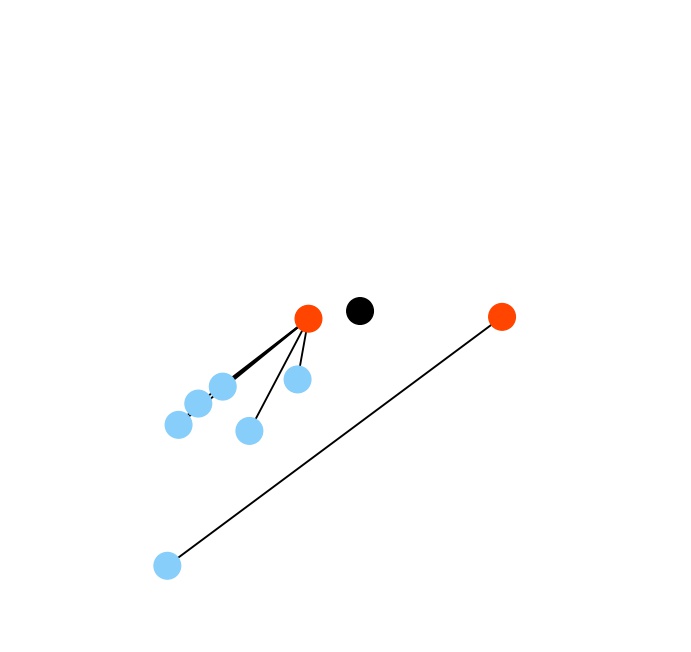} 
\end{minipage}%% 
\begin{minipage}[b]{0.20\linewidth}
	\centering
	\includegraphics[width=0.9\linewidth,fbox]{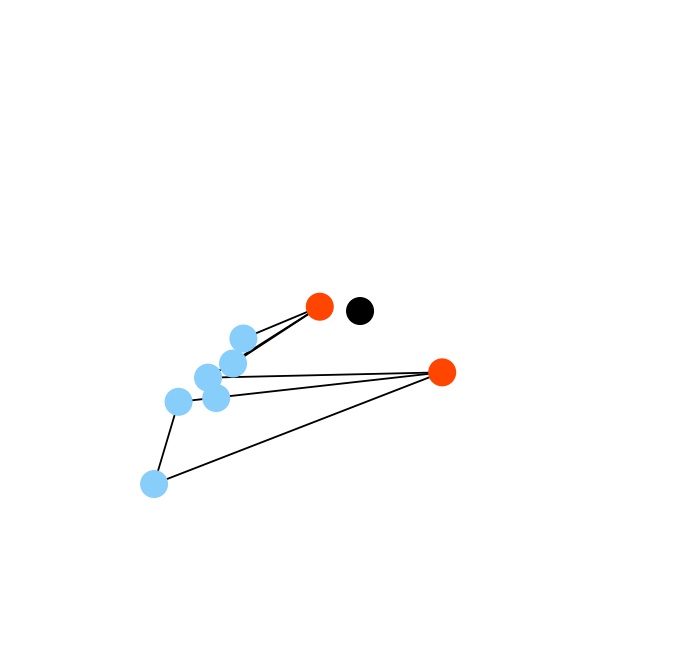} 
\end{minipage}%% 
\begin{minipage}[b]{0.20\linewidth}
	\centering
	\includegraphics[width=0.9\linewidth,fbox]{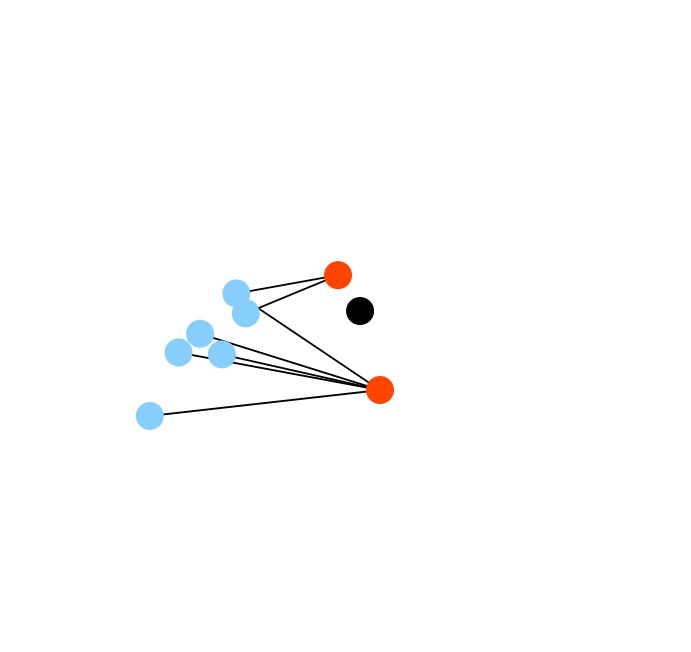} 
\end{minipage}%% 
\begin{minipage}[b]{0.20\linewidth}
	\centering
	\includegraphics[width=0.9\linewidth,fbox]{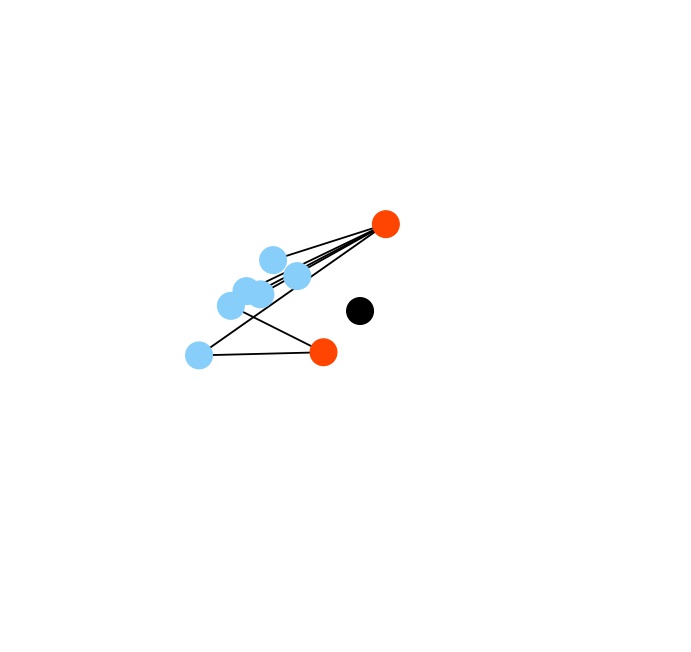} 
\end{minipage}%%
\vspace{-1mm}
\caption*{\textmd{(d) Dynamical Pack Control $ N=8 $}}
\caption{\textmd{{\color{black}Examples of communication networks $ G^t $ evolving over different episode time-steps on Formation Control and Dynamic Pack Control. Black circles describe landmarks; agents are represented in blue, leader agents in red. Connections indicate the heat kernel connectivity weights generated by CDC.}}}
\label{fig:hkenvironments2} 
\end{figure}

In this section, we provide a qualitative evaluation of the communication patterns and associated topological structures that have emerged using CDC on the four environments. {\color{black}Figure \ref{fig:hkenvironments1} and \ref{fig:hkenvironments2}} show the communication networks $ G^t_H $ evolving over time at a given episode during execution: black circles represent the landmarks,  blue circles indicate the normal agents, and the red circles are the leaders. Their coordinates within the two-dimensional area indicate the navigation trajectories. The lines connecting pairs of agents represent the time-varying edge weights, $\bm{H}^t$. Each $ H^t_{u,v} $ element quantifies the amount of diffused heat between the two nodes.
%{\color{black}Figure \ref{fig:hkenvironments1} and \ref{fig:hkenvironments2}} illustrate how those quantities evolve over time and which agents are involved in the communication at any given time.
% end color_red

{\color{black} As expected, different patterns emerge in different environments; see Figure \ref{fig:hkenvironments1} and Figure \ref{fig:hkenvironments2}. For instance, in Formation Control, the dynamic graphs are dense in the early stages of the episodes, and become sparser later on when the formation is found. } The degree of topological adjustment observed over time indicate initial bursts of communication activity at the beginning of an episode; towards the end the communication, this seems to have stabilised and consists of messages shared only across neighbours, which seems to be sufficient to maintain the polygonal shape. A different situation can be observed in  Dynamic Pack Control; see Figure \ref{fig:hkenvironments2}(f). Here,  there is an intense communication activity between leaders and members at an early stage, and the emerging topology approximates a bipartite graph between red and blue nodes. This is an expected and plausible pattern, given the nature of this environment; the leaders need to share information with the members, which otherwise would not know be able to locate the landmarks. 

{\color{black}
In addition to the above qualitative interpretation based on graph topologies, we can also quantify the emergence of different communication patterns by looking at changes in the statistics of the degree centrality (i.e. the number of connections of each agent) over time. Specifically, we compare the statistics attained at the beginning and end of an episode using the connectivity graph generated by CDC. Table \ref{tab:degree_centrality} shows the mean and variance of the centrality degree, across all nodes, for each environment. Changes in variance, for instance, may indicate the formation of clusters. Here it can be noted that in Navigation Control, Line Control and Formation Control, the variance is significantly lower at the end of the episodes; this is expected since the best strategy in such tasks consists of spreading the number of connections across all nodes. A different pattern emerges in Dynamic Pack Control where the formation of clusters is necessary since the workers need to connect with the leaders. These clusters are also visible in Figure \ref{fig:hkenvironments2} (f).

\begin{table}[h!]
	\begin{center}
		\scalebox{0.9}{
			{\color{black}
				\begin{tabular}{l|l|l}
					\multicolumn{3}{c}{ Average Degree Centrality} \\
					Environment & Beginning of episode   & End of episode   \\
					\hline			
					Navigation Control $ N = 10 $    & $ 1.7\pm(1.5) $     & $ 2.4\pm(0.5) $   \\
					Line Control $ N = 10 $    & $ 2.5\pm(0.9) $     & $ 1.8\pm(0.4) $    \\
					Formation Control $ N = 10 $    & $ 2.2 \pm (1.7) $   & $ 2.1 \pm (0.3) $  \\
					Dynamic Pack Control $ N = 8 $    & $ 1.4\pm(0.91) $  & $ 1.6\pm(1.4) $ \\
					\hline
				\end{tabular}
			}
		}
		\caption{\textmd{{\color{black}Mean and standard deviation for the centrality degree calculated using the connectivity graphs generated by CDC. Metrics are calculated utilising the graph produced in the first (beginning) and last step (end) of the episodes at execution time.}}}
		\label{tab:degree_centrality}
	\end{center}
\end{table}
} % end color_red

%	\begin{figure}[t!]
%		%\begin{figure}[t!] 
%		%%%%%%%%%%%%%%%%%%%%%%%
%		%%%%%%%%%%%%%%%%%%%%%%%
%		\begin{minipage}[b]{0.5\linewidth}
%			\centering
%			\includegraphics[width=0.8\linewidth]{img_avg_simple_formation_po10_episode2_layout_zoom.jpg} 
%		\end{minipage}%% 
%		\begin{minipage}[b]{0.5 \linewidth}
%			\centering
%			\includegraphics[width=0.7\linewidth]{img_avg_simple_formation_po10_episode2_hkhp.jpg} 
%		\end{minipage} 
%		\vspace{-5mm}
%		\caption*{\textmd{(a) Formation Control}}
%		\vspace{3mm}
%		%%%%%%%%%%%%%%%%%%%%%%%
%		\begin{minipage}[b]{0.5\linewidth}
%			\centering
%			\includegraphics[width=0.8\linewidth]{img_avg_simple_spread_pack_leader_8_2_episode0_layout_zoom.jpg} 
%		\end{minipage}%% 
%		\begin{minipage}[b]{0.5 \linewidth}
%			\centering
%			\includegraphics[width=0.7\linewidth]{img_avg_simple_spread_pack_leader_8_2_episode0_hkhp.jpg} 
%		\end{minipage} 
%		\vspace{-5mm}
%		\caption*{\textmd{(b) Dynamic Pack Control}}
%		%	%%%%%%%%%%%%%%%%%%%%%%%	
%		%	\caption{\textmd{Averaged communication graphs for Formation Control (line above) and Pack Control (line below).}}
%		\caption{\textmd{Averaged communication graphs for (a) Formation Control and (b) Dynamic Pack Control. On the left, node sizes indicate the eigenvector centrality, where the connections the stable heat kernel values, while numbers the node labels. On the right, the heat kernel values are shown as heatmaps, where axis numbers correspond to node labels.}}
%		\label{fig:graphheatmap} 
%		%\end{figure}
%	\end{figure}
\begin{figure}[H] 
\begin{minipage}[b]{0.23\linewidth}
	\centering
	\includegraphics[width=1\linewidth]{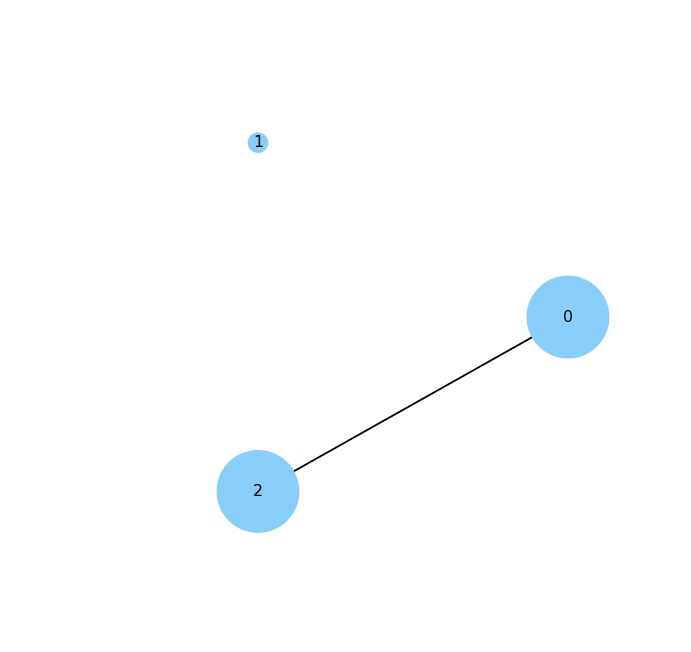} 
\end{minipage}%% 
\begin{minipage}[b]{0.23 \linewidth}
	\centering
	\includegraphics[width=1\linewidth]{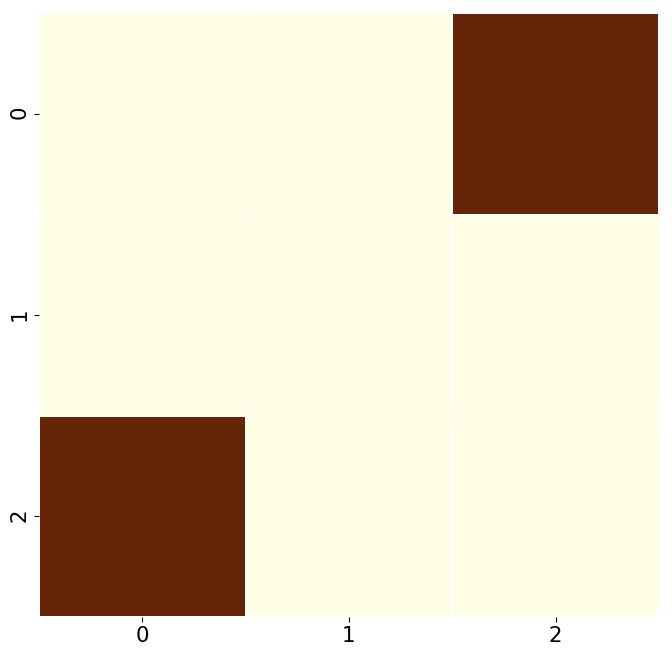} 
\end{minipage} 
\hspace{1cm}
\begin{minipage}[b]{0.23\linewidth}
	\centering
	\includegraphics[width=1\linewidth]{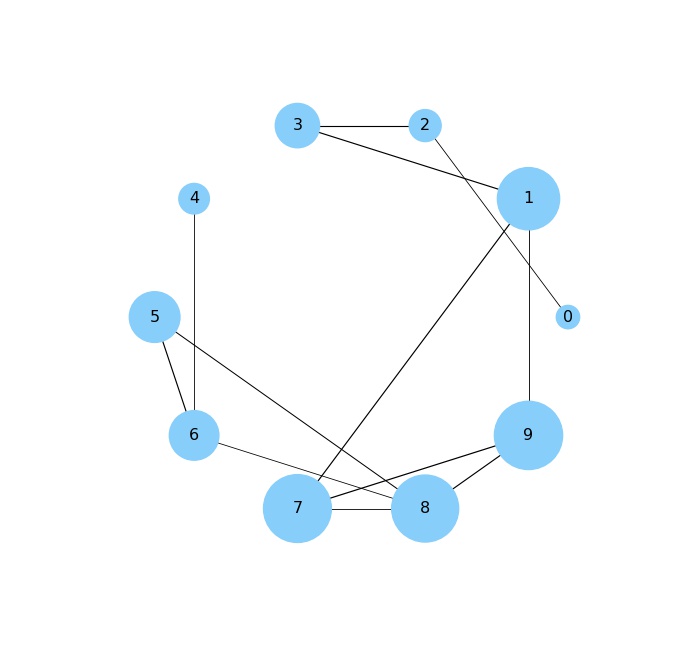} 
\end{minipage}%% 
\begin{minipage}[b]{0.23 \linewidth}
	\centering
	\includegraphics[width=1\linewidth]{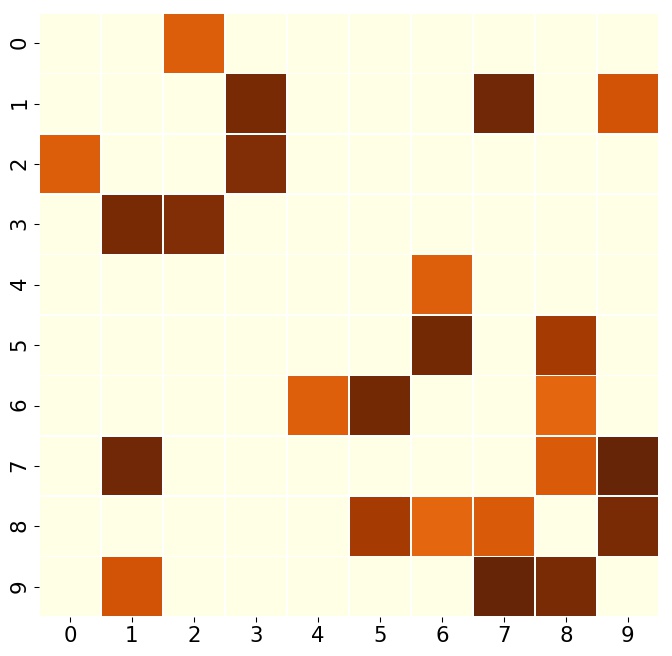} 
\end{minipage} 
\caption*{\hspace{0.5cm}\textmd{(a) Navigation Control $ N = 3$} \hspace{1.5cm}\textmd{(b) Navigation Control $ N = 10$}}
\vspace{5mm}
%%%%%%%%%%%%%%%%%%%%%%%
%%%%%%%%%%%%%%%%%%%%%%%
\begin{minipage}[b]{0.23\linewidth}
	\centering
	\includegraphics[width=1\linewidth]{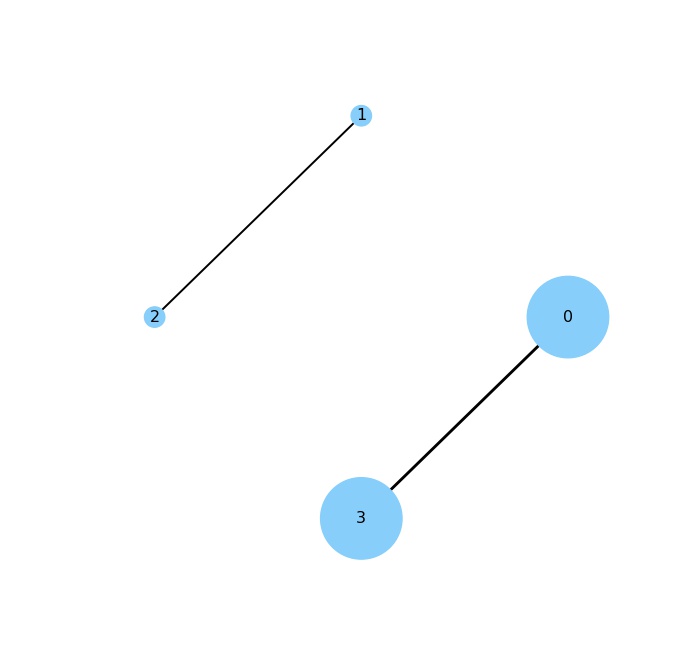} 
\end{minipage}%% 
\begin{minipage}[b]{0.23 \linewidth}
	\centering
	\includegraphics[width=1\linewidth]{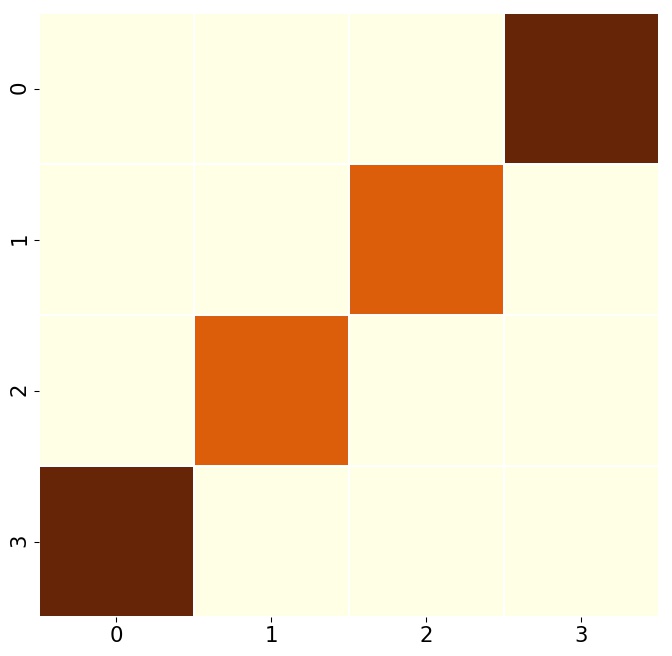} 
\end{minipage} 
\hspace{1cm}
\begin{minipage}[b]{0.23\linewidth}
	\centering
	\includegraphics[width=1\linewidth]{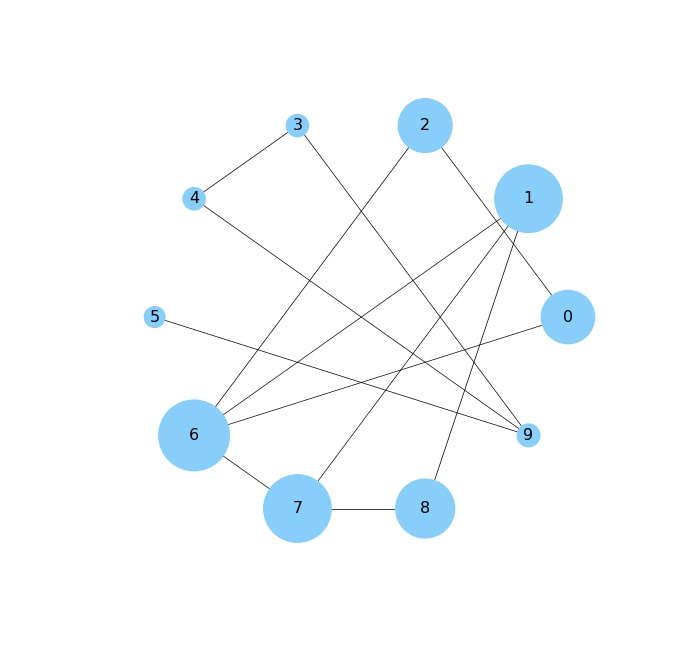} 
\end{minipage}%% 
\begin{minipage}[b]{0.23 \linewidth}
	\centering
	\includegraphics[width=1\linewidth]{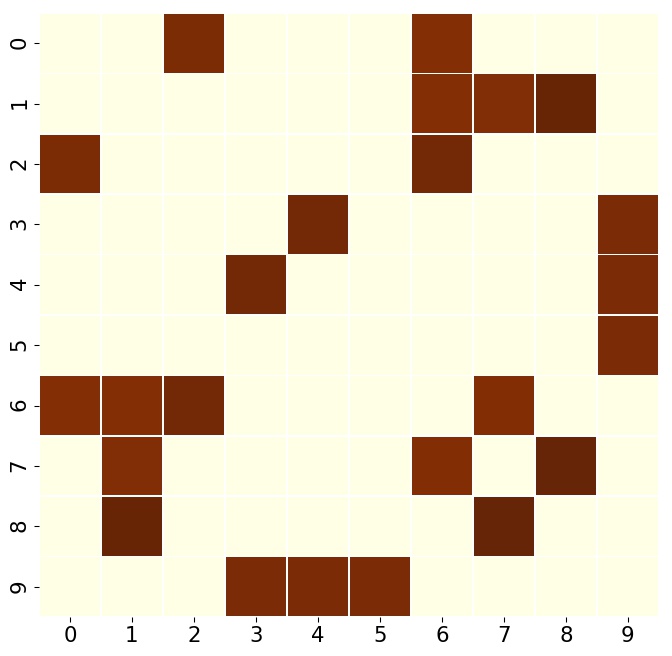} 
\end{minipage} 
\caption*{\hspace{0.5cm}\textmd{(c) Line Control $ N = 4$} \hspace{3cm}\textmd{(d) Line Control $ N = 10$}}
\vspace{3mm}
%%%%%%%%%%%%%%%%%%%%%%%\\
%%%%%%%%%%%%%%%%%%%%%%%
\begin{minipage}[b]{0.23\linewidth}
	\centering
	\includegraphics[width=1\linewidth]{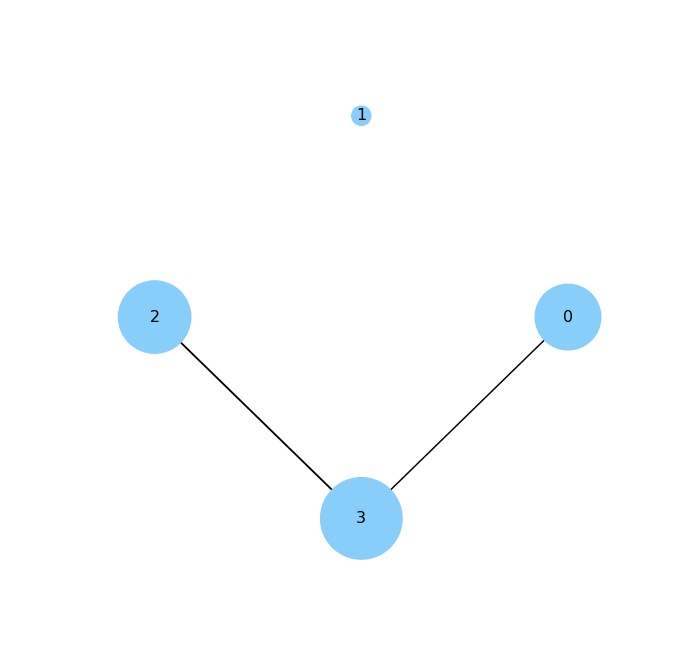} 
\end{minipage}%% 
\begin{minipage}[b]{0.23 \linewidth}
	\centering
	\includegraphics[width=1\linewidth]{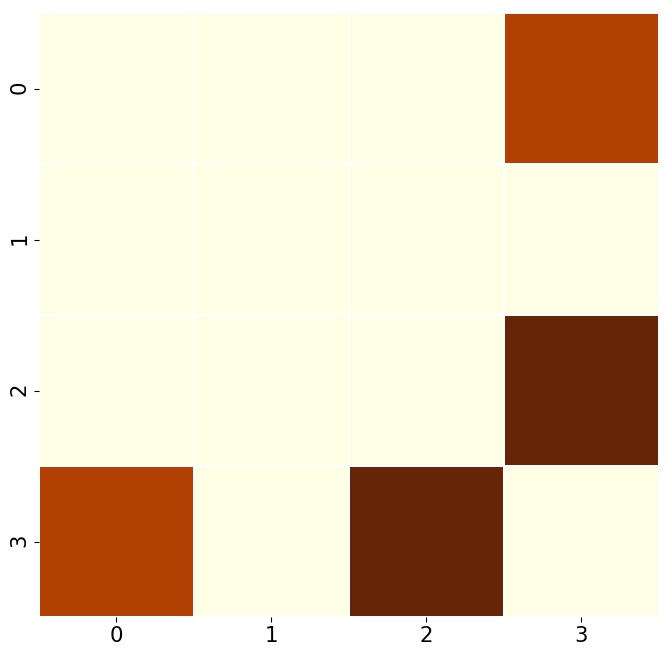} 
\end{minipage} 
\hspace{1cm}
\begin{minipage}[b]{0.23\linewidth}
	\centering
	\includegraphics[width=1\linewidth]{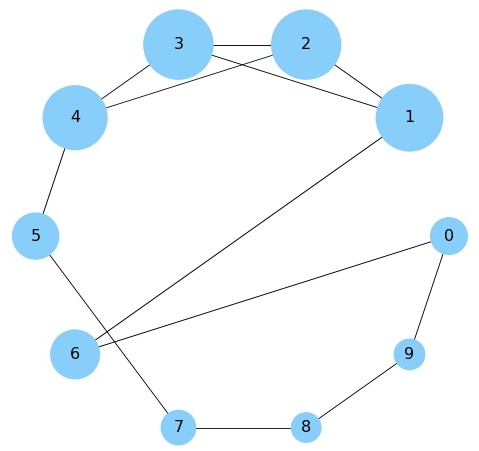} 
\end{minipage}%% 
\begin{minipage}[b]{0.23 \linewidth}
	\centering
	\includegraphics[width=1\linewidth]{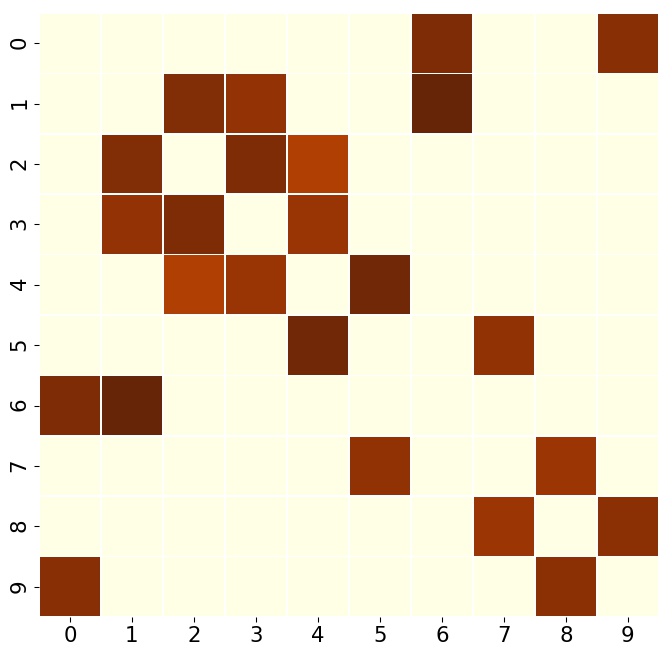} 
\end{minipage} 
\caption*{\hspace{0.5cm}\textmd{(e) Formation Control $ N = 4$} \hspace{2cm}\textmd{(f) Formation Control $ N = 10$}}
\vspace{3mm}
%%%%%%%%%%%%%%%%%%%%%%%\\
%%%%%%%%%%%%%%%%%%%%%%%\\
\begin{minipage}[b]{0.23\linewidth}
	\centering
	\includegraphics[width=1\linewidth]{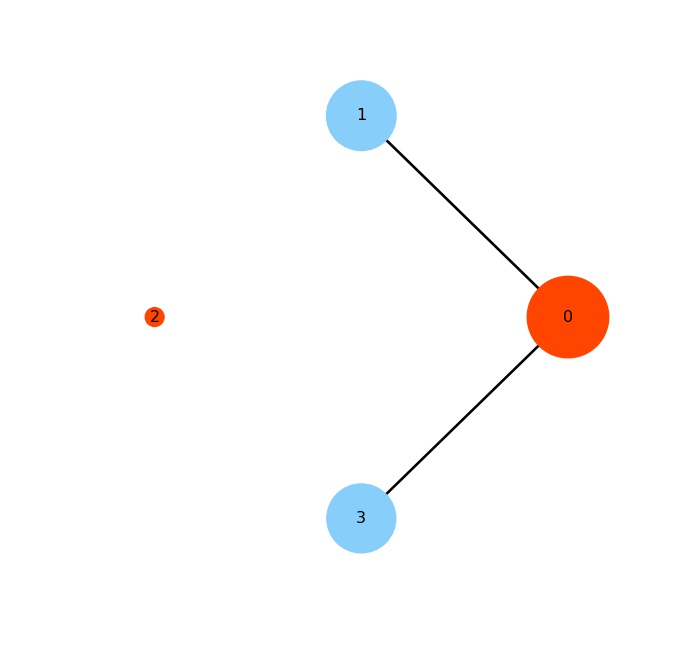} 
\end{minipage}%% 
\begin{minipage}[b]{0.23 \linewidth}
	\centering
	\includegraphics[width=1\linewidth]{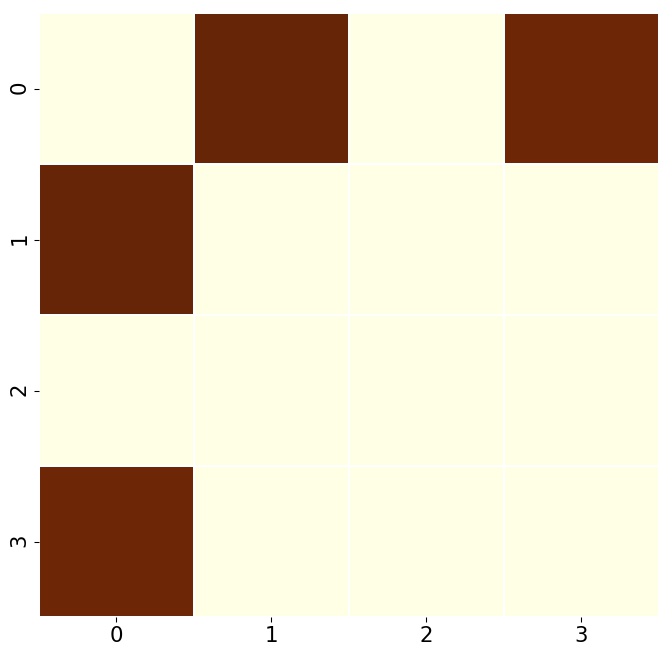} 
\end{minipage} 
\hspace{1cm}
\begin{minipage}[b]{0.23\linewidth}
	\centering
	\includegraphics[width=1\linewidth]{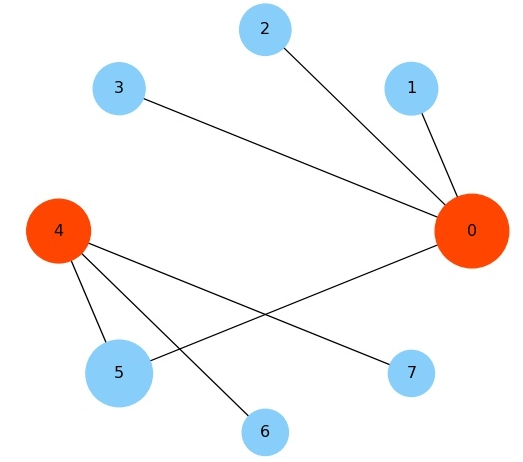} 
\end{minipage}%% 
\begin{minipage}[b]{0.23 \linewidth}
	\centering
	\includegraphics[width=1\linewidth]{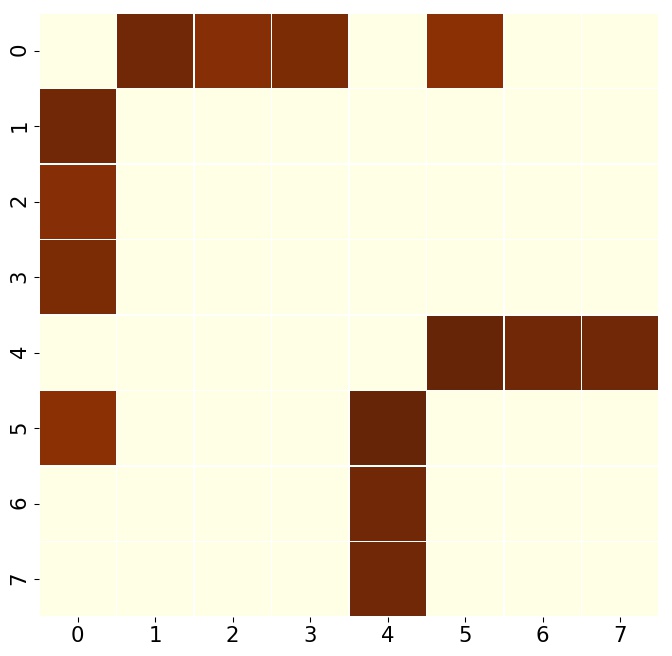} 
\end{minipage} 
\caption*{\hspace{0.5cm}\textmd{(e) Dynamic Pack Control $ N = 4$} \hspace{1cm}\textmd{(f) Dynamic Pack Control $ N = 8$}}
\vspace{3mm}
%%%%%%%%%%%%%%%%%%%%%%%\\
%	\caption{\textmd{{\color{black}Communication graphs averaged over an episode. For each environment, on the left, node sizes indicate the eigenvector centrality, connections the stable heat kernel values, while numbers the node labels. Here, a circular layout is used to represent the graphs in order to provide an alternative view where connection patterns can result easier to detect. On the right, diffused values are shown as heatmaps, where axis numbers correspond to node labels.}}}
%	\caption{\textmd{Averaged communication graphs for Formation Control (line above) and Pack Control (line below).}}
\caption{\textmd{{\color{black}Averaged communication graphs for all the environments. On the left side of each figure, the node sizes describe the eigenvector centrality, the connections represent the heat kernel values and the numbers indicate the node labels. On the right, the heat kernel values are shown as heatmaps, where axis numbers correspond to node labels.}}}
\label{fig:graphheatmap} 
\end{figure}

Further appreciation for the role played by the heat kernel in driving the communication strategy can be gained by observing Figure \ref{fig:graphheatmap} which provides visualisations {\color{black}for all the environments}. On the left, the connection weights are visualised using a circular layout. Here the nodes represent agents, and the size of each node is proportional to the node's eigenvector centrality. The eigenvector centrality is a popular graph spectral measure \cite{bonacich2007some}, utilised to determine the influence of a node considering both its adjacent connections and the importance of its neighbouring node.
This measure is calculated using the stable heat diffused values averaged over an episode, i.e. $ H_{u,v} = (\sum^T_{t=1} H^t_{u,v})/T $. The resulting graph structure reflects the overall communication patterns emerged while solving the given tasks. On the right, we visualise the squared $N \times N$ matrix of averaged pairwise diffusion values as a heatmap (red values are higher). It can be noted that, in  Pack Control, two communities of agents are formed, each one with a leader. Here, as expected, leaders appear to be influential nodes (red nodes), and the heatmap shows that the connections between individual members and leaders are very strong. A different pattern emerges instead in Formation Control, where there is no evidence of communities since all nodes are connected to nearly form a circular shape. The corresponding heatmap shows the heat kernel values connecting neighbouring agents tend to assume higher values compared to more distant agents.

{\color{black}
\subsection{Ablation studies}

We have carried out a number of studies to assess the relative importance of each new component contributing to CDC. First, we investigate the relative merits of the heat kernel over two alternative and simpler information propagation mechanisms: (a) a {\it global average} approach, where the observations of all other agents are averaged and provided to the agent to inform its action, and (b) the {\it nearest neighbours} approach, where only the observations of the agent's two nearest neighbours are averaged. For each one of these two mechanisms, we compare a version using our proposed critic (Section \ref{sec:learning}), which uses a recurrent architecture (specifically an LSTM), and a version using a traditional critic, i.e. based on a feed-forward neural network. To better characterise the benefits of a recurrent network, we have also investigated an LSTM-based version of MADDPG. In addition, we have implemented a version of CDC that use a {\it softmax attention}, i.e. the heat kernel connectivity weights have been replaced by a softmax function.  To ensure a fair comparison, only the necessary architectural changes have been carried out in order to keep the modelling capacity across different versions comparable.   
%	For instance, in the nearest neighbours, Softmax and average versions of CDC the main difference consists in the fact that only the message aggregation module differs from the original model.

\begin{figure*}[h!] 
	\begin{minipage}[b]{0.9\linewidth}
		\centering
		\includegraphics[width=1\linewidth]{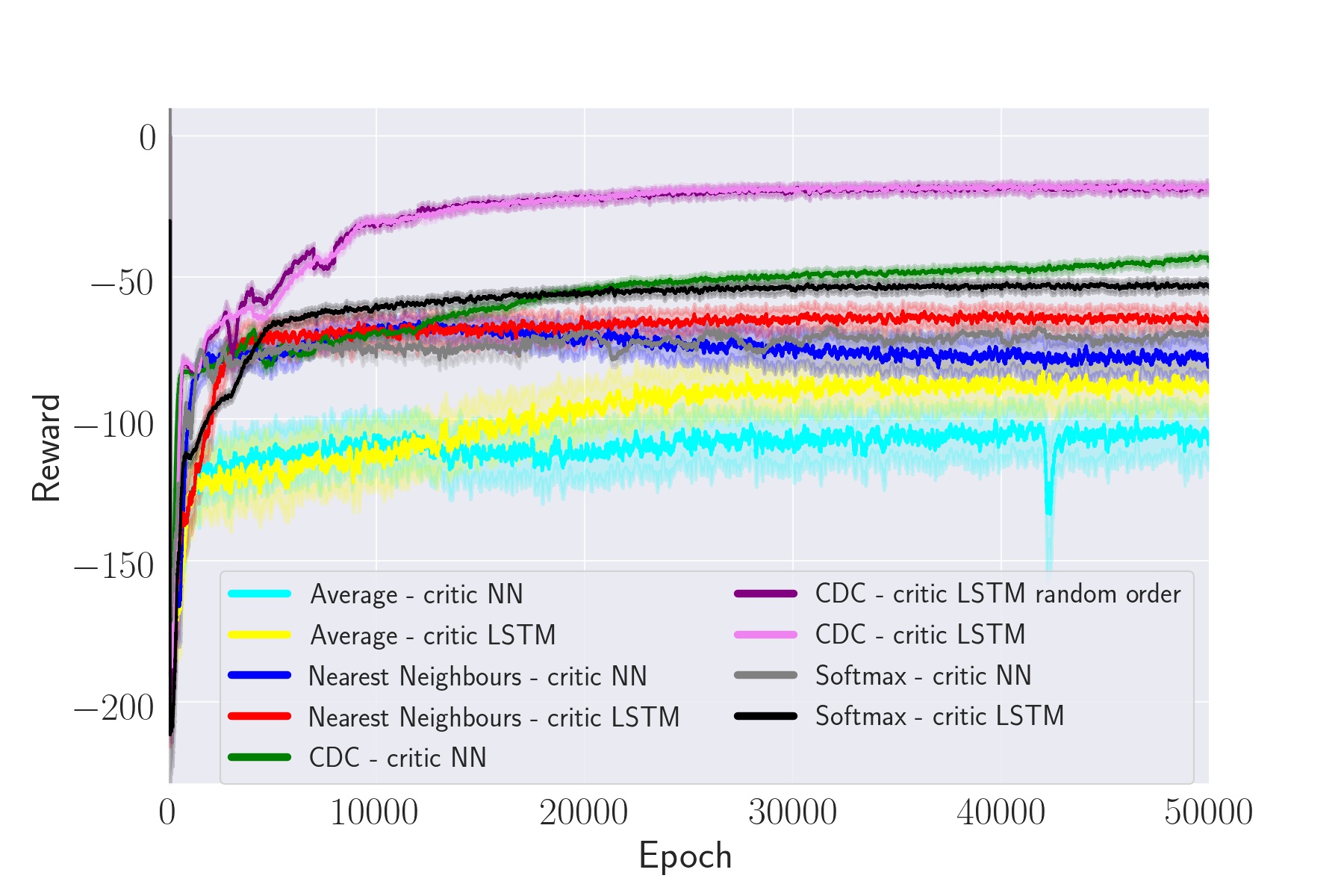} 
	\end{minipage}%% 
	\vspace{-2mm}
	%	\caption*{\textmd{Pack Control $ N = 4 $}}
	\caption{\textmd{{\color{black}Learning curves of different versions of the proposed model on Formation Control ($ N=4 $)}. 
	}}
	\label{fig:supp_ablationstudy} 
\end{figure*}

In Figure \ref{fig:supp_ablationstudy}, it can be noted that the proposed CDC using the heat kernel achieves the highest performance by a significant margin. The other modified versions of CDC, with and without LSTM, also outperform the simpler communication methods. There is evidence to suggest that averaging  local information coming from the nearest neighbours is a better strategy compared to using a global average; the latter cannot discard unnecessary information and results in nosier embeddings and worse communication. Overall, we have observed that the LSTM-based critic is beneficial compared to the simpler alternative. This is an expected result because, by design, the LSTM's hidden state  filters out irrelevant information content from the sequence of inputs. Another observed finding is that the order of the agents does not affect the final performance of the model. This is explained by the fact that each of LSTM-based critics observe the entire sequence of observations and actions before producing the feedback to return. 
{\color{black}Furthermore, the softmax version of CDC has been found to be less performant that the original CDC thus confirming the important role played by the heat kernel in aggregating the messages across the communication network.}

%Finally, \ref{fig:supp_ablationstudy} also shows an additional comparison where we kept the final model of CDC and replaced the HK weights with a simple softmax calculated on the connectivity weights. With this additional comparison we aim to show the heat kernel significantly improve the model performance by allowing a better aggregation of the generated messages.

%		\paragraph{Choosing the heat kernel threshold}\label{sec:supp_threshold}
In order to choose an appropriate threshold for the heat kernel equation (see Eq. \ref{eq:threshold}) we have run a set of experiments whereby we monitor how the success rate behaves using different parameter values. Table \ref{tab:supp_thresholdresults} reports on the performance of CDC on Formation Control when the threshold parameter $ s $ varies over a grid of possible values. In turn, this threshold determines whether the heat kernel values are stable or not. The best performance is obtained using $ s = 0.05 $, which is the value used in all our experiments. To select the specific thresholds reported in Table \ref{tab:supp_thresholdresults}, we tried a range of values suggested in related works \cite{chung2016characterising,xiao2005characterising}.
\begin{table}[h!]
	\begin{center}
		\scalebox{0.9}{
			{\color{black}
				\begin{tabular}{lllll}
					\textbf{Method}     & \multicolumn{3}{l}{\textbf{Formation Control $ N = 4 $}	} \\
					\hline
					& Reward & Time & Success Rate \\
					\hline			
					CDC $ s = 0.01$    & $ -4.48\pm(1.62) $     & $ 13.52\pm(9.83) $   & $ 0.93\pm(0.21)  $ \\
					CDC $ s = 0.025$    & $ -4.33\pm(1.28) $     & $ 14.01\pm(9.74) $   & $ 0.94\pm(0.24) $ \\
					CDC $ s = 0.05$    & $ -4.22 \pm (1.46) $   & $ 11.82 \pm (5.49) $ & $ 0.99 \pm (0.1) $ \\
					CDC $ s = 0.075$    & $ -4.34\pm(1.43) $     & $ 12.88\pm(9.13) $   & $ 0.95\pm(0.22) $ \\
					CDC $ s = 0.1$     & $ -4.31\pm(1.57)  $    & $ 12.52\pm(8.39) $   & $ 0.96\pm(0.2) $ \\
					\hline
				\end{tabular}
			}
		}
		\caption{\textmd{{\color{black}Comparison of CDC results using different values for threshold $s$.}}}
		\label{tab:supp_thresholdresults}
	\end{center}
\end{table}

} % end color_red

\section{Conclusions} \label{sec:conclusions}

In this work, we have presented a novel approach to deep multi-agent reinforcement learning that models agents as nodes of a state-dependent graph, and uses the overall topology of the graph to facilitate communication and cooperation. The inter-agent communication patterns are represented by a connectivity graph that is used to decide which messages should be shared with others, how often, and with whom. A key novelty of this approach is represented by the fact that the graph topology is inferred directly from observations and is utilised as an attention mechanism guiding the agents throughout the sequential decision process. Unlike other recently proposed architectures that rely on graph convolutional networks to extract features, but we make use of a graph diffusion process  to simulate how the information propagates over the communication network and is aggregated. Our experimental results on four different environments have demonstrated that, compared to other state-of-the-art baselines, CDC can achieve superior performance on navigation tasks of increasing complexity, and remarkably so when the number of agents increases. We have also found that visualising the graphs learnt by the agents can shed some light on the role played by the diffusion process in mediating the communication strategy that ultimately yields highly rewarding policies. 
{\color{black} The current LSTM-based critic could potentially be replaced by a graph neural network equipped with an attention mechanism capable of tailoring individual feedback according to the agents' needs.}

%Amongst the exising multi-agent communication models, TarMAC reaches similar performance; in this model,t the agents learn to aggregate other agents' messages through an attentional mechanism, while in our CDC this operation is done by exploiting the a graph diffusion process capable to operate on the generated graph of connectivities.

This work represents an initial attempt to leverage well-known graph-theoretical properties in the context of a multi-agent communication strategy, and paves the way for future exploration along related directions. For instance, further constraints could be imposed on the graph edges to regulate the overall communication process, e.g. using a notion of flow conservation \cite{jia2019graph}. Further investigations could be directed towards the effects of adopting a decentralised critic modelling the communication content together with the agents' state-action values to provide a richer individual feedback.

\backmatter

\bmhead{Supplementary information}

Supplementary material is provided in the Appendix as suggested by the provided template.

\section*{Statements and Declarations}

\subsection*{Funding}
GM acknowledges support from a UKRI AI Turing Acceleration Fellowship (EPSRC EP/V024868/1).

\subsection*{Conflict of interest/Competing interests}
No competing and finacial interests to disclose.

\subsection*{Ethics approval}
Not applicable.

\subsection*{Consent to participate}
The authors give their consent to participate.

\subsection*{Consent for publication}
The authors give their consent for publication.

\subsection*{Availability of data and materials}

Environments will be made available upon paper publication.
\subsection*{Availability of data and materials}
All code will be made available upon paper publication.

\subsection*{Authors' contributions}
Authors' contributions follow the authors' order convention.

\bigskip

\begin{appendices}

%\section{Section title of first appendix}\label{secA1}

%\onecolumn
%\clearpage
%\newpage
%\appendix
\section*{\fontsize{17}{17}\selectfont Appendix}

\section{Varying the number of agents}\label{sec:supp_varna}
\begin{table*}[h]
	\begin{center}
		\begin{tabular}{llll}
			%			\hline
			%			\multicolumn{4}{c}{\textbf{DYNAMIC EVALUATION}}   \\
			\hline 
			\textbf{\# agents}	 	&  \textbf{DDPG}  & \textbf{CDC} \\ 
			\hline 
			3                 &  $ 2.34\pm0.61 $  &  $ \bm{1.06\pm0.12}  $    \\ 
			4                  &  $ 3.52\pm1.67 $  &  $ \bm{1.09\pm0.1} $    \\ 
			5                  &  $ 3.90\pm 1.68  $  &  $ \bm{1.08\pm0.15} $    \\ 
			6                  &  $ 4.44 \pm 1.7 $  &  $ \bm{1.08\pm0.18} $    \\ 
			7                  &  $ 5.21\pm1.98 $  &  $ \bm{1.12\pm0.12}  $    \\ 
			8                  &  $ 6.49\pm2.17 $  &  $ \bm{1.13\pm0.11} $    \\ 
			\hline 
		\end{tabular}
		\caption{Comparison of DDPG and CDC on Dynamic Pack Control. Both algorithms were trained with 4 agents and tested with 3-8. The performance metric used here is the distance of the the farthest agent to the landmark.}
		\label{tab:supp_varna}
	\end{center}
\end{table*}

We tested whether CDC is capable of handling a different number of agents at test time. Table \ref{tab:supp_varna} shows how the performance of DDPG and CDC compares when they are both trained using 4 learners, but 3-8 agents are used at test time. We report on the maximum distance between the farthest agent and the landmark, which is invariant to the number of agents. It can be noted that CDC can handle systems with a varying number of agents, outperforming DDPG and keeping the final performance competitive with other methods that have been trained with a larger number of agents (see Table \ref{tab:mainresults}).

\end{appendices}


\bibliography{refs}   


\begin{thebibliography}{109}
\ifx \bisbn   \undefined \def \bisbn  #1{ISBN #1}\fi
\ifx \binits  \undefined \def \binits#1{#1}\fi
\ifx \bauthor  \undefined \def \bauthor#1{#1}\fi
\ifx \batitle  \undefined \def \batitle#1{#1}\fi
\ifx \bjtitle  \undefined \def \bjtitle#1{#1}\fi
\ifx \bvolume  \undefined \def \bvolume#1{\textbf{#1}}\fi
\ifx \byear  \undefined \def \byear#1{#1}\fi
\ifx \bissue  \undefined \def \bissue#1{#1}\fi
\ifx \bfpage  \undefined \def \bfpage#1{#1}\fi
\ifx \blpage  \undefined \def \blpage #1{#1}\fi
\ifx \burl  \undefined \def \burl#1{\textsf{#1}}\fi
\ifx \doiurl  \undefined \def \doiurl#1{\url{https://doi.org/#1}}\fi
\ifx \betal  \undefined \def \betal{\textit{et al.}}\fi
\ifx \binstitute  \undefined \def \binstitute#1{#1}\fi
\ifx \binstitutionaled  \undefined \def \binstitutionaled#1{#1}\fi
\ifx \bctitle  \undefined \def \bctitle#1{#1}\fi
\ifx \beditor  \undefined \def \beditor#1{#1}\fi
\ifx \bpublisher  \undefined \def \bpublisher#1{#1}\fi
\ifx \bbtitle  \undefined \def \bbtitle#1{#1}\fi
\ifx \bedition  \undefined \def \bedition#1{#1}\fi
\ifx \bseriesno  \undefined \def \bseriesno#1{#1}\fi
\ifx \blocation  \undefined \def \blocation#1{#1}\fi
\ifx \bsertitle  \undefined \def \bsertitle#1{#1}\fi
\ifx \bsnm \undefined \def \bsnm#1{#1}\fi
\ifx \bsuffix \undefined \def \bsuffix#1{#1}\fi
\ifx \bparticle \undefined \def \bparticle#1{#1}\fi
\ifx \barticle \undefined \def \barticle#1{#1}\fi
\bibcommenthead
\ifx \bconfdate \undefined \def \bconfdate #1{#1}\fi
\ifx \botherref \undefined \def \botherref #1{#1}\fi
\ifx \url \undefined \def \url#1{\textsf{#1}}\fi
\ifx \bchapter \undefined \def \bchapter#1{#1}\fi
\ifx \bbook \undefined \def \bbook#1{#1}\fi
\ifx \bcomment \undefined \def \bcomment#1{#1}\fi
\ifx \oauthor \undefined \def \oauthor#1{#1}\fi
\ifx \citeauthoryear \undefined \def \citeauthoryear#1{#1}\fi
\ifx \endbibitem  \undefined \def \endbibitem {}\fi
\ifx \bconflocation  \undefined \def \bconflocation#1{#1}\fi
\ifx \arxivurl  \undefined \def \arxivurl#1{\textsf{#1}}\fi
\csname PreBibitemsHook\endcsname

%%% 1
\bibitem{sutton1998introduction}
\begin{botherref}
\oauthor{\bsnm{Sutton}, \binits{R.S.}},
\oauthor{\bsnm{Barto}, \binits{A.G.}}:
Introduction to reinforcement learning.
MIT press Cambridge
(1998)
\end{botherref}
\endbibitem

%%% 2
\bibitem{lecun2015deep}
\begin{barticle}
\bauthor{\bsnm{LeCun}, \binits{Y.}},
\bauthor{\bsnm{Bengio}, \binits{Y.}},
\bauthor{\bsnm{Hinton}, \binits{G.}}:
\batitle{Deep learning}.
\bjtitle{nature}
\bvolume{521}(\bissue{7553}),
\bfpage{436}--\blpage{444}
(\byear{2015})
\end{barticle}
\endbibitem

%%% 3
\bibitem{schmidhuber2015deep}
\begin{barticle}
\bauthor{\bsnm{Schmidhuber}, \binits{J.}}:
\batitle{Deep learning in neural networks: An overview}.
\bjtitle{Neural networks}
\bvolume{61},
\bfpage{85}--\blpage{117}
(\byear{2015})
\end{barticle}
\endbibitem

%%% 4
\bibitem{silver2016mastering}
\begin{barticle}
\bauthor{\bsnm{Silver}, \binits{D.}},
\bauthor{\bsnm{Huang}, \binits{A.}},
\bauthor{\bsnm{Maddison}, \binits{C.J.}},
\bauthor{\bsnm{Guez}, \binits{A.}},
\bauthor{\bsnm{Sifre}, \binits{L.}},
\bauthor{\bsnm{Van Den~Driessche}, \binits{G.}},
\bauthor{\bsnm{Schrittwieser}, \binits{J.}},
\bauthor{\bsnm{Antonoglou}, \binits{I.}},
\bauthor{\bsnm{Panneershelvam}, \binits{V.}},
\bauthor{\bsnm{Lanctot}, \binits{M.}}, \betal:
\batitle{Mastering the game of go with deep neural networks and tree search}.
\bjtitle{nature}
\bvolume{529}(\bissue{7587}),
\bfpage{484}
(\byear{2016})
\end{barticle}
\endbibitem

%%% 5
\bibitem{mnih2015human}
\begin{barticle}
\bauthor{\bsnm{Mnih}, \binits{V.}},
\bauthor{\bsnm{Kavukcuoglu}, \binits{K.}},
\bauthor{\bsnm{Silver}, \binits{D.}},
\bauthor{\bsnm{Rusu}, \binits{A.A.}},
\bauthor{\bsnm{Veness}, \binits{J.}},
\bauthor{\bsnm{Bellemare}, \binits{M.G.}},
\bauthor{\bsnm{Graves}, \binits{A.}},
\bauthor{\bsnm{Riedmiller}, \binits{M.}},
\bauthor{\bsnm{Fidjeland}, \binits{A.K.}},
\bauthor{\bsnm{Ostrovski}, \binits{G.}}, \betal:
\batitle{Human-level control through deep reinforcement learning}.
\bjtitle{Nature}
\bvolume{518}(\bissue{7540}),
\bfpage{529}
(\byear{2015})
\end{barticle}
\endbibitem

%%% 6
\bibitem{vinyals2019grandmaster}
\begin{botherref}
\oauthor{\bsnm{Vinyals}, \binits{O.}},
\oauthor{\bsnm{Babuschkin}, \binits{I.}},
\oauthor{\bsnm{Czarnecki}, \binits{W.M.}},
\oauthor{\bsnm{Mathieu}, \binits{M.}},
\oauthor{\bsnm{Dudzik}, \binits{A.}},
\oauthor{\bsnm{Chung}, \binits{J.}},
\oauthor{\bsnm{Choi}, \binits{D.H.}},
\oauthor{\bsnm{Powell}, \binits{R.}},
\oauthor{\bsnm{Ewalds}, \binits{T.}},
\oauthor{\bsnm{Georgiev}, \binits{P.}}, et al.:
Grandmaster level in starcraft ii using multi-agent reinforcement learning.
Nature,
1--5
(2019)
\end{botherref}
\endbibitem

%%% 7
\bibitem{tanner2005towards}
\begin{bchapter}
\bauthor{\bsnm{Tanner}, \binits{H.G.}},
\bauthor{\bsnm{Kumar}, \binits{A.}}:
\bctitle{Towards decentralization of multi-robot navigation functions}.
In: \bbtitle{Proceedings of the 2005 IEEE International Conference on Robotics
  and Automation},
pp. \bfpage{4132}--\blpage{4137}
(\byear{2005}).
\bcomment{IEEE}
\end{bchapter}
\endbibitem

%%% 8
\bibitem{brunet1995multi}
\begin{bchapter}
\bauthor{\bsnm{Brunet}, \binits{C.-A.}},
\bauthor{\bsnm{Gonzalez-Rubio}, \binits{R.}},
\bauthor{\bsnm{Tetreault}, \binits{M.}}:
\bctitle{A multi-agent architecture for a driver model for autonomous road
  vehicles}.
In: \bbtitle{Proceedings 1995 Canadian Conference on Electrical and Computer
  Engineering},
vol. \bseriesno{2},
pp. \bfpage{772}--\blpage{775}
(\byear{1995}).
\bcomment{IEEE}
\end{bchapter}
\endbibitem

%%% 9
\bibitem{dresner2004multiagent}
\begin{bchapter}
\bauthor{\bsnm{Dresner}, \binits{K.}},
\bauthor{\bsnm{Stone}, \binits{P.}}:
\bctitle{Multiagent traffic management: A reservation-based intersection
  control mechanism}.
In: \bbtitle{Proceedings of the Third International Joint Conference on
  Autonomous Agents and Multiagent Systems-Volume 2},
pp. \bfpage{530}--\blpage{537}
(\byear{2004}).
\bcomment{IEEE Computer Society}
\end{bchapter}
\endbibitem

%%% 10
\bibitem{lee2008multi}
\begin{barticle}
\bauthor{\bsnm{Lee}, \binits{J.-H.}},
\bauthor{\bsnm{Kim}, \binits{C.-O.}}:
\batitle{Multi-agent systems applications in manufacturing systems and supply
  chain management: a review paper}.
\bjtitle{International Journal of Production Research}
\bvolume{46}(\bissue{1}),
\bfpage{233}--\blpage{265}
(\byear{2008})
\end{barticle}
\endbibitem

%%% 11
\bibitem{hernandez2017survey}
\begin{botherref}
\oauthor{\bsnm{Hernandez-Leal}, \binits{P.}},
\oauthor{\bsnm{Kaisers}, \binits{M.}},
\oauthor{\bsnm{Baarslag}, \binits{T.}},
\oauthor{\bparticle{de} \bsnm{Cote}, \binits{E.M.}}:
A survey of learning in multiagent environments: Dealing with non-stationarity.
arXiv preprint arXiv:1707.09183
(2017)
\end{botherref}
\endbibitem

%%% 12
\bibitem{rahaie2009toward}
\begin{bchapter}
\bauthor{\bsnm{Rahaie}, \binits{Z.}},
\bauthor{\bsnm{Beigy}, \binits{H.}}:
\bctitle{Toward a solution to multi-agent credit assignment problem}.
In: \bbtitle{2009 International Conference of Soft Computing and Pattern
  Recognition},
pp. \bfpage{563}--\blpage{568}
(\byear{2009}).
\bcomment{IEEE}
\end{bchapter}
\endbibitem

%%% 13
\bibitem{harati2007knowledge}
\begin{barticle}
\bauthor{\bsnm{Harati}, \binits{A.}},
\bauthor{\bsnm{Ahmadabadi}, \binits{M.N.}},
\bauthor{\bsnm{Araabi}, \binits{B.N.}}:
\batitle{Knowledge-based multiagent credit assignment: A study on task type and
  critic information}.
\bjtitle{IEEE systems journal}
\bvolume{1}(\bissue{1}),
\bfpage{55}--\blpage{67}
(\byear{2007})
\end{barticle}
\endbibitem

%%% 14
\bibitem{yliniemi2014multi}
\begin{bchapter}
\bauthor{\bsnm{Yliniemi}, \binits{L.}},
\bauthor{\bsnm{Tumer}, \binits{K.}}:
\bctitle{Multi-objective multiagent credit assignment through difference
  rewards in reinforcement learning}.
In: \bbtitle{Asia-Pacific Conference on Simulated Evolution and Learning},
pp. \bfpage{407}--\blpage{418}
(\byear{2014}).
\bcomment{Springer}
\end{bchapter}
\endbibitem

%%% 15
\bibitem{agogino2004unifying}
\begin{bchapter}
\bauthor{\bsnm{Agogino}, \binits{A.K.}},
\bauthor{\bsnm{Tumer}, \binits{K.}}:
\bctitle{Unifying temporal and structural credit assignment problems}.
In: \bbtitle{AAMAS},
vol. \bseriesno{4},
pp. \bfpage{980}--\blpage{987}
(\byear{2004})
\end{bchapter}
\endbibitem

%%% 16
\bibitem{vorobeychik2017does}
\begin{barticle}
\bauthor{\bsnm{Vorobeychik}, \binits{Y.}},
\bauthor{\bsnm{Joveski}, \binits{Z.}},
\bauthor{\bsnm{Yu}, \binits{S.}}:
\batitle{Does communication help people coordinate?}
\bjtitle{PloS one}
\bvolume{12}(\bissue{2}),
\bfpage{0170780}
(\byear{2017})
\end{barticle}
\endbibitem

%%% 17
\bibitem{demichelis2008language}
\begin{barticle}
\bauthor{\bsnm{Demichelis}, \binits{S.}},
\bauthor{\bsnm{Weibull}, \binits{J.W.}}:
\batitle{Language, meaning, and games: A model of communication, coordination,
  and evolution}.
\bjtitle{American Economic Review}
\bvolume{98}(\bissue{4}),
\bfpage{1292}--\blpage{1311}
(\byear{2008})
\end{barticle}
\endbibitem

%%% 18
\bibitem{miller2004communication}
\begin{barticle}
\bauthor{\bsnm{Miller}, \binits{J.H.}},
\bauthor{\bsnm{Moser}, \binits{S.}}:
\batitle{Communication and coordination}.
\bjtitle{Complexity}
\bvolume{9}(\bissue{5}),
\bfpage{31}--\blpage{40}
(\byear{2004})
\end{barticle}
\endbibitem

%%% 19
\bibitem{kearns2012experiments}
\begin{barticle}
\bauthor{\bsnm{Kearns}, \binits{M.}}:
\batitle{Experiments in social computation}.
\bjtitle{Communications of the ACM}
\bvolume{55}(\bissue{10}),
\bfpage{56}--\blpage{67}
(\byear{2012})
\end{barticle}
\endbibitem

%%% 20
\bibitem{foerster2016learning}
\begin{bchapter}
\bauthor{\bsnm{Foerster}, \binits{J.}},
\bauthor{\bsnm{Assael}, \binits{I.A.}},
\bauthor{\bparticle{de} \bsnm{Freitas}, \binits{N.}},
\bauthor{\bsnm{Whiteson}, \binits{S.}}:
\bctitle{Learning to communicate with deep multi-agent reinforcement learning}.
In: \bbtitle{Advances in Neural Information Processing Systems},
pp. \bfpage{2137}--\blpage{2145}
(\byear{2016})
\end{bchapter}
\endbibitem

%%% 21
\bibitem{sukhbaatar2016learning}
\begin{bchapter}
\bauthor{\bsnm{Sukhbaatar}, \binits{S.}},
\bauthor{\bsnm{Fergus}, \binits{R.}}, \betal:
\bctitle{Learning multiagent communication with backpropagation}.
In: \bbtitle{Advances in Neural Information Processing Systems},
pp. \bfpage{2244}--\blpage{2252}
(\byear{2016})
\end{bchapter}
\endbibitem

%%% 22
\bibitem{singh2018learning}
\begin{botherref}
\oauthor{\bsnm{Singh}, \binits{A.}},
\oauthor{\bsnm{Jain}, \binits{T.}},
\oauthor{\bsnm{Sukhbaatar}, \binits{S.}}:
Learning when to communicate at scale in multiagent cooperative and competitive
  tasks.
ICLR
(2019)
\end{botherref}
\endbibitem

%%% 23
\bibitem{pesce2019improving}
\begin{botherref}
\oauthor{\bsnm{Pesce}, \binits{E.}},
\oauthor{\bsnm{Montana}, \binits{G.}}:
Improving coordination in multi-agent deep reinforcement learning through
  memory-driven communication.
Deep Reinforcement Learning Workshop, (NeurIPS 2018), Montreal, Canada
(2019)
\end{botherref}
\endbibitem

%%% 24
\bibitem{jiang2018learning}
\begin{botherref}
\oauthor{\bsnm{Jiang}, \binits{J.}},
\oauthor{\bsnm{Lu}, \binits{Z.}}:
Learning attentional communication for multi-agent cooperation.
arXiv preprint arXiv:1805.07733
(2018)
\end{botherref}
\endbibitem

%%% 25
\bibitem{mao2018modelling}
\begin{botherref}
\oauthor{\bsnm{Mao}, \binits{H.}},
\oauthor{\bsnm{Zhang}, \binits{Z.}},
\oauthor{\bsnm{Xiao}, \binits{Z.}},
\oauthor{\bsnm{Gong}, \binits{Z.}}:
Modelling the dynamic joint policy of teammates with attention multi-agent
  ddpg.
arXiv preprint arXiv:1811.07029
(2018)
\end{botherref}
\endbibitem

%%% 26
\bibitem{liu2020multi}
\begin{bchapter}
\bauthor{\bsnm{Liu}, \binits{Y.}},
\bauthor{\bsnm{Wang}, \binits{W.}},
\bauthor{\bsnm{Hu}, \binits{Y.}},
\bauthor{\bsnm{Hao}, \binits{J.}},
\bauthor{\bsnm{Chen}, \binits{X.}},
\bauthor{\bsnm{Gao}, \binits{Y.}}:
\bctitle{Multi-agent game abstraction via graph attention neural network.}
In: \bbtitle{AAAI},
pp. \bfpage{7211}--\blpage{7218}
(\byear{2020})
\end{bchapter}
\endbibitem

%%% 27
\bibitem{hoshen2017vain}
\begin{bchapter}
\bauthor{\bsnm{Hoshen}, \binits{Y.}}:
\bctitle{Vain: Attentional multi-agent predictive modeling}.
In: \bbtitle{Advances in Neural Information Processing Systems},
pp. \bfpage{2701}--\blpage{2711}
(\byear{2017})
\end{bchapter}
\endbibitem

%%% 28
\bibitem{das2018tarmac}
\begin{botherref}
\oauthor{\bsnm{Das}, \binits{A.}},
\oauthor{\bsnm{Gervet}, \binits{T.}},
\oauthor{\bsnm{Romoff}, \binits{J.}},
\oauthor{\bsnm{Batra}, \binits{D.}},
\oauthor{\bsnm{Parikh}, \binits{D.}},
\oauthor{\bsnm{Rabbat}, \binits{M.}},
\oauthor{\bsnm{Pineau}, \binits{J.}}:
Tarmac: Targeted multi-agent communication.
arXiv preprint arXiv:1810.11187
(2018)
\end{botherref}
\endbibitem

%%% 29
\bibitem{iqbal2018actor}
\begin{botherref}
\oauthor{\bsnm{Iqbal}, \binits{S.}},
\oauthor{\bsnm{Sha}, \binits{F.}}:
Actor-attention-critic for multi-agent reinforcement learning.
ICML
(2019)
\end{botherref}
\endbibitem

%%% 30
\bibitem{wang2019learning}
\begin{botherref}
\oauthor{\bsnm{Wang}, \binits{T.}},
\oauthor{\bsnm{Wang}, \binits{J.}},
\oauthor{\bsnm{Zheng}, \binits{C.}},
\oauthor{\bsnm{Zhang}, \binits{C.}}:
Learning nearly decomposable value functions via communication minimization.
arXiv preprint arXiv:1910.05366
(2019)
\end{botherref}
\endbibitem

%%% 31
\bibitem{zhang2008graph}
\begin{barticle}
\bauthor{\bsnm{Zhang}, \binits{F.}},
\bauthor{\bsnm{Hancock}, \binits{E.R.}}:
\batitle{Graph spectral image smoothing using the heat kernel}.
\bjtitle{Pattern Recognition}
\bvolume{41}(\bissue{11}),
\bfpage{3328}--\blpage{3342}
(\byear{2008})
\end{barticle}
\endbibitem

%%% 32
\bibitem{chung2016classifying}
\begin{bchapter}
\bauthor{\bsnm{Chung}, \binits{A.W.}},
\bauthor{\bsnm{Pesce}, \binits{E.}},
\bauthor{\bsnm{Monti}, \binits{R.P.}},
\bauthor{\bsnm{Montana}, \binits{G.}}:
\bctitle{Classifying hcp task-fmri networks using heat kernels}.
In: \bbtitle{2016 International Workshop on Pattern Recognition in NeuroImaging
  (PRNI)},
pp. \bfpage{1}--\blpage{4}
(\byear{2016}).
\bcomment{IEEE}
\end{bchapter}
\endbibitem

%%% 33
\bibitem{chung2016characterising}
\begin{barticle}
\bauthor{\bsnm{Chung}, \binits{A.W.}},
\bauthor{\bsnm{Schirmer}, \binits{M.}},
\bauthor{\bsnm{Krishnan}, \binits{M.L.}},
\bauthor{\bsnm{Ball}, \binits{G.}},
\bauthor{\bsnm{Aljabar}, \binits{P.}},
\bauthor{\bsnm{Edwards}, \binits{A.D.}},
\bauthor{\bsnm{Montana}, \binits{G.}}:
\batitle{Characterising brain network topologies: a dynamic analysis approach
  using heat kernels}.
\bjtitle{Neuroimage}
\bvolume{141},
\bfpage{490}--\blpage{501}
(\byear{2016})
\end{barticle}
\endbibitem

%%% 34
\bibitem{degris2012off}
\begin{botherref}
\oauthor{\bsnm{Degris}, \binits{T.}},
\oauthor{\bsnm{White}, \binits{M.}},
\oauthor{\bsnm{Sutton}, \binits{R.S.}}:
Off-policy actor-critic.
arXiv preprint arXiv:1205.4839
(2012)
\end{botherref}
\endbibitem

%%% 35
\bibitem{silver2014deterministic}
\begin{bchapter}
\bauthor{\bsnm{Silver}, \binits{D.}},
\bauthor{\bsnm{Lever}, \binits{G.}},
\bauthor{\bsnm{Heess}, \binits{N.}},
\bauthor{\bsnm{Degris}, \binits{T.}},
\bauthor{\bsnm{Wierstra}, \binits{D.}},
\bauthor{\bsnm{Riedmiller}, \binits{M.}}:
\bctitle{Deterministic policy gradient algorithms}.
In: \bbtitle{ICML}
(\byear{2014})
\end{bchapter}
\endbibitem

%%% 36
\bibitem{lillicrapHPHETS15}
\begin{botherref}
\oauthor{\bsnm{Lillicrap}, \binits{T.P.}},
\oauthor{\bsnm{Hunt}, \binits{J.J.}},
\oauthor{\bsnm{Pritzel}, \binits{A.}},
\oauthor{\bsnm{Heess}, \binits{N.}},
\oauthor{\bsnm{Erez}, \binits{T.}},
\oauthor{\bsnm{Tassa}, \binits{Y.}},
\oauthor{\bsnm{Silver}, \binits{D.}},
\oauthor{\bsnm{Wierstra}, \binits{D.}}:
Continuous control with deep reinforcement learning.
CoRR
\textbf{abs/1509.02971}
(2015)
\end{botherref}
\endbibitem

%%% 37
\bibitem{lowe2017multi}
\begin{bchapter}
\bauthor{\bsnm{Lowe}, \binits{R.}},
\bauthor{\bsnm{Wu}, \binits{Y.}},
\bauthor{\bsnm{Tamar}, \binits{A.}},
\bauthor{\bsnm{Harb}, \binits{J.}},
\bauthor{\bsnm{Abbeel}, \binits{O.P.}},
\bauthor{\bsnm{Mordatch}, \binits{I.}}:
\bctitle{Multi-agent actor-critic for mixed cooperative-competitive
  environments}.
In: \bbtitle{Advances in Neural Information Processing Systems},
pp. \bfpage{6379}--\blpage{6390}
(\byear{2017})
\end{bchapter}
\endbibitem

%%% 38
\bibitem{stone2000multiagent}
\begin{barticle}
\bauthor{\bsnm{Stone}, \binits{P.}},
\bauthor{\bsnm{Veloso}, \binits{M.}}:
\batitle{Multiagent systems: A survey from a machine learning perspective}.
\bjtitle{Autonomous Robots}
\bvolume{8}(\bissue{3}),
\bfpage{345}--\blpage{383}
(\byear{2000})
\end{barticle}
\endbibitem

%%% 39
\bibitem{parsons2002game}
\begin{barticle}
\bauthor{\bsnm{Parsons}, \binits{S.}},
\bauthor{\bsnm{Wooldridge}, \binits{M.}}:
\batitle{Game theory and decision theory in multi-agent systems}.
\bjtitle{Autonomous Agents and Multi-Agent Systems}
\bvolume{5}(\bissue{3}),
\bfpage{243}--\blpage{254}
(\byear{2002})
\end{barticle}
\endbibitem

%%% 40
\bibitem{shoham2008multiagent}
\begin{botherref}
\oauthor{\bsnm{Shoham}, \binits{Y.}},
\oauthor{\bsnm{Leyton-Brown}, \binits{K.}}:
Multiagent systems: Algorithmic, game-theoretic, and logical foundations.
Cambridge University Press
(2008)
\end{botherref}
\endbibitem

%%% 41
\bibitem{nguyen2020deep}
\begin{botherref}
\oauthor{\bsnm{Nguyen}, \binits{T.T.}},
\oauthor{\bsnm{Nguyen}, \binits{N.D.}},
\oauthor{\bsnm{Nahavandi}, \binits{S.}}:
Deep reinforcement learning for multiagent systems: A review of challenges,
  solutions, and applications.
IEEE transactions on cybernetics
(2020)
\end{botherref}
\endbibitem

%%% 42
\bibitem{hernandez2019survey}
\begin{barticle}
\bauthor{\bsnm{Hernandez-Leal}, \binits{P.}},
\bauthor{\bsnm{Kartal}, \binits{B.}},
\bauthor{\bsnm{Taylor}, \binits{M.E.}}:
\batitle{A survey and critique of multiagent deep reinforcement learning}.
\bjtitle{Autonomous Agents and Multi-Agent Systems}
\bvolume{33}(\bissue{6}),
\bfpage{750}--\blpage{797}
(\byear{2019})
\end{barticle}
\endbibitem

%%% 43
\bibitem{albrecht2018autonomous}
\begin{barticle}
\bauthor{\bsnm{Albrecht}, \binits{S.V.}},
\bauthor{\bsnm{Stone}, \binits{P.}}:
\batitle{Autonomous agents modelling other agents: A comprehensive survey and
  open problems}.
\bjtitle{Artificial Intelligence}
\bvolume{258},
\bfpage{66}--\blpage{95}
(\byear{2018})
\end{barticle}
\endbibitem

%%% 44
\bibitem{busoniu2008comprehensive}
\begin{barticle}
\bauthor{\bsnm{Busoniu}, \binits{L.}},
\bauthor{\bsnm{Babuska}, \binits{R.}},
\bauthor{\bsnm{De~Schutter}, \binits{B.}}:
\batitle{A comprehensive survey of multiagent reinforcement learning}.
\bjtitle{IEEE Transactions on Systems, Man, and Cybernetics, Part C
  (Applications and Reviews)}
\bvolume{38}(\bissue{2}),
\bfpage{156}--\blpage{172}
(\byear{2008})
\end{barticle}
\endbibitem

%%% 45
\bibitem{tuyls2012multiagent}
\begin{barticle}
\bauthor{\bsnm{Tuyls}, \binits{K.}},
\bauthor{\bsnm{Weiss}, \binits{G.}}:
\batitle{Multiagent learning: Basics, challenges, and prospects}.
\bjtitle{Ai Magazine}
\bvolume{33}(\bissue{3}),
\bfpage{41}
(\byear{2012})
\end{barticle}
\endbibitem

%%% 46
\bibitem{laurent2011world}
\begin{barticle}
\bauthor{\bsnm{Laurent}, \binits{G.J.}},
\bauthor{\bsnm{Matignon}, \binits{L.}},
\bauthor{\bsnm{Fort-Piat}, \binits{L.}}, \betal:
\batitle{The world of independent learners is not markovian}.
\bjtitle{International Journal of Knowledge-based and Intelligent Engineering
  Systems}
\bvolume{15}(\bissue{1}),
\bfpage{55}--\blpage{64}
(\byear{2011})
\end{barticle}
\endbibitem

%%% 47
\bibitem{kraemer2016multi}
\begin{barticle}
\bauthor{\bsnm{Kraemer}, \binits{L.}},
\bauthor{\bsnm{Banerjee}, \binits{B.}}:
\batitle{Multi-agent reinforcement learning as a rehearsal for decentralized
  planning}.
\bjtitle{Neurocomputing}
\bvolume{190},
\bfpage{82}--\blpage{94}
(\byear{2016})
\end{barticle}
\endbibitem

%%% 48
\bibitem{foerster2017counterfactual}
\begin{botherref}
\oauthor{\bsnm{Foerster}, \binits{J.}},
\oauthor{\bsnm{Farquhar}, \binits{G.}},
\oauthor{\bsnm{Afouras}, \binits{T.}},
\oauthor{\bsnm{Nardelli}, \binits{N.}},
\oauthor{\bsnm{Whiteson}, \binits{S.}}:
Counterfactual multi-agent policy gradients.
arXiv preprint arXiv:1705.08926
(2017)
\end{botherref}
\endbibitem

%%% 49
\bibitem{wang2020r}
\begin{botherref}
\oauthor{\bsnm{Wang}, \binits{R.E.}},
\oauthor{\bsnm{Everett}, \binits{M.}},
\oauthor{\bsnm{How}, \binits{J.P.}}:
R-maddpg for partially observable environments and limited communication.
arXiv preprint arXiv:2002.06684
(2020)
\end{botherref}
\endbibitem

%%% 50
\bibitem{hochreiter1997long}
\begin{barticle}
\bauthor{\bsnm{Hochreiter}, \binits{S.}},
\bauthor{\bsnm{Schmidhuber}, \binits{J.}}:
\batitle{Long short-term memory}.
\bjtitle{Neural computation}
\bvolume{9}(\bissue{8}),
\bfpage{1735}--\blpage{1780}
(\byear{1997})
\end{barticle}
\endbibitem

%%% 51
\bibitem{lin2018efficient}
\begin{bchapter}
\bauthor{\bsnm{Lin}, \binits{K.}},
\bauthor{\bsnm{Zhao}, \binits{R.}},
\bauthor{\bsnm{Xu}, \binits{Z.}},
\bauthor{\bsnm{Zhou}, \binits{J.}}:
\bctitle{Efficient large-scale fleet management via multi-agent deep
  reinforcement learning}.
In: \bbtitle{Proceedings of the 24th ACM SIGKDD International Conference on
  Knowledge Discovery \& Data Mining},
pp. \bfpage{1774}--\blpage{1783}
(\byear{2018})
\end{bchapter}
\endbibitem

%%% 52
\bibitem{scardovi2008synchronization}
\begin{bchapter}
\bauthor{\bsnm{Scardovi}, \binits{L.}},
\bauthor{\bsnm{Sepulchre}, \binits{R.}}:
\bctitle{Synchronization in networks of identical linear systems}.
In: \bbtitle{Decision and Control, 2008. CDC 2008. 47th IEEE Conference On},
pp. \bfpage{546}--\blpage{551}
(\byear{2008}).
\bcomment{IEEE}
\end{bchapter}
\endbibitem

%%% 53
\bibitem{wen2012consensus}
\begin{barticle}
\bauthor{\bsnm{Wen}, \binits{G.}},
\bauthor{\bsnm{Duan}, \binits{Z.}},
\bauthor{\bsnm{Yu}, \binits{W.}},
\bauthor{\bsnm{Chen}, \binits{G.}}:
\batitle{Consensus in multi-agent systems with communication constraints}.
\bjtitle{International Journal of Robust and Nonlinear Control}
\bvolume{22}(\bissue{2}),
\bfpage{170}--\blpage{182}
(\byear{2012})
\end{barticle}
\endbibitem

%%% 54
\bibitem{wunder2009communication}
\begin{bchapter}
\bauthor{\bsnm{Wunder}, \binits{M.}},
\bauthor{\bsnm{Littman}, \binits{M.}},
\bauthor{\bsnm{Stone}, \binits{M.}}:
\bctitle{Communication, credibility and negotiation using a cognitive hierarchy
  model}.
In: \bbtitle{Workshop\# 19: MSDM 2009},
p. \bfpage{73}
(\byear{2009})
\end{bchapter}
\endbibitem

%%% 55
\bibitem{ito2011innovations}
\begin{botherref}
\oauthor{\bsnm{It{\=o}}, \binits{T.}},
\oauthor{\bsnm{Zhang}, \binits{M.}},
\oauthor{\bsnm{Robu}, \binits{V.}},
\oauthor{\bsnm{Fatima}, \binits{S.}},
\oauthor{\bsnm{Matsuo}, \binits{T.}},
\oauthor{\bsnm{Yamaki}, \binits{H.}}:
Innovations in Agent-Based Complex Automated Negotiations.
Springer
(2011)
\end{botherref}
\endbibitem

%%% 56
\bibitem{fox2000probabilistic}
\begin{barticle}
\bauthor{\bsnm{Fox}, \binits{D.}},
\bauthor{\bsnm{Burgard}, \binits{W.}},
\bauthor{\bsnm{Kruppa}, \binits{H.}},
\bauthor{\bsnm{Thrun}, \binits{S.}}:
\batitle{A probabilistic approach to collaborative multi-robot localization}.
\bjtitle{Autonomous robots}
\bvolume{8}(\bissue{3}),
\bfpage{325}--\blpage{344}
(\byear{2000})
\end{barticle}
\endbibitem

%%% 57
\bibitem{peng2017multiagent}
\begin{botherref}
\oauthor{\bsnm{Peng}, \binits{P.}},
\oauthor{\bsnm{Yuan}, \binits{Q.}},
\oauthor{\bsnm{Wen}, \binits{Y.}},
\oauthor{\bsnm{Yang}, \binits{Y.}},
\oauthor{\bsnm{Tang}, \binits{Z.}},
\oauthor{\bsnm{Long}, \binits{H.}},
\oauthor{\bsnm{Wang}, \binits{J.}}:
Multiagent bidirectionally-coordinated nets for learning to play starcraft
  combat games.
arXiv preprint arXiv:1703.10069
(2017)
\end{botherref}
\endbibitem

%%% 58
\bibitem{kim2020communication}
\begin{bchapter}
\bauthor{\bsnm{Kim}, \binits{W.}},
\bauthor{\bsnm{Park}, \binits{J.}},
\bauthor{\bsnm{Sung}, \binits{Y.}}:
\bctitle{Communication in multi-agent reinforcement learning: Intention
  sharing}.
In: \bbtitle{International Conference on Learning Representations}
(\byear{2020})
\end{bchapter}
\endbibitem

%%% 59
\bibitem{vaswani2017attention}
\begin{bchapter}
\bauthor{\bsnm{Vaswani}, \binits{A.}},
\bauthor{\bsnm{Shazeer}, \binits{N.}},
\bauthor{\bsnm{Parmar}, \binits{N.}},
\bauthor{\bsnm{Uszkoreit}, \binits{J.}},
\bauthor{\bsnm{Jones}, \binits{L.}},
\bauthor{\bsnm{Gomez}, \binits{A.N.}},
\bauthor{\bsnm{Kaiser}, \binits{{\L}.}},
\bauthor{\bsnm{Polosukhin}, \binits{I.}}:
\bctitle{Attention is all you need}.
In: \bbtitle{Advances in Neural Information Processing Systems},
pp. \bfpage{5998}--\blpage{6008}
(\byear{2017})
\end{bchapter}
\endbibitem

%%% 60
\bibitem{chung1997spectral}
\begin{botherref}
\oauthor{\bsnm{Chung}, \binits{F.R.}},
\oauthor{\bsnm{Graham}, \binits{F.C.}}:
Spectral graph theory.
American Mathematical Soc.
(1997)
\end{botherref}
\endbibitem

%%% 61
\bibitem{brouwer2011spectra}
\begin{botherref}
\oauthor{\bsnm{Brouwer}, \binits{A.E.}},
\oauthor{\bsnm{Haemers}, \binits{W.H.}}:
Spectra of graphs.
Springer
(2011)
\end{botherref}
\endbibitem

%%% 62
\bibitem{cvetkovic1980spectra}
\begin{botherref}
\oauthor{\bsnm{Cvetkovic}, \binits{D.M.}},
\oauthor{\bsnm{DM}, \binits{C.}}, et al.:
Spectra of graphs. theory and application
(1980)
\end{botherref}
\endbibitem

%%% 63
\bibitem{differentialgeometry}
\begin{botherref}
\oauthor{\bsnm{Schoen}, \binits{R.}},
\oauthor{\bsnm{Shing-Tung Yau~Mack}, \binits{C.A.}}:
Lectures on Differential Geometry.
International Press
(1994)
\end{botherref}
\endbibitem

%%% 64
\bibitem{kloster2014heat}
\begin{bchapter}
\bauthor{\bsnm{Kloster}, \binits{K.}},
\bauthor{\bsnm{Gleich}, \binits{D.F.}}:
\bctitle{Heat kernel based community detection}.
In: \bbtitle{Proceedings of the 20th ACM SIGKDD International Conference on
  Knowledge Discovery and Data Mining},
pp. \bfpage{1386}--\blpage{1395}
(\byear{2014}).
\bcomment{ACM}
\end{bchapter}
\endbibitem

%%% 65
\bibitem{lafferty2005diffusion}
\begin{barticle}
\bauthor{\bsnm{Lafferty}, \binits{J.}},
\bauthor{\bsnm{Lebanon}, \binits{G.}}:
\batitle{Diffusion kernels on statistical manifolds}.
\bjtitle{Journal of Machine Learning Research}
\bvolume{6}(\bissue{Jan}),
\bfpage{129}--\blpage{163}
(\byear{2005})
\end{barticle}
\endbibitem

%%% 66
\bibitem{xu2020graph}
\begin{botherref}
\oauthor{\bsnm{Xu}, \binits{B.}},
\oauthor{\bsnm{Shen}, \binits{H.}},
\oauthor{\bsnm{Cao}, \binits{Q.}},
\oauthor{\bsnm{Cen}, \binits{K.}},
\oauthor{\bsnm{Cheng}, \binits{X.}}:
Graph convolutional networks using heat kernel for semi-supervised learning.
arXiv preprint arXiv:2007.16002
(2020)
\end{botherref}
\endbibitem

%%% 67
\bibitem{klicpera2019diffusion}
\begin{bchapter}
\bauthor{\bsnm{Klicpera}, \binits{J.}},
\bauthor{\bsnm{Wei{\ss}enberger}, \binits{S.}},
\bauthor{\bsnm{G{\"u}nnemann}, \binits{S.}}:
\bctitle{Diffusion improves graph learning}.
In: \bbtitle{Advances in Neural Information Processing Systems},
pp. \bfpage{13354}--\blpage{13366}
(\byear{2019})
\end{bchapter}
\endbibitem

%%% 68
\bibitem{kschischang2001factor}
\begin{barticle}
\bauthor{\bsnm{Kschischang}, \binits{F.R.}},
\bauthor{\bsnm{Frey}, \binits{B.J.}},
\bauthor{\bsnm{Loeliger}, \binits{H.-A.}}, \betal:
\batitle{Factor graphs and the sum-product algorithm}.
\bjtitle{IEEE Transactions on information theory}
\bvolume{47}(\bissue{2}),
\bfpage{498}--\blpage{519}
(\byear{2001})
\end{barticle}
\endbibitem

%%% 69
\bibitem{kuyer2008multiagent}
\begin{bchapter}
\bauthor{\bsnm{Kuyer}, \binits{L.}},
\bauthor{\bsnm{Whiteson}, \binits{S.}},
\bauthor{\bsnm{Bakker}, \binits{B.}},
\bauthor{\bsnm{Vlassis}, \binits{N.}}:
\bctitle{Multiagent reinforcement learning for urban traffic control using
  coordination graphs}.
In: \bbtitle{Joint European Conference on Machine Learning and Knowledge
  Discovery in Databases},
pp. \bfpage{656}--\blpage{671}
(\byear{2008}).
\bcomment{Springer}
\end{bchapter}
\endbibitem

%%% 70
\bibitem{guestrin2002multiagent}
\begin{bchapter}
\bauthor{\bsnm{Guestrin}, \binits{C.}},
\bauthor{\bsnm{Koller}, \binits{D.}},
\bauthor{\bsnm{Parr}, \binits{R.}}:
\bctitle{Multiagent planning with factored mdps}.
In: \bbtitle{Advances in Neural Information Processing Systems},
pp. \bfpage{1523}--\blpage{1530}
(\byear{2002})
\end{bchapter}
\endbibitem

%%% 71
\bibitem{liao2021review}
\begin{botherref}
\oauthor{\bsnm{Liao}, \binits{W.}},
\oauthor{\bsnm{Bak-Jensen}, \binits{B.}},
\oauthor{\bsnm{Pillai}, \binits{J.R.}},
\oauthor{\bsnm{Wang}, \binits{Y.}},
\oauthor{\bsnm{Wang}, \binits{Y.}}:
A review of graph neural networks and their applications in power systems.
arXiv preprint arXiv:2101.10025
(2021)
\end{botherref}
\endbibitem

%%% 72
\bibitem{zhou2021ast}
\begin{barticle}
\bauthor{\bsnm{Zhou}, \binits{H.}},
\bauthor{\bsnm{Ren}, \binits{D.}},
\bauthor{\bsnm{Xia}, \binits{H.}},
\bauthor{\bsnm{Fan}, \binits{M.}},
\bauthor{\bsnm{Yang}, \binits{X.}},
\bauthor{\bsnm{Huang}, \binits{H.}}:
\batitle{Ast-gnn: An attention-based spatio-temporal graph neural network for
  interaction-aware pedestrian trajectory prediction}.
\bjtitle{Neurocomputing}
\bvolume{445},
\bfpage{298}--\blpage{308}
(\byear{2021})
\end{barticle}
\endbibitem

%%% 73
\bibitem{huang2019stgat}
\begin{bchapter}
\bauthor{\bsnm{Huang}, \binits{Y.}},
\bauthor{\bsnm{Bi}, \binits{H.}},
\bauthor{\bsnm{Li}, \binits{Z.}},
\bauthor{\bsnm{Mao}, \binits{T.}},
\bauthor{\bsnm{Wang}, \binits{Z.}}:
\bctitle{Stgat: Modeling spatial-temporal interactions for human trajectory
  prediction}.
In: \bbtitle{Proceedings of the IEEE/CVF International Conference on Computer
  Vision},
pp. \bfpage{6272}--\blpage{6281}
(\byear{2019})
\end{bchapter}
\endbibitem

%%% 74
\bibitem{mohamed2020social}
\begin{bchapter}
\bauthor{\bsnm{Mohamed}, \binits{A.}},
\bauthor{\bsnm{Qian}, \binits{K.}},
\bauthor{\bsnm{Elhoseiny}, \binits{M.}},
\bauthor{\bsnm{Claudel}, \binits{C.}}:
\bctitle{Social-stgcnn: A social spatio-temporal graph convolutional neural
  network for human trajectory prediction}.
In: \bbtitle{Proceedings of the IEEE/CVF Conference on Computer Vision and
  Pattern Recognition},
pp. \bfpage{14424}--\blpage{14432}
(\byear{2020})
\end{bchapter}
\endbibitem

%%% 75
\bibitem{xu2021learning}
\begin{botherref}
\oauthor{\bsnm{Xu}, \binits{Z.}},
\oauthor{\bsnm{Zhang}, \binits{B.}},
\oauthor{\bsnm{Bai}, \binits{Y.}},
\oauthor{\bsnm{Li}, \binits{D.}},
\oauthor{\bsnm{Fan}, \binits{G.}}:
Learning to coordinate via multiple graph neural networks.
arXiv preprint arXiv:2104.03503
(2021)
\end{botherref}
\endbibitem

%%% 76
\bibitem{wang2019stmarl}
\begin{botherref}
\oauthor{\bsnm{Wang}, \binits{Y.}},
\oauthor{\bsnm{Xu}, \binits{T.}},
\oauthor{\bsnm{Niu}, \binits{X.}},
\oauthor{\bsnm{Tan}, \binits{C.}},
\oauthor{\bsnm{Chen}, \binits{E.}},
\oauthor{\bsnm{Xiong}, \binits{H.}}:
Stmarl: A spatio-temporal multi-agent reinforcement learning approach for
  traffic light control.
arXiv preprint arXiv:1908.10577
(2019)
\end{botherref}
\endbibitem

%%% 77
\bibitem{li2020deep}
\begin{botherref}
\oauthor{\bsnm{Li}, \binits{S.}},
\oauthor{\bsnm{Gupta}, \binits{J.K.}},
\oauthor{\bsnm{Morales}, \binits{P.}},
\oauthor{\bsnm{Allen}, \binits{R.}},
\oauthor{\bsnm{Kochenderfer}, \binits{M.J.}}:
Deep implicit coordination graphs for multi-agent reinforcement learning.
arXiv preprint arXiv:2006.11438
(2020)
\end{botherref}
\endbibitem

%%% 78
\bibitem{jiang2018graph}
\begin{botherref}
\oauthor{\bsnm{Jiang}, \binits{J.}},
\oauthor{\bsnm{Dun}, \binits{C.}},
\oauthor{\bsnm{Huang}, \binits{T.}},
\oauthor{\bsnm{Lu}, \binits{Z.}}:
Graph convolutional reinforcement learning.
arXiv preprint arXiv:1810.09202
(2018)
\end{botherref}
\endbibitem

%%% 79
\bibitem{chen2020gama}
\begin{barticle}
\bauthor{\bsnm{Chen}, \binits{H.}},
\bauthor{\bsnm{Liu}, \binits{Y.}},
\bauthor{\bsnm{Zhou}, \binits{Z.}},
\bauthor{\bsnm{Hu}, \binits{D.}},
\bauthor{\bsnm{Zhang}, \binits{M.}}:
\batitle{Gama: Graph attention multi-agent reinforcement learning algorithm for
  cooperation}.
\bjtitle{Applied Intelligence}
\bvolume{50}(\bissue{12}),
\bfpage{4195}--\blpage{4205}
(\byear{2020})
\end{barticle}
\endbibitem

%%% 80
\bibitem{seraj2021heterogeneous}
\begin{botherref}
\oauthor{\bsnm{Seraj}, \binits{E.}},
\oauthor{\bsnm{Wang}, \binits{Z.}},
\oauthor{\bsnm{Paleja}, \binits{R.}},
\oauthor{\bsnm{Sklar}, \binits{M.}},
\oauthor{\bsnm{Patel}, \binits{A.}},
\oauthor{\bsnm{Gombolay}, \binits{M.}}:
Heterogeneous graph attention networks for learning diverse communication.
arXiv preprint arXiv:2108.09568
(2021)
\end{botherref}
\endbibitem

%%% 81
\bibitem{su2020counterfactual}
\begin{botherref}
\oauthor{\bsnm{Su}, \binits{J.}},
\oauthor{\bsnm{Adams}, \binits{S.}},
\oauthor{\bsnm{Beling}, \binits{P.A.}}:
Counterfactual multi-agent reinforcement learning with graph convolution
  communication.
arXiv preprint arXiv:2004.00470
(2020)
\end{botherref}
\endbibitem

%%% 82
\bibitem{yuan2021graphcomm}
\begin{botherref}
\oauthor{\bsnm{Yuan}, \binits{Q.}},
\oauthor{\bsnm{Fu}, \binits{X.}},
\oauthor{\bsnm{Li}, \binits{Z.}},
\oauthor{\bsnm{Luo}, \binits{G.}},
\oauthor{\bsnm{Li}, \binits{J.}},
\oauthor{\bsnm{Yang}, \binits{F.}}:
Graphcomm: Efficient graph convolutional communication for multi-agent
  cooperation.
IEEE Internet of Things Journal
(2021)
\end{botherref}
\endbibitem

%%% 83
\bibitem{niu2021multi}
\begin{bchapter}
\bauthor{\bsnm{Niu}, \binits{Y.}},
\bauthor{\bsnm{Paleja}, \binits{R.}},
\bauthor{\bsnm{Gombolay}, \binits{M.}}:
\bctitle{Multi-agent graph-attention communication and teaming}.
In: \bbtitle{Proceedings of the 20th International Conference on Autonomous
  Agents and MultiAgent Systems},
pp. \bfpage{964}--\blpage{973}
(\byear{2021})
\end{bchapter}
\endbibitem

%%% 84
\bibitem{littman1994markov}
\begin{bchapter}
\bauthor{\bsnm{Littman}, \binits{M.L.}}:
\bctitle{Markov games as a framework for multi-agent reinforcement learning}.
In: \bbtitle{Machine Learning Proceedings 1994},
pp. \bfpage{157}--\blpage{163}.
\bpublisher{Elsevier}, \blocation{???}
(\byear{1994})
\end{bchapter}
\endbibitem

%%% 85
\bibitem{kondor2002diffusion}
\begin{bchapter}
\bauthor{\bsnm{Kondor}, \binits{R.}},
\bauthor{\bsnm{Lafferty}, \binits{J.}}:
\bctitle{Diffusion kernels on graphs and other discrete input spaces. icml
  2002}.
In: \bbtitle{Proc},
pp. \bfpage{315}--\blpage{322}
(\byear{2002})
\end{bchapter}
\endbibitem

%%% 86
\bibitem{fiedler1989laplacian}
\begin{barticle}
\bauthor{\bsnm{Fiedler}, \binits{M.}}:
\batitle{Laplacian of graphs and algebraic connectivity}.
\bjtitle{Banach Center Publications}
\bvolume{25}(\bissue{1}),
\bfpage{57}--\blpage{70}
(\byear{1989})
\end{barticle}
\endbibitem

%%% 87
\bibitem{al2009new}
\begin{barticle}
\bauthor{\bsnm{Al-Mohy}, \binits{A.H.}},
\bauthor{\bsnm{Higham}, \binits{N.J.}}:
\batitle{A new scaling and squaring algorithm for the matrix exponential}.
\bjtitle{SIAM Journal on Matrix Analysis and Applications}
\bvolume{31}(\bissue{3}),
\bfpage{970}--\blpage{989}
(\byear{2009})
\end{barticle}
\endbibitem

%%% 88
\bibitem{cheng2005heritage}
\begin{barticle}
\bauthor{\bsnm{Cheng}, \binits{A.H.-D.}},
\bauthor{\bsnm{Cheng}, \binits{D.T.}}:
\batitle{Heritage and early history of the boundary element method}.
\bjtitle{Engineering Analysis with Boundary Elements}
\bvolume{29}(\bissue{3}),
\bfpage{268}--\blpage{302}
(\byear{2005})
\end{barticle}
\endbibitem

%%% 89
\bibitem{mesbahi2010graph}
\begin{botherref}
\oauthor{\bsnm{Mesbahi}, \binits{M.}},
\oauthor{\bsnm{Egerstedt}, \binits{M.}}:
Graph theoretic methods in multiagent networks.
Princeton University Press
(2010)
\end{botherref}
\endbibitem

%%% 90
\bibitem{balch1998behavior}
\begin{barticle}
\bauthor{\bsnm{Balch}, \binits{T.}},
\bauthor{\bsnm{Arkin}, \binits{R.C.}}:
\batitle{Behavior-based formation control for multirobot teams}.
\bjtitle{IEEE transactions on robotics and automation}
\bvolume{14}(\bissue{6}),
\bfpage{926}--\blpage{939}
(\byear{1998})
\end{barticle}
\endbibitem

%%% 91
\bibitem{agarwal2019learning}
\begin{botherref}
\oauthor{\bsnm{Agarwal}, \binits{A.}},
\oauthor{\bsnm{Kumar}, \binits{S.}},
\oauthor{\bsnm{Sycara}, \binits{K.}}:
Learning transferable cooperative behavior in multi-agent teams.
arXiv preprint arXiv:1906.01202
(2019)
\end{botherref}
\endbibitem

%%% 92
\bibitem{mordatch2017emergence}
\begin{botherref}
\oauthor{\bsnm{Mordatch}, \binits{I.}},
\oauthor{\bsnm{Abbeel}, \binits{P.}}:
Emergence of grounded compositional language in multi-agent populations.
arXiv preprint arXiv:1703.04908
(2017)
\end{botherref}
\endbibitem

%%% 93
\bibitem{schmidhuber1996general}
\begin{bchapter}
\bauthor{\bsnm{Schmidhuber}, \binits{J.}}:
\bctitle{A general method for multi-agent reinforcement learning in
  unrestricted environments}.
In: \bbtitle{Adaptation, Coevolution and Learning in Multiagent Systems: Papers
  from the 1996 AAAI Spring Symposium},
pp. \bfpage{84}--\blpage{87}
(\byear{1996})
\end{bchapter}
\endbibitem

%%% 94
\bibitem{kingma2014adam}
\begin{botherref}
\oauthor{\bsnm{Kingma}, \binits{D.P.}},
\oauthor{\bsnm{Ba}, \binits{J.}}:
Adam: A method for stochastic optimization.
arXiv preprint arXiv:1412.6980
(2014)
\end{botherref}
\endbibitem

%%% 95
\bibitem{van1995python}
\begin{botherref}
\oauthor{\bsnm{Van~Rossum}, \binits{G.}},
\oauthor{\bsnm{Drake~Jr}, \binits{F.L.}}:
Python tutorial.
Centrum voor Wiskunde en Informatica Amsterdam, The Netherlands
(1995)
\end{botherref}
\endbibitem

%%% 96
\bibitem{paszke2017automatic}
\begin{botherref}
\oauthor{\bsnm{Paszke}, \binits{A.}},
\oauthor{\bsnm{Gross}, \binits{S.}},
\oauthor{\bsnm{Chintala}, \binits{S.}},
\oauthor{\bsnm{Chanan}, \binits{G.}},
\oauthor{\bsnm{Yang}, \binits{E.}},
\oauthor{\bsnm{DeVito}, \binits{Z.}},
\oauthor{\bsnm{Lin}, \binits{Z.}},
\oauthor{\bsnm{Desmaison}, \binits{A.}},
\oauthor{\bsnm{Antiga}, \binits{L.}},
\oauthor{\bsnm{Lerer}, \binits{A.}}:
Automatic differentiation in PyTorch
(2017)
\end{botherref}
\endbibitem

%%% 97
\bibitem{hagberg2008exploring}
\begin{botherref}
\oauthor{\bsnm{Hagberg}, \binits{A.}},
\oauthor{\bsnm{Swart}, \binits{P.}},
\oauthor{\bsnm{S~Chult}, \binits{D.}}:
Exploring network structure, dynamics, and function using networkx.
Technical report,
Los Alamos National Lab.(LANL), Los Alamos, NM (United States)
(2008)
\end{botherref}
\endbibitem

%%% 98
\bibitem{liu2020when2com}
\begin{bchapter}
\bauthor{\bsnm{Liu}, \binits{Y.-C.}},
\bauthor{\bsnm{Tian}, \binits{J.}},
\bauthor{\bsnm{Glaser}, \binits{N.}},
\bauthor{\bsnm{Kira}, \binits{Z.}}:
\bctitle{When2com: Multi-agent perception via communication graph grouping}.
In: \bbtitle{Proceedings of the IEEE/CVF Conference on Computer Vision and
  Pattern Recognition},
pp. \bfpage{4106}--\blpage{4115}
(\byear{2020})
\end{bchapter}
\endbibitem

%%% 99
\bibitem{breazeal2005effects}
\begin{bchapter}
\bauthor{\bsnm{Breazeal}, \binits{C.}},
\bauthor{\bsnm{Kidd}, \binits{C.D.}},
\bauthor{\bsnm{Thomaz}, \binits{A.L.}},
\bauthor{\bsnm{Hoffman}, \binits{G.}},
\bauthor{\bsnm{Berlin}, \binits{M.}}:
\bctitle{Effects of nonverbal communication on efficiency and robustness in
  human-robot teamwork}.
In: \bbtitle{2005 IEEE/RSJ International Conference on Intelligent Robots and
  Systems},
pp. \bfpage{708}--\blpage{713}
(\byear{2005}).
\bcomment{IEEE}
\end{bchapter}
\endbibitem

%%% 100
\bibitem{mech2007wolves}
\begin{bbook}
\bauthor{\bsnm{Mech}, \binits{L.D.}},
\bauthor{\bsnm{Boitani}, \binits{L.}}:
\bbtitle{Wolves: Behavior, Ecology, and Conservation}.
\bpublisher{University of Chicago Press}, \blocation{???}
(\byear{2007})
\end{bbook}
\endbibitem

%%% 101
\bibitem{quick2012bottlenose}
\begin{barticle}
\bauthor{\bsnm{Quick}, \binits{N.J.}},
\bauthor{\bsnm{Janik}, \binits{V.M.}}:
\batitle{Bottlenose dolphins exchange signature whistles when meeting at sea}.
\bjtitle{Proceedings of the Royal Society B: Biological Sciences}
\bvolume{279}(\bissue{1738}),
\bfpage{2539}--\blpage{2545}
(\byear{2012})
\end{barticle}
\endbibitem

%%% 102
\bibitem{schaller2009serengeti}
\begin{botherref}
\oauthor{\bsnm{Schaller}, \binits{G.B.}}:
The Serengeti lion: a study of predator-prey relations.
University of Chicago press
(2009)
\end{botherref}
\endbibitem

%%% 103
\bibitem{montesello1998implicit}
\begin{bchapter}
\bauthor{\bsnm{Montesello}, \binits{F.}},
\bauthor{\bsnm{D’Angelo}, \binits{A.}},
\bauthor{\bsnm{Ferrari}, \binits{C.}},
\bauthor{\bsnm{Pagello}, \binits{E.}}:
\bctitle{Implicit coordination in a multi-agent system using a behavior-based
  approach}.
In: \bbtitle{Distributed Autonomous Robotic Systems 3},
pp. \bfpage{351}--\blpage{360}.
\bpublisher{Springer}, \blocation{???}
(\byear{1998})
\end{bchapter}
\endbibitem

%%% 104
\bibitem{grupen2022multi}
\begin{bchapter}
\bauthor{\bsnm{Grupen}, \binits{N.A.}},
\bauthor{\bsnm{Lee}, \binits{D.D.}},
\bauthor{\bsnm{Selman}, \binits{B.}}:
\bctitle{Multi-agent curricula and emergent implicit signaling}.
In: \bbtitle{Proceedings of the 21st International Conference on Autonomous
  Agents and Multiagent Systems},
pp. \bfpage{553}--\blpage{561}
(\byear{2022})
\end{bchapter}
\endbibitem

%%% 105
\bibitem{gildert2018need}
\begin{barticle}
\bauthor{\bsnm{Gildert}, \binits{N.}},
\bauthor{\bsnm{Millard}, \binits{A.G.}},
\bauthor{\bsnm{Pomfret}, \binits{A.}},
\bauthor{\bsnm{Timmis}, \binits{J.}}:
\batitle{The need for combining implicit and explicit communication in
  cooperative robotic systems}.
\bjtitle{Frontiers in Robotics and AI}
\bvolume{5},
\bfpage{65}
(\byear{2018})
\end{barticle}
\endbibitem

%%% 106
\bibitem{haakansson2013communication}
\begin{bbook}
\bauthor{\bsnm{H{\aa}kansson}, \binits{G.}},
\bauthor{\bsnm{Westander}, \binits{J.}}:
\bbtitle{Communication in Humans and Other Animals}.
\bpublisher{John Benjamins Publishing Company Amsterdam}, \blocation{???}
(\byear{2013})
\end{bbook}
\endbibitem

%%% 107
\bibitem{bonacich2007some}
\begin{barticle}
\bauthor{\bsnm{Bonacich}, \binits{P.}}:
\batitle{Some unique properties of eigenvector centrality}.
\bjtitle{Social networks}
\bvolume{29}(\bissue{4}),
\bfpage{555}--\blpage{564}
(\byear{2007})
\end{barticle}
\endbibitem

%%% 108
\bibitem{xiao2005characterising}
\begin{bchapter}
\bauthor{\bsnm{Xiao}, \binits{B.}},
\bauthor{\bsnm{Wilson}, \binits{R.C.}},
\bauthor{\bsnm{Hancock}, \binits{E.R.}}:
\bctitle{Characterising graphs using the heat kernel}.
(\byear{2005})
\end{bchapter}
\endbibitem

%%% 109
\bibitem{jia2019graph}
\begin{bchapter}
\bauthor{\bsnm{Jia}, \binits{J.}},
\bauthor{\bsnm{Schaub}, \binits{M.T.}},
\bauthor{\bsnm{Segarra}, \binits{S.}},
\bauthor{\bsnm{Benson}, \binits{A.R.}}:
\bctitle{Graph-based semi-supervised \& active learning for edge flows}.
In: \bbtitle{Proceedings of the 25th ACM SIGKDD International Conference on
  Knowledge Discovery \& Data Mining},
pp. \bfpage{761}--\blpage{771}
(\byear{2019})
\end{bchapter}
\endbibitem

\end{thebibliography}

%\bibliography{refs}% common bib file


\end{document}